\documentclass[journal]{IEEEtran}
\usepackage{graphicx}
\usepackage{subfigure}
\usepackage{url}
\usepackage[ruled,linesnumbered]{algorithm2e}
\usepackage{amsmath}
\usepackage{amsfonts}
\usepackage{multirow}
\usepackage{picins}

\usepackage{cite}
\ifCLASSINFOpdf
\else
\fi
\begin{document}
\title{DiffuTraj: A Stochastic Vessel Trajectory Prediction Approach via Guided Diffusion Process}

\author{
Changlin~Li,
Yanglei~Gan,
Tian~Lan, ~\IEEEmembership{Member,~IEEE}, 
Yuxiang~Cai,
Xueyi~Liu,
Run~Lin,
Qiao~Liu$^*$, ~\IEEEmembership{Member,~IEEE}
\thanks{This work was supported by the National Natural Science Foundation of China (U22B2061), the National Key R\&D Program of China (2022YFB4300603), and Sichuan Science and Technology Program (2023YFG0151).}
\thanks{The authors are with the School of Computer Science and Engineering, University of Electronic Science and Technology of China, Qingshuihe Campus:No.2006, China (e-mail: changlinli@std.uestc.edu.cn; yangleigan@std.uestc.edu.cn; lantian1029@uestc.edu.cn; yuxiangcai@std.uestc.edu.cn; xueyiliu@std.uestc.edu.cn; runlin@std.uestc.edu.cn; qliu@uestc.edu.cn).}
\thanks{$^*$ Corresponding author.}
}


\markboth{Journal of \LaTeX\ Class Files,~Vol.~xx, No.~x, xxxxxx~xxxx}%
{Shell \MakeLowercase{\textit{Li et al.}}: Bare Demo of IEEEtran.cls for IEEE Journals}



\maketitle

\begin{abstract}
Maritime vessel maneuvers, characterized by their inherent complexity and indeterminacy, requires vessel trajectory prediction system capable of modeling the multi-modality nature of future motion states. Conventional stochastic trajectory prediction methods utilize latent variables to represent the multi-modality of vessel motion, however, tends to overlook the complexity and dynamics inherent in maritime behavior. In contrast, we explicitly simulate the transition of vessel motion from uncertainty towards a state of certainty, effectively handling future indeterminacy in dynamic scenes. In this paper, we present a novel framework (\textit{DiffuTraj}) to conceptualize the trajectory prediction task as a guided reverse process of motion pattern uncertainty diffusion, in which we progressively remove uncertainty from maritime regions to delineate the intended trajectory. Specifically, we encode the previous states of the target vessel, vessel-vessel interactions, and the environment context as guiding factors for trajectory generation. Subsequently, we devise a transformer-based conditional denoiser to capture spatio-temporal dependencies, enabling the generation of trajectories better aligned for particular maritime environment. Comprehensive experiments on vessel trajectory prediction benchmarks demonstrate the superiority of our method.
\end{abstract}

\IEEEpeerreviewmaketitle

\section{Introduction}
\IEEEPARstart{V}{essel} trajectory prediction is essential in maritime safety systems, aiding in functions, such as maritime traffic monitoring \cite{perera2012maritime}, collision avoidance \cite{pedrielli2019real}, anomaly detection \cite{ristic2008statistical}, and threat assessment \cite{lane2010maritime}. Most research utilizes the Automatic Identification System (AIS) for its extensive global data. However, developing accurate prediction models is challenging due to the complex nature of human behavior and various factors affecting vessel movement. \cite{rudenko2020human,capobianco2021uncertainty,jia2023conditional}.

Traditional physics-based models \cite{li2003survey, mazzarella2015knowledge} have proven effectiveness for short-term predictions but struggle with long-term accuracy due to their idealized assumptions. In response to this limitation, a shift towards Long Short-Term Memory (LSTM) networks and related variants \cite{tang2022model, wang2020ship, wang2020vessel} have emerged as representative approaches for vessel trajectory prediction. These models have been extended to the encoder-decoder framework, capitalizing on its success in sequence-to-sequence tasks \cite{forti2020prediction, capobianco2021deep, xiao2022bidirectional}. Despite the promising results from these advancements, vessel trajectory prediction continues to face several critical challenges:

\begin{figure}[t]
    \centering
    \includegraphics[scale=0.80]{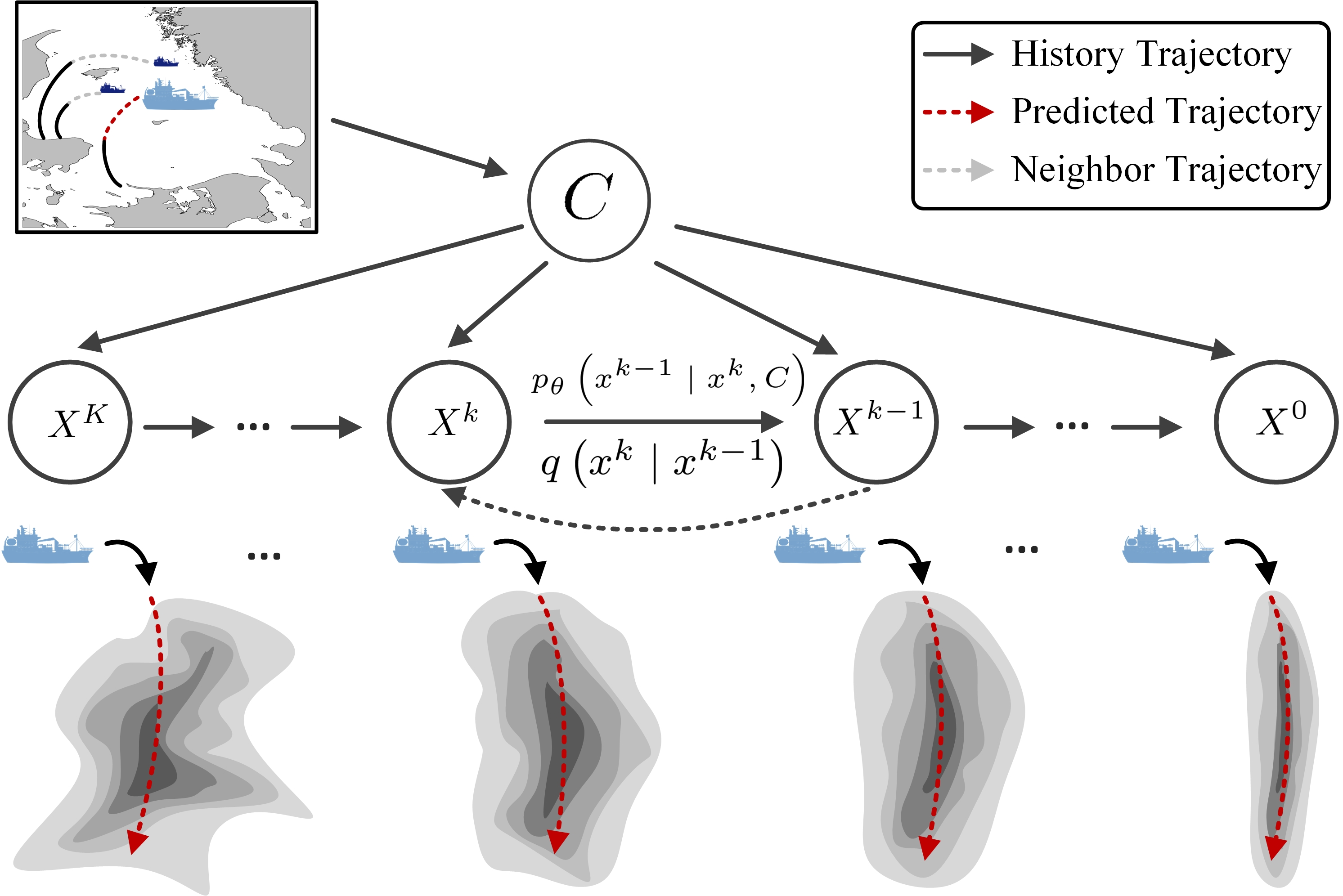}
    \caption{Illustration of the gradual transition in vessel motion from a state of uncertainty to certainty throughout the reverse diffusion process.}
    \label{fig1}
\end{figure}

\begin{itemize}

    \item \textbf{The prediction of vessel trajectories is inherently probabilistic and multi-modal}, underscoring the importance of accurately forecasting an unbiased distribution encompassing various potential future paths. Existing works have embraced probabilistic techniques, such as Gaussian process \cite{rong2019ship} and Bayesian learning \cite{capobianco2021uncertainty}, which offer enhanced representations of uncertainty and multi-modal aspects in maritime movement. However, these approaches face scalability challenges \cite{han2023interaction}, limiting their effectiveness in capturing complex patterns. Recent works \cite{gupta2018social,wang2021prediction, jia2023conditional} utilize deep generative models to spread the distribution over all possible future trajectories. However, it is noteworthy that these models utilize latent variables instead of explicitly modeling the uncertainty inherent in data distribution, introducing limitations in capturing the full complexity of vessel motions \cite{yang2021novel,pang2021trajectory,dhariwal2021diffusion}.
    % \item \textit{Accurate vessel trajectory prediction also entails the consideration of potential interactions among multiple agents}. Solely relying on independent marginal trajectory distributions for each agent can result in unrealistic and conflicting predictions. Recent studies \cite{liu2022stmgcn,wang2023novel} employ a graph convolutional neural network (GCN) to aggregate spatio-temporal interactions. For stochastic prediction methods, the trajectories of surrounding vessels are often encoded as contextual condition to model the interaction dynamics and  influence the generation process \cite{han2023interaction}. However, it's restricted to scenarios involving encounters between two vessels.
    % \cite{han2023interaction} or face over-smoothing problems on node features \cite{Liang23STGlow}, potentially leading to the loss of distinctive behavioral characteristics of vessels.
    \item \textbf{Vessel motions are significantly influenced by the surrounding scene context} \cite{yao2023lidar,wang2023novel}, encompassing both physical constraints and distinct motion patterns varying across geographical regions. In this regard, it becomes essential to incorporate physical priors into the modeling process \cite{jiang2023motiondiffuser,bansal2023universal} to avoid potential collision risks. Additionally, spatial interactions between neighboring vessels should be taken into account. Even if vessels have relatively certain long-term goals, their actual motions are influenced by the complex interaction dynamics between different vessels, especially during close encounters \cite{han2023interaction}. Such integration is crucial for developing realistic maritime scenarios that accurately reflect the diverse conditions and navigational behaviors encountered in different maritime environments.
    
\end{itemize}

In light of these challenges, we present DiffuTraj, a novel diffusion-based stochastic 
vessel trajectory prediction model. DiffuTraj directs each stage of the reverse diffusion process, utilizing a mix of interactive and physical information as guiding factors for trajectory generation. This includes leveraging past trajectories of the ego and surrounding vessels, along with the scene context, to offer a range of diverse and plausible future trajectories. In contrast to other latent variable generative models, we explicitly model the evolution of vessel motion as it progresses from a state of uncertainty to a state of certainty.

DiffuTraj operates through a twofold implementation: 1) the forward diffusion process methodically introduces Gaussian noise into pristine future trajectories; 2) the reverse diffusion process strategically reduces indeterminacy, transforming ambiguous predictions into clear, deterministic paths. Specifically, DiffuTraj encodes historical trajectories of both the target vessel and its nearby counterparts to capture potential interactions among multiple agents, transforming this data into state embeddings. In addition, the model incorporates static scene context with past trajectories, forming a detailed 2D pixel-level heatmap. This heatmap is subsequently converted into a state vector through a convolutional neural network (CNN), ensuring the inclusion of physical prior in the modeling process. The embeddings from the grayscale heatmap and past motion trajectories are concatenated to serve as condition to guide the reverse diffusion learning process. A Transformer-based decoder captures the complex spatio-temporal dependencies, ensuring a comprehensive understanding of the evolving maritime environment. During inference, instead of employing an autoregressive approach, DiffuTraj generates all future trajectories concurrently. It begins by sampling noisy trajectories  from a predefined Gaussian distribution, then progressively deriving accurate future trajectories through the learned reverse diffusion process, avoiding the limitations caused by error accumulation.

Our main contributions can be summarized as follows:
\begin{itemize}
    \item We present a novel stochastic vessel trajectory prediction framework based on guided motion pattern uncertainty diffusion process, which gradually reduces uncertainty from maritime regions to define the intended trajectory.
    \item To the best of our knowledge, our model is the first to introduce scene context jointly with spatial interactions of the vessels to generate spatial-sensitive and physically-plausible future trajectories.
    \item We conduct extensive experiments on real-world AIS dataset. Experimental results demonstrate that our model outperforms the state-of-the-art methods.
\end{itemize}

 

\section{Related work}
\textbf{Vessel Trajectory Prediction} \quad In the field of maritime navigation systems, traditional approaches predominantly rely on well-established kinematic models \cite{li2003survey,millefiori2016modeling} and statistical techniques \cite{mazzarella2015knowledge,alizadeh2021prediction}. However, given the challenges in modeling the intricate long-term dynamics of vessel trajectories, there is a burgeoning interest in harnessing machine learning and deep learning methodologies. Long Short-Term Memory (LSTM) networks and their variants have emerged as prominent methods in vessel trajectory prediction, adeptly capturing temporal dynamics. Gao et al. \cite{gao2018online} have introduced a bidirectional LSTM framework to enhance the correlation between past and future trajectories, while Wang et al. \cite{wang2020ship} have incorporated an attention mechanism, enabling a focused analysis of various hidden information aspects. Additionally, Gao et al. \cite{gao2021novel} have leveraged cubic interpolation alongside LSTM networks for multi-step trajectory prediction, integrating physical constraints to refine the accuracy of their predictions.

Despite these advancements, LSTM models still face challenges, particularly with handling long-term dependencies and computational efficiency, which are crucial in complex maritime environments. To address these issues, the encoder-decoder architecture, known for its proficiency in sequence-to-sequence tasks, has become increasingly popular in vessel trajectory prediction \cite{murray2020dual, forti2020prediction, capobianco2021deep, xiao2022bidirectional}. Those approaches often integrate various seq2seq networks to model vessel behavior using bidirectional historical trajectory data. Furthermore, Nguyen et al. \cite{nguyen2021traisformer} have introduced a Transformer-based framework, reconceptualizing trajectory prediction as a classification problem of discrete values by embedding attributes into high-dimensional discrete vectors.
Besides, recent studies \cite{liu2022stmgcn,wang2023novel} have incorporated spatial interactions among nearby vessels based on a graph convolutional neural network (GCN) to enhance prediction accuracy.
% Furthermore, Nguyen et al. \cite{nguyen2021traisformer} have introduced a Transformer-based framework, reconceptualizing trajectory prediction as a classification problem of discrete values by embedding attributes into high-dimensional discrete vectors. 
However, our approach distinguishes from these deterministic frameworks as it aims to model the multi-modality of future vessel motion states. In our work, we incorporate information from both spatial interactions and the scene context of vessels to represent the diverse future motion states.
\mbox{}\\
\textbf{Uncertainty Modeling} \quad In the realm of vessel trajectory prediction, recent research efforts have increasingly focused on capturing multi-modalities and modeling the future distribution of trajectories through enhanced uncertainty modeling. For example, Rong et al. \cite{rong2019ship} have characterized the uncertainty in future positions along the vessel trajectories by continuous probability distributions, while Liu et al \cite{liu2022stmgcn} have assumed the future vessel positions follow a Gaussian distribution. Capobianco et al. \cite{capobianco2021uncertainty} have extended the deep learning framework by generating the corresponding prediction uncertainty via Bayesian modeling of epistemic and aleatoric uncertainties.
Generative Adversarial Networks (GANs), as explored in studies by Jia et al. \cite{jia2023conditional} and Wang et al. \cite{wang2021prediction}, have been proposed to generate multiple potential future trajectories. Zhang et al. \cite{zhang2024dynamic} have proposed a multi-stage trajectory prediction method that employs GANs for trajectory augmentation. 
The experimental results of GANs show a remarkable improvement over seq2seq models. Concurrently, Han et al. \cite{han2023interaction} and Nguyen et al. \cite{nguyen2021geotracknet} have leveraged Conditional Variational Autoencoders (CVAEs) and Variational Recurrent Neural Networks (VRNNs) to structure and learn distributions through variational inference. However, these stochastic methods typically rely on latent variables, and are likely to lead to issues such as unstable training or the generation of unnatural trajectories \cite{dhariwal2021diffusion, gu2022stochastic}. In our study, we explicitly simulate the transition of vessel motion from uncertainty to certainty, effectively managing future indeterminacy in dynamic scenes.
\mbox{}\\
\textbf{Denoising Diffusion Probabilistic Models} \quad Denoising diffusion probabilistic models (Diffusion models) \cite{ho2020denoising, song2019generative} are a class of generative models inspired by non-equilibrium thermodynamics \cite{sohl2015deep}. Most existing works of diffusion models have demonstrated high sample quality across a wide range of real-world domains, such as vision \cite{dhariwal2021diffusion, rombach2022high, nichol2021improved} and audio \cite{chen2021wavegrad, kong2021diffwave}. In particular, Gu et al. \cite{gu2022stochastic} have pioneered the adaptation of the diffusion framework, focusing on short-term predictions conditioned on past trajectories. Diffuser \cite{janner2022planning} has generated trajectories for planning and control in robotics applications, incorporating guidance during test-time. Rempe et al. \cite{rempe2023trace} have incorporated map conditioning into an expressive feature grid queried in denoising and applied classifier-free sampling during the inference stage. In the vehicle traffic domain, CTG \cite{zhong2023guided} has developed a controllable model emphasizing adherence to formal traffic rules and ensuring the physical feasibility of generated trajectories. Another emerging focus within diffusion models is enhancing the sampling process for real-time applications. DDIM by Song et al. \cite{song2020denoising} has initially deduced the state of the original data and then deterministically forecasts the transition towards the next expected timestamp using a non-Markov process. Further advancing this line, Mao et al. \cite{mao2023leapfrog} have introduced LED, the first model to improve trajectory prediction sampling efficiency by employing a trainable leapfrog initializer to reduce the number of denoising steps required. Building on these developments, we are among the first to use diffusion models for predicting the future trajectories of vessels.

\section{Preliminaries}
\subsection{Problem Formulation}
The goal of our work is to predict future trajectories for each target vessel based on the scene context, past observations of itself and surrounding vessels. Here, the scene information is fed as a high-resolution scene image \(\mathcal{I}\).
Let \(\mathbf{x}^{tar}_{-L+1:0} = [\mathbf{x}_{-L+1}^{tar}, \mathbf{x}^{tar}_{-L+2},...,\mathbf{x}^{tar}_0]\) denote as the observed past state trajectory over \(L\) timestamps where the state \(\mathbf{x}^{tar}_t = ( lat, lon)\) respectively indicated the 2D latitude and longitude spatial coordinates at timestamp \(t\).
Let \(\mathbf{X}^{neigh}\) be the neighbouring vessel trajectory set, we denote \(\mathbf{X}^{neigh} = \left\{\mathbf{x}_{-L+1:0}^{i}\right\}_{i=1}^N\) as the past trajectories of \(N\) surrounding vessels, where \(\mathbf{x}^{i}_{-L+1:0}\) is the trajectory of the \(i\)-th neighbour. The corresponding ground-truth future trajectory then can be written as \(\mathbf{x}^{tar}_{1:H} = [\mathbf{x}^{tar}_1, \mathbf{x}^{tar}_2,...,\mathbf{x}^{tar}_H]\) over the next \(H\) timestamps. For clarity, we use \(\mathbf{x}\) without the former superscript for the target vessel trajectory and formulate \(\mathbf{x}_{1:H}^0 = \mathbf{x}_{1:H}\) to represent the future states.

\begin{figure*}[h]
    \centering
    \includegraphics[scale=0.95]{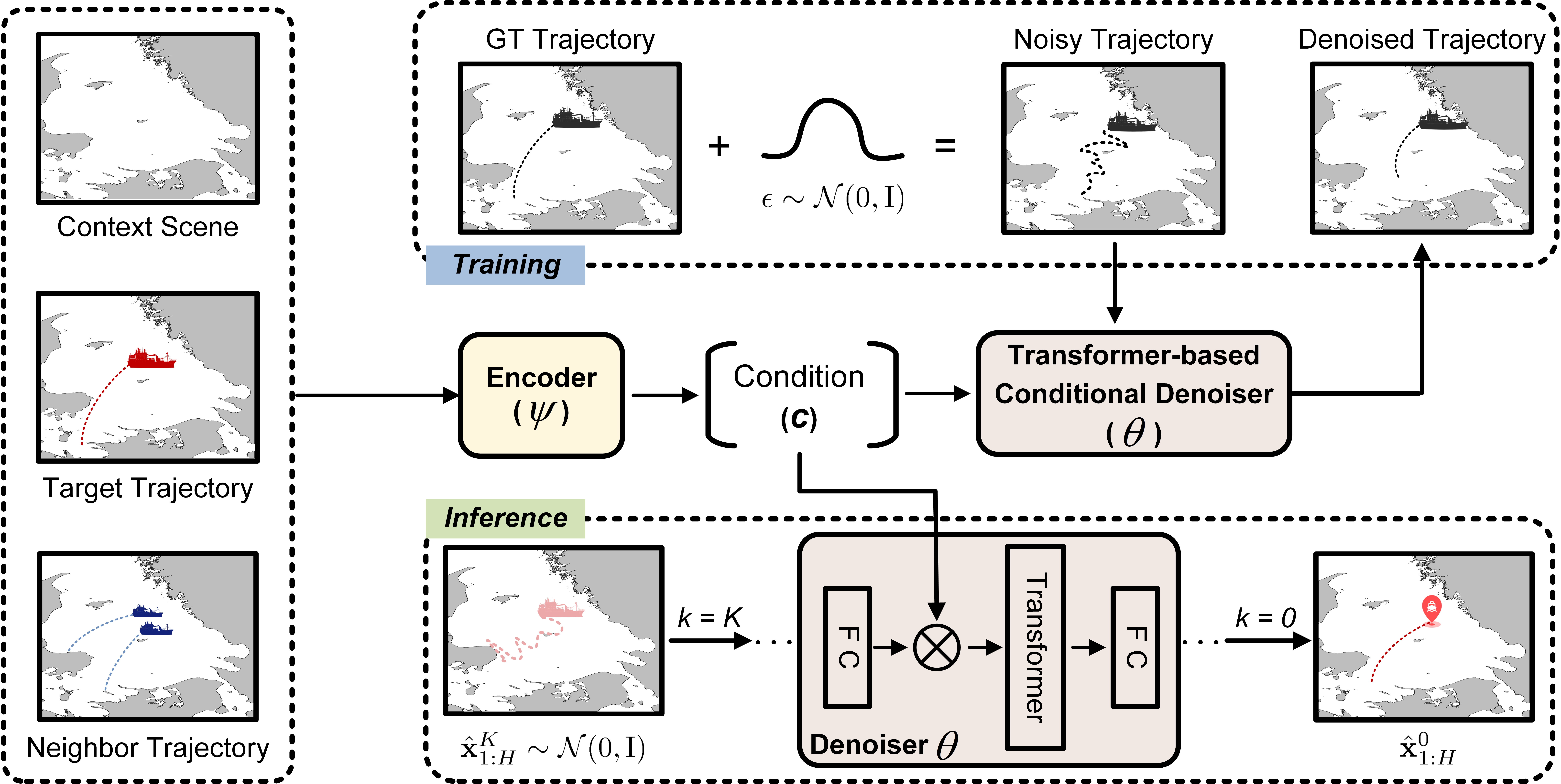}
    \caption{The framework of our DiffuTraj model. $\otimes$ denotes multiplication. $FC$ denotes fully-connected layer. The comprehensive details of the Encoder ($\psi$) and the Transformer-based Conditional Denoiser ($\theta$) are elaborated in section \ref{sec4.2} and \ref{sec4.3}, respectively.}
    \label{fig2}
\end{figure*}

\subsection{Diffusion Model In Trajectory Prediction}
\textbf{Preliminaries} \quad Diffusion models consist of a forward diffusion process and a backward denoising process. The core idea is to learn and refine the trajectory distribution by gradually reducing the indeterminacy from noise distribution with a parameterized Markov chain.
The forward diffusion process is formulated as the posterior distribution $q(\mathbf{x}^{1: K}_{1:H} \mid \mathbf{x}^0_{1:H})$, and it is fixed to a Markov chain that intentionlly adds a series of Gaussian noise in \(K\) steps:
\begin{equation}
    q(\mathbf{x}^{1: K}_{1:H} \mid \mathbf{x}^0_{1:H}):=\prod_{k=1}^K q(\mathbf{x}^k_{1:H} \mid \mathbf{x}^{k-1}_{1:H})
\end{equation}
\begin{equation}
    q(\mathbf{x}^k_{1:H} \mid \mathbf{x}^{k-1}_{1:H}):=\mathcal{N}(\mathrm{x}^k_{1:H} ; \sqrt{1-\beta_k} \mathbf{x}^{k-1}_{1:H}, \beta_k \mathbf{I}),
\end{equation}
where $\beta_k \in[0,1]$ is the noise variance following a predefined schedule. Starting from a clean future trajectory, it can be shown in a closed form for any step \(k\) as:
\begin{equation}\label{eq:forward diffusion}
    q(\mathbf{x}^k_{1:H} \mid \mathbf{x}^0_{1:H} ):=\mathcal{N}(\mathbf{x}^k_{1:H} ; \sqrt{\bar{\alpha}_k} \mathbf{x}^0_{1:H},(1-\bar{\alpha}_k) \mathbf{I}),
\end{equation}
where \(\alpha_k:=1-\beta_k\) and \(\bar{\alpha}_k:=\Pi_{s=1}^k \alpha_s\). In the opposite direction, the future states generation procedure can be considered as a reverse dynamics of the above
diffusion process defined as a Markov chain with learned Gaussian transitions starting at $p(\mathbf{x}^K_{1:H})=\mathcal{N}(0, \mathbf{I})$. We formulate the reverse diffusion process as follows:
\begin{equation}
    p_\theta(\mathbf{x}^{0: K}_{1:H} ):=p(\mathbf{x}^K_{1:H}) \prod_{k=1}^K p_\theta(\mathbf{x}^{k-1}_{1:H} \mid \mathbf{x}^k_{1:H} )
\end{equation}
\begin{equation}
    p_\theta(\mathbf{x}^{k-1}_{1:H} \mid \mathbf{x}^k_{1:H} ):=\mathcal{N}(\mathbf{x}^{k-1}_{1:H};{\mu}_\theta(\mathbf{x}^k_{1:H}, k ), \Sigma_\theta(\mathbf{x}^k_{1:H}, k)),
\end{equation}
\newline where \(\Sigma_\theta(\mathbf{x}^k_{1:H}, k)\) is often fixed at \(\sigma_k^2 \mathbf{I}=\beta_k \mathbf{I}\), and \(\mu_\theta(\mathbf{x}^k_{1:H}, k)\) is the estimated mean parameterized by \(\theta\) in practice. To train the diffusion model, we uniformly sample \(k\) from \(\{1,2, \ldots, K\}\) and minimize the following \textbf{KL} divergence:
\begin{equation}
    \mathcal{L}_k=D_{\mathrm{KL}}(q(\mathbf{x}^{k-1}_{1:H} \mid \mathbf{x}^k_{1:H})|| p_\theta(\mathbf{x}^{k-1}_{1:H} \mid \mathbf{x}^k_{1:H})).
\end{equation}
For more tractable training, the approximate posterior is often replaced by $q(\mathbf{x}^{k-1}_{1:H} \mid \mathbf{x}^k_{1:H}, \mathbf{x}^0_{1:H})$ and can be represented by Gaussian distribution as follows:
\begin{equation}
    q(\mathbf{x}^{k-1}_{1:H} \mid \mathbf{x}^k_{1:H}, \mathbf{x}^0_{1:H})=\mathcal{N}(\mathbf{x}^{k-1}_{1:H} ; \tilde{\mu}_k(\mathbf{x}^k_{1:H}, \mathbf{x}^0_{1:H}, k), \tilde{\beta}_k \mathrm{I}),
\end{equation}
where the closed form of the mean and variance are denoted as:
\begin{equation}
    \tilde{\mu}_k(\mathbf{x}^k_{1:H}, \mathbf{x}^0_{1:H}, k)=\frac{\sqrt{\bar{\alpha}_{k-1}} \beta_k}{1-\bar{\alpha}_k} \mathbf{x}^0_{1:H}+\frac{\sqrt{\alpha_k}(1-\bar{\alpha}_{k-1})}{1-\bar{\alpha}_k} \mathbf{x}^k_{1:H}
\end{equation}
\begin{equation}
    \tilde{\beta}_k=\frac{1-\bar{\alpha}_{k-1}}{1-\bar{\alpha}_k} \beta_k \mathrm{I}.
\end{equation}
Since both forward diffusion process and reverse process are Gaussian, we naturally convert the training objective into calculating the difference between the means as:
\begin{equation}
    \mathcal{L}_k=\frac{1}{2 \sigma_k^2}\|\tilde{\mu}_k(\mathbf{x}^k_{1:H}, \mathbf{x}^0_{1:H}, k)-\mu_\theta(\mathbf{x}^k_{1:H}, k)\|^2.
\end{equation}
We apply a noise prediction model $\epsilon_\theta(\mathbf{x}^k_{1:H}, k)$ to reparameterize:
\begin{equation}
    \mu_\theta(\mathbf{x}^k_{1:H}, k)=\frac{1}{\sqrt{\alpha_k}} \mathbf{x}^k_{1:H}-\frac{1-\alpha_k}{\sqrt{1-\bar{\alpha}_k} \sqrt{\alpha_k}} \epsilon_\theta(\mathbf{x}^k_{1:H}, k),
\end{equation}
and obtain a simplified loss function that leads to better generation quality as shown in \cite{ho2020denoising}:
\begin{equation}\label{eq:loss}
    \mathcal{L}_k=\mathbb{E}_{k, \mathbf{x}^0_{1:H}, \epsilon}[\|\epsilon-\epsilon_\theta(\mathbf{x}^k_{1:H}, k)\|^2],
\end{equation}
where $\epsilon \sim \mathcal{N}(0, \mathbf{I})$ is the noise used to obtain $\mathbf{x}^k_{1:H}$ from $\mathbf{x}^0_{1:H}$ in the forward diffusion process.
\newline\textbf{Conditional Diffusion Model} \quad In this work, we are interested in the conditional setting of learning the reverse diffusion process.
Figure \ref{fig1} illustrates the initial ambiguous region $\mathbf{x}^K$ under the noise distribution and the target trajectory $\mathbf{x}^0$ under the data distribution. We describe the diffusion process as $(\mathbf{x}^0, \mathbf{x}^1, \ldots, \mathbf{x}^K)$, with $K$ representing the maximum diffusion steps. This process incrementally introduces uncertainty until the actual trajectory becomes a noisy, indistinct region. Conversely, the reverse process $(\mathbf{x}^K, \mathbf{x}^{K-1}, \ldots, \mathbf{x}^0)$ is learned to gradually diminish this uncertainty to reconstruct the trajectories. Throughout the reverse process, the conditioning variable $\mathbf{c}$ is applied at each step. Both diffusion and reverse diffusion are modeled as Markov chains with Gaussian transitions.
Specifically, DiffuTraj performs trajectory prediction by modeling the joint distribution as follows:
\begin{equation}
    p_\theta(\mathbf{x}_{1: H}^{0: K} \mid \mathbf{c})=p_\theta(\mathbf{x}_{1: H}^K) \prod_{k=1}^K p_\theta(\mathbf{x}_{1: H}^{k-1} \mid \mathbf{x}_{1: H}^k, \mathbf{c}),
\end{equation}
where $\mathbf{x}_{1: H}^K \sim \mathcal{N}(0, \mathbf{I})$ and \(\mathbf{c}\) is the state embedding derived by using neural networks to encode the historical sequence \((\mathbf{x}^{tar}_{-L+1:0}, \mathbf{X}^{neigh}, \mathcal{I})\) including observed trajectories and scene context, and
\begin{equation}
    p_\theta(\mathbf{x}_{1: H}^{k-1} \mid \mathbf{x}_{1: H}^k, \mathbf{c})=\mathcal{N}(\mathbf{x}_{1: H}^{k-1} ;  \mu_\theta(\mathbf{x}_{1: H}^k, k \mid \mathbf{c}), \sigma_k^2 \mathbf{I}).
\end{equation}
The conditional denoiser then can be learned by minimizing a conditioned variant of equation [\ref{eq:loss}] at every step \(k\) as follows:
\begin{equation}\label{eq:condition loss}
    \mathcal{L}_k=E_{k, \mathbf{x}^0_{1:H}, \epsilon}[\|\epsilon-\epsilon_\theta(\mathbf{x}^k_{1:H}, k \mid\mathbf{c})\|^2].
\end{equation}

\section{Methodology}
In this section, we introduce our DiffuTraj method, which performs stochastic trajectory prediction by denoising diffusion process. To implement accurate and realistic prediction, the future route taken by each vessel needs to be influenced not only by its own state history, but also the states of surrounding vessels and scene context around its path. Our model takes all these cues into account. Specifically, DiffuTraj consists of two trainable modules, an encoder with parameters \(\psi\) to extract joint condition features, and a decoder parameterized by \(\theta\) for the backward denoising process. An overview of the whole architecture is depicted in Figure \ref{fig2}. We will introduce each part in detail in the following.

\subsection{Forward Diffusion Process}
In DiffuTraj, the forward diffusion process transforms a clean future trajectory \(\mathbf{x}_{1:H}^0\) to a white Gaussian noise vector \(\mathbf{x}_{1:H}^k\) in \(k\) diffusion steps based on equation [\ref{eq:forward diffusion}].
Thus, \(\mathbf{x}^k_{1:H}\) can be directly obtained as:
\begin{equation}
    \mathbf{x}^k_{1:H}=\sqrt{\bar{\alpha}_k} \mathbf{x}^0_{1:H}+\sqrt{1-\bar{\alpha}_K} \epsilon,
\end{equation}
where \(\epsilon\) is sampled from \(\mathcal{N}(0, \mathbf{I})\) with the same size as \(\mathbf{x}_{1: H}^0\).

\subsection{Conditioning the Backward Denoising Process}\label{sec4.2}
The encoder network \(\psi\) consists of three major components. We first explore two pure LSTM-based encoders to learn the state embeddings \(\mathbf{Z}_{en}\) and \(\mathbf{Z}_{neigh}\) by observed history trajectories of ego and surrounding vessels. Then we integrate the static scene context with past trajectory of the target vessel into a fused heatmap at the 2D pixel-level, and then use a 2D-CNN to extract joint features into embedding \(\mathbf{Z}_{map}\). Finally, these embeddings are concatenated along the feature dimension to form the condition:
\begin{algorithm}[tb]
   \caption{Training}
   \label{alg:example}
   \Repeat{converged}{
      $\mathbf{x}^0_{1:H} \sim q(\mathbf{x}^0_{1:H})$\;
      $k \sim {Uniform}(\{1,2, \ldots, K\})$\;
      $\epsilon \sim \mathcal{N}(0, \mathbf{I})$\;
      obtain condition $\mathbf{c}$ by encoder $\psi$\;
      Take gradient descent step on \newline $\nabla_\theta\left\|\epsilon-\epsilon_\theta(\sqrt{\bar{\alpha}_k} \mathbf{x}^0_{1:H} + \sqrt{1-\bar{\alpha}_k} \epsilon, k \mid \mathbf{c})\right\|^2 $\;
   }
\end{algorithm}
\begin{equation}
    \mathbf{c}=\operatorname{concat}([\mathbf{Z}_{en}, \mathbf{Z}_{neigh}, \mathbf{Z}_{map}]) \in \mathbb{R}^{3 \times d}.
\end{equation}
\textbf{Encoding Historical Information} \quad To encode the observed history \(\mathbf{x}_{-L+1:0} = [\mathbf{x}_{-L+1}, \mathbf{x}_{-L+2},...,\mathbf{x}_0]\) of the target vessel, an LSTM network is used to capture the temporal dependency between all states and encode them into a high dimensional feature representation \(\mathbf{Z}_{en}\), denoted as:
\begin{equation}
    \mathbf{Z}_{en}=L S T M_{e n}(\mathbf{x}_{-L+1:0}).
\end{equation} 
\textbf{Encoding Spatial Interactions} \quad We first abstract the scene for each timestamp \(t\) as a spatio-temporal directed graph \(G=(V, E)\) to model the neighbor's information of each vessel, where nodes represent vessels and edges represent their spatial interactions. The Euclidean distance \(\ell_2 = \|(x_i-x_j)^2+(y_i-y_j)^2\|_2\) is calculated as a proxy for whether vessels are influenced by each other, where the threshold for impact distance is set as a hyperparameter and \((x_i,y_i),(x_j,y_j)\) are 2D longitude and latitude coordinates. As the number of neighbors for each vessel is time-varying, their past trajectories needs to be aggregated. Given the neighbouring vessel  trajectory set \(\mathbf{X}^{neigh} = \left\{\mathbf{x}_{-L+1:0}^{i}\right\}_{i=1}^N\), the aggregated state \(\mathbf{x}^{sum}\) is calculated as follows:
\begin{equation}
    \mathbf{x}^{sum} = \sum_{i=1}^{N}{\mathbf{x}_{-L+1:0}^{i}},
\end{equation}
where we apply element-wise addition to preserve fixed structure of encoder inputs and count information. Accordingly, the subsequent steps are analogous to the previous encoder, but handle the features of the concatenation of ego and neighbors for a more informative representation \(\mathbf{Z}_{neigh}\), denoted as: 
\begin{equation}
    \mathbf{Z}_{neigh}=L S T M_{neigh}(\mathbf{x}^{sum} \oplus \mathbf{x}_{-L+1:0}).
\end{equation}
\textbf{Trajectory-on-Map Representation} \quad One main benefit of using a top-down high definition input representation is the simplicity to encode spatial information. However, 
encoding the image into a hidden state directly is discouraged since any meaningful spatial signals will get highly conflated when flattened into a vector \cite{mangalam2021goals}. To learn a meaningful context representation which effectively leverages scene context (image-like) with observed trajectory (2D coordinates), we create pixel-wise alignment between the different modalities. Compared to prior works \cite{kosaraju2019social, sadeghian2019sophie} based on image views that extract visual features by encoding them with convolutional neural networks (CNNs) separately, our approach operates on a trajectory-on-map representation, which integrates both static scene context and past trajectory into a fused heatmap, and better captures the joint features in the same space.
We first use a Gaussian kernel to convert the past trajectory \(\mathbf{x}_{-L+1:0} = [\mathbf{x}_{-L+1}, \mathbf{x}_{-L+2},...,\mathbf{x}_0]\) into a trajectory heatmap \(H\) of the same size as the image \(\mathcal{I}\), denoted as:
\begin{equation}
    K_t(x, y)=\exp \left(-\frac{\left(x-x_t\right)^2+\left(y-y_t\right)^2}{2 \sigma^2}\right)
\end{equation}
\begin{equation}
    H(x, y)=\sum_{t={-L+1}}^0 K_t(x, y),
\end{equation}
\newline where \(\sigma\) is the standard deviation of the Gaussian kernel and \((x_t, y_t)\) represents the longitude and latitude coordinates at timestamp \(t\). If two Gaussians overlap, we take the element-wise addition to ensure the intensity and smoothness of the trajectory heatmap.
To extract joint features from both scene context and past trajectory, we employ a 2D-CNN. The trajectory heatmap is then fused with the map image as:
\begin{equation}
    \mathbf{Z}_{map} = CNN(\alpha \odot H + (1-\alpha) \odot \mathcal{I}),
\end{equation}
where \(\alpha\) is a fixed parameter sampled from the uniform distribution on \([0,1)\). 

\begin{algorithm}[tb]
   \caption{Inference}
   \label{alg:example}
   $\hat{\mathbf{x}}^K_{1:H} \sim \mathcal{N}(0, \mathbf{I})$\;
   \For{$k=K, K-s, \ldots, s$}{
      Obtain condition \(\mathbf{c}\) by encoder \(\psi\)\;
      $\hat{\mathbf{x}}^{0} = \frac{1}{\bar{\alpha}_k}(\hat{\mathbf{x}}_{1: H}^k - \sqrt{1 - \bar{\alpha}_k}\epsilon_\theta(\hat{\mathbf{x}}_{1: H}^k, k \mid \mathbf{c}))$\;
      $\hat{\mathbf{x}}_{1: H}^{k-s} = \sqrt{\bar{\alpha}_{k-s}}\hat{\mathbf{x}}^{0} + \sqrt{1 - \bar{\alpha}_{k-s}}\epsilon_\theta(\hat{\mathbf{x}}_{1: H}^k, k \mid \mathbf{c})$\;
   }
   \Return \(\hat{\mathbf{x}}^0_{1:H}\)
\end{algorithm}

\begin{table*}[]
\caption{Trajectory forecast errors for different methods at varying lead times up to 4 hours(3 to 24 timestamps). The best results are highlighted in \textbf{bold}. The second best results are \underline{underlined}. All baseline models were re-implemented for consistency in comparison. $\dag$ denotes a reimplementation of the model by deterministically selecting the highest classification score. Note that ADE and FDE are in \textit{km}.}
\setlength{\tabcolsep}{3.5pt} % 减小列间距
\renewcommand{\arraystretch}{1.3}
\begin{tabular}{cccccccccccccccccc}
\hline
Data                        & Method           & \multicolumn{8}{c}{ADE}                                                                                                               & \multicolumn{8}{c}{FDE}                                                                                                               \\ \hline
                            &                  & 0.5\textit{h}              & 1\textit{h}              & 1.5\textit{h}              & 2\textit{h}             & 2.5\textit{h}             & 3\textit{h}             & 3.5\textit{h}             & 4\textit{h}             & 0.5\textit{h}              & 1\textit{h}              & 1.5\textit{h}              & 2\textit{h}             & 2.5\textit{h}             & 3\textit{h}             & 3.5\textit{h}             & 4\textit{h}             \\ \hline
\multirow{7}{*}{\rotatebox{90}{Danish Straits}} & Vanilla LSTM     & 1.057          & 2.316          & 3.679          & 5.126          & 6.236          & 7.688          & 9.276          & 10.827         & 1.768          & 4.556          & 7.554          & 10.707         & 13.095         & 16.374         & 20.084         & 23.949         \\
                            & LSTM-Seq2Seq     & 0.861          & 1.563          & 2.147          & 2.808          & 3.588          & 4.656          & 5.410          & 6.736          & 1.362          & 2.746          & 3.997          & 5.577          & 7.222          & 9.505          & 11.777         & 14.273         \\
                            & LSTM-Seq2Seq-Att & 0.884          & 1.544          & 2.132          & 2.806          & 3.700          & 4.443          & 5.593          & 6.626          & 1.397          & 2.685          & 3.944          & 5.487          & 7.510          & 9.509          & 11.833         & 14.777         \\
                            & TrAISformer$^\dag$      & 1.191          & 1.830          & 2.467          & 3.211          & 3.993          & 4.750          & 5.472          & 6.324          & 1.573          & 2.940          & 4.558          & 6.358          & 8.288          & 10.268         & 12.296         & 14.701         \\ \cline{2-18} 
                            & GANs              & 0.550          & 1.202          & 1.903          & 2.602          & 3.457          & 4.234          & 4.844          & 5.747          & 0.859          & 2.149          & 3.602          & 5.043          & 6.750          & 8.254          & 9.656          & 11.349         \\
                            & CVAE             & 0.524          & 0.976          & 1.513          & 2.124          & 2.730          & 3.476          & 3.998          & 4.482          & 0.575          & 1.311          & 2.343          & 3.536          & 4.781          & 6.235          & 7.557          & 7.697          \\ \cline{2-18} 
                            & DiffuTraj             & \textbf{0.210} & \textbf{0.436} & \textbf{0.787} & \textbf{1.172} & \textbf{1.698} & \textbf{2.238} & \textbf{2.849} & \textbf{3.458} & \textbf{0.242} & \textbf{0.704} & \textbf{1.411} & \textbf{2.204} & \textbf{3.385} & \textbf{4.378} & \textbf{5.711} & \textbf{7.110} \\ \hline
\\                        
                            \hline
                            &                  & 0.5\textit{h}              & 1\textit{h}              & 1.5\textit{h}              & 2\textit{h}             & 2.5\textit{h}             & 3\textit{h}             & 3.5\textit{h}             & 4\textit{h}             & 0.5\textit{h}              & 1\textit{h}              & 1.5\textit{h}              & 2\textit{h}             & 2.5\textit{h}             & 3\textit{h}             & 3.5\textit{h}             & 4\textit{h}             \\ \hline
\multirow{7}{*}{\rotatebox{90}{Baltic Sea}} & Vanilla LSTM     & 1.183          & 2.223          & 3.320          & 4.480          & 5.717          & 7.057          & 8.480          & 9.975         & 1.841          & 4.057          & 6.383          & 8.849         & 11.545         & 14.556         & 17.769         & 21.097         \\
                            & LSTM-Seq2Seq     & 0.981          & 1.542          & 2.236          & 3.007          & 3.852          & 4.785          & 5.670          & 6.693          & 1.461          & 2.665          & 4.081          & 5.725          & 7.602          & 9.601          & 11.439         & 13.613         \\
                            & LSTM-Seq2Seq-Att & 0.944          & 1.551          & 2.225          & 2.930          & 3.933          & 4.885          & 5.767          & 6.557          & 1.400          & 2.663          & 4.074          & 5.548          & 7.639          & 9.838          & 11.904         & 13.250         \\
                            & TrAISformer$^\dag$      & 1.340          & 2.037          & 2.805          & 3.628          & 4.536          & 5.526          & 6.524          & 7.642          & 1.790          & 3.254          & 4.963          & 6.849          & 9.075          & 11.487         & 13.754         & 16.427         \\ \cline{2-18} 
                            & GANs              & 0.785          & 1.499          & 2.255          & 3.020          & 3.852          & 4.962          & 5.688          & 6.796          & 1.188          & 2.636          & 4.152          & 5.749          & 7.496          & 9.697          & 11.409          & 14.063         \\
                            & CVAE             & 0.644          & 1.068          & 1.630          & 2.270          & 2.771          & 3.413          & 3.990          & 4.800          & 0.697          & 1.387          & 2.288          & 3.514          & 4.165          & 5.804          & 6.743          & 8.630          \\ \cline{2-18} 
                            & DiffuTraj             & \textbf{0.273} & \textbf{0.528} & \textbf{0.872} & \textbf{1.314} & \textbf{1.718} & \textbf{2.412} & \textbf{2.908} & \textbf{3.969} & \textbf{0.320} & \textbf{0.866} & \textbf{1.571} & \textbf{2.424} & \textbf{3.238} & \textbf{4.716} & \textbf{5.569} & \textbf{7.879} \\ \hline
\end{tabular}
\label{tab1}
\end{table*}

\begin{table}[]
\caption{Average displacement errors for different variations of DiffuTraj at varying lead times up to 4 hours on the Danish Straits testing set.}
\resizebox{\linewidth}{!}{
\renewcommand{\arraystretch}{1.3}
\begin{tabular}{ccccccccc}
\hline
Method       & \multicolumn{8}{c}{Average Displacement Error}                                                                                                               \\ \hline
             & 0.5\textit{h}              & 1\textit{h}              & 1.5\textit{h}              & 2\textit{h}             & 2.5\textit{h}             & 3\textit{h}             & 3.5\textit{h}             & 4\textit{h}             \\ \hline
w/o All        & 1.773          & 3.290          & 4.954          & 7.014          & 9.083               & 11.077               & 13.427               & 15.788               \\ \hline
w/o His\&Neigh & 0.455          & 0.893          & 1.412          & 2.045          & 2.714               & 3.621               & 4.902               & 5.954               \\
w/o Map\&Neigh & 0.228          & 0.535          & 1.041          & 1.720          & 2.583          & 3.661          & 5.017          & 5.860          \\ \hline
w/o His        & 0.436          & 0.876          & 1.295          & 1.867          & 2.489               & 3.275               & 4.541               & 5.622               \\
w/o Neigh      & 0.227          & 0.446          & 0.800          & 1.208          & 1.742               & 2.418               & 3.120               & 3.889               \\
w/o Map        & 0.223          & 0.531          & 1.042          & 1.705          & 2.548               & 3.597               & 4.956               & 5.803               \\ \hline
DiffuTraj    & \textbf{0.210} & \textbf{0.436} & \textbf{0.787} & \textbf{1.172} & \textbf{1.698} & \textbf{2.238} & \textbf{2.849} & \textbf{3.458} \\ \hline
\end{tabular}}
\label{tab2}
\end{table}

\begin{table}[]
\caption{Final displacement errors for different variations of DiffuTraj at varying lead times up to 4 hours on the Danish Straits testing set.}
\resizebox{\linewidth}{!}{
\renewcommand{\arraystretch}{1.3}
\begin{tabular}{ccccccccc}
\hline
Method       & \multicolumn{8}{c}{Final Displacement Error}                                                                                                               \\ \hline
             & 0.5\textit{h}              & 1\textit{h}              & 1.5\textit{h}              & 2\textit{h}             & 2.5\textit{h}             & 3\textit{h}             & 3.5\textit{h}             & 4\textit{h}             \\ \hline
w/o All        & 2.611          & 5.440          & 8.517          & 12.351         & 16.109               & 19.728               & 24.079               & 28.505               \\ \hline
w/o His\&Neigh & 0.626          & 1.415          & 2.408          & 3.672          & 4.962               & 6.922               & 9.506               & 11.319               \\
w/o Map\&Neigh & 0.271          & 0.949          & 2.094          & 3.543          & 5.367          & 7.505          & 10.538          & 12.269         \\ \hline
w/o His        & 0.599          & 1.408          & 2.184          & 3.326          & 4.550               & 6.153               & 8.763               & 11.288               \\
w/o Neigh      & 0.266          & 0.754          & 1.431          & 2.303          & 3.397               & 4.865               & 6.371               & 7.961               \\
w/o Map        & 0.261          & 0.963          & 2.062          & 3.432          & 5.241          & 7.477      & 10.245        & 11.738      \\ \hline
DiffuTraj    & \textbf{0.242} & \textbf{0.704} & \textbf{1.411} & \textbf{2.204} & \textbf{3.385} & \textbf{4.378} & \textbf{5.711} & \textbf{7.110} \\ \hline
\end{tabular}}
\label{tab3}
\end{table}

\subsection{Denoising Network}\label{sec4.3}
The denoising network \(\theta\) in this work is essentially a noise prediction network that takes the diffused future trajectory \(\mathbf{x}_{1:H}^k = [\mathbf{x}_1^k, \mathbf{x}_2^k,...,\mathbf{x}_H^k]\), denoising step \(k\) and condition \(\mathbf{c}\) as inputs. Specifically for the decoder architecture, we first calculate the time embedding \(k_{emb}\) as:
\begin{equation}
    {k_{emb}} = [\beta_k, \sin(\beta_k), \cos(\beta_k)],
\end{equation}
where \(\beta_k\) is from a fixed schedule and it is concatenated with the condition \(\mathbf{c}\) along the variable dimension to generate a new condition \(\mathbf{c}^{k}\). The input \(\mathbf{x}_{1:H}^k\) is transformed by an linear projection to obtain the diffused future trajectory embedding \(\mathbf{e}_1^k\) with the same feature dimension as \(\mathbf{c}^{k}\). Then we apply several fully-connected layers and take element-wise multiplication to explicitly upsample both input embedding \(\mathbf{e}_1^k\) and condition \(\mathbf{c}^{k}\) in feature dimensions to obtain the aggregated representation \(\mathbf{e}_2^k\), denoted as:
\begin{equation}
    w = sigmoid(FC_1(\mathbf{c}^{k}))
\end{equation}
\begin{equation}
    b = FC_2(\mathbf{c}^{k})
\end{equation}
\begin{equation}
    \mathbf{e}_2^k = \mathbf{W} \cdot 
 FC_3(\mathbf{e}_1^k) + \mathbf{B},
\end{equation}
where \(w \in R^{1 \times D}\) is broadcasted over timestamps to form \(\mathbf{W} \in R^{H \times D}\), and \(b \in R^{1 \times D}\) performs the same steps to form \(\mathbf{B} \in R^{H \times D}\). 
We also introduce the positional embedding before the last decoder layer to inject information about the relative position of different timestamps. Finally, a Transformer-based decoder consists of multiple self-attention layers is used to model the complex spatio-temporal dependencies of the fused features, where the self-attention module facilitates heightened interactions among timestamps by leveraging query, key, and values derived from the sequence representation, denoted as:
\begin{equation}
    \mathbf{e}_3^k = SelfAttention(PositionEncoding(\mathbf{e}_2^k))
\end{equation}
\begin{equation}
    \mathbf{e}^k = FFN(\mathbf{e}_3^k),
\end{equation}
where $FFN(\cdot)$ denotes a feed-forward network (FFN). With several fully-connected layers with the same aggregation operations, we downsample the output sequence \(\mathbf{e}^k\) to the trajectory dimension to form \(\epsilon_\theta(\mathbf{x}_{1: H}^k, k \mid \mathbf{c})\) with the same size as \(\mathbf{x}_{1: H}^k\). 
The denoised output is calculated in a deterministic way, denoted as:
\begin{equation}
    \hat{\mathbf{x}}_{1: H}^{k-s} = \sqrt{\bar{\alpha}_{k-s}}\hat{\mathbf{x}}^{0} + \sqrt{1 - \bar{\alpha}_{k-s}}\epsilon_\theta({\mathbf{x}}_{1: H}^k, k \mid \mathbf{c}),
\end{equation}
where \(s\) is the step decrement used to accelerate the reverse diffusion process and \(\hat{\mathbf{x}}^{0}\) is the estimation of the original trajectory states at step \(k\), calculated as:
\begin{equation}
    \hat{\mathbf{x}}^{0} = \frac{1}{\bar{\alpha}_k}(\hat{\mathbf{x}}_{1: H}^k - \sqrt{1 - \bar{\alpha}_k}\epsilon_\theta({\mathbf{x}}_{1: H}^k, k \mid \mathbf{c})).
\end{equation}

\begin{figure}[t]
    \centering
    \includegraphics[scale=0.28]{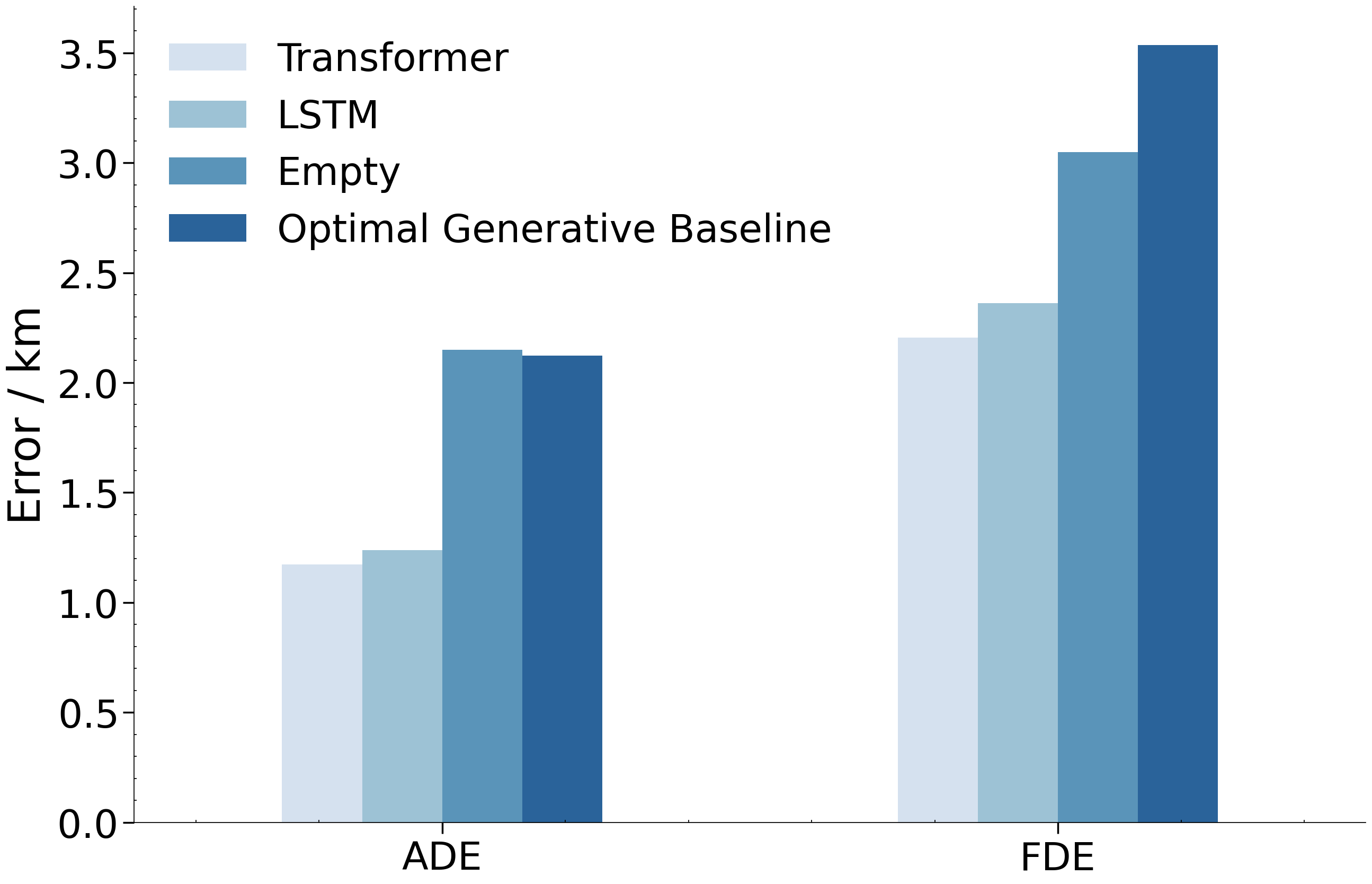}
    \caption{Ablation studies on decoder backbone network designs under the 2-hour prediction horizon on the Danish Straits dataset. \textbf{Empty} indicates trajectory generation without using the decoder backbone network.}
    \label{fig3}
\end{figure}

\subsection{Training Objective}
% The training objective consists of a penalty-reduced pixel-wise logistic regression with focal loss, and a MSE loss of noise. We discuss each component in detail in following part. 
The training procedure is shown in Algorithm 1. As mentioned in equation [\ref{eq:condition loss}], we minimize the following simplified MSE loss function:

\begin{equation}
    \mathcal{L}_k=\|\epsilon-\epsilon_\theta(\mathrm{x}_{1: H}^k, k \mid \mathbf{c})\|^2,
\end{equation}
where $\epsilon \sim \mathcal{N}(0, \mathrm{I})$ and the training is performed at each step $k \in \left\{1,2, \cdots, K\right\}$.

% \textbf{Pixel-wise Logistic Regression} \quad For each ground truth trajectory point of the added noise \(\epsilon\), we splat them onto a heatmap \(H_{\epsilon}\) using a Gaussian kernel as mentioned. The training objective between \(H_{\epsilon}\) and the noise prediction heatmap \(\hat{H_{\epsilon}}\) is a penalty-reduced pixelwise logistic regression with focal loss, denoted as:
% \[\mathcal{L}_m=-\frac{1}{N} \sum_{x y}\left\{\begin{array}{cl}(1-\hat{H_{\epsilon}}(x, y))^ \alpha \log (\hat{H_{\epsilon}}(x, y))
% \\ {if} H_{\epsilon}(x, y)>0.5 
% \\
% \\ ((1-{H_{\epsilon}(x, y)})^ \beta(\hat{H_{\epsilon}}(x, y))^\alpha
% \\ {otherwise}
% \end{array}\right.\]
% where α\alpha and β\beta are hyper-parameters of the focal loss[][], and NN is the number of peaks of the ground truth trajectory points. The normalization by \(N\) is chosen to standardize all positive instances of focal loss to 1. We use \(\alpha = 2\) and \(\beta = 4\) in our experiments, following [][].

% The overall training objective is represented as:
% L=Lm+Lk\mathcal{L}=\mathcal{L}_m+\mathcal{L}_k

\subsection{Inference}
The reverse diffusion process employs the non-Markovian denoising strategy DDIM \cite{song2020denoising}, which initiates by reconstructing the original data from a noisy state and then deterministically estimates the direction to the next target timestamp. Once both the encoder \(\psi\) and decoder \(\theta\) are trained, we can generate plausible trajectories following the inference procedure shown in Algorithm 2. Note that the initial input is a noise Gaussian $\hat{\mathbf{x}}_{1: H}^K \sim \mathcal{N}(0, \mathrm{I})$ and the iterations are executed $K/s$ times. We obtain the denoised trajectory $\hat{\mathbf{x}}_{1: H}^0$ as final prediction by iteratively applying the denoising steps with decreased timestamps.

\section{Experiments}
\subsection{Experimental Setup}
\textbf{Datasets} \quad In our study, we conduct evaluations using real-world AIS datasets, sourced from the Danish Maritime Authority (DMA)\footnote{\url{https://dma.dk/safety-at-sea/navigational-information/ais-data}}. We perform our experiments on AIS data derived from vessels sailing in different regions to evaluate vessel trajectory generation across various geographical characteristics and motion patterns. Additionally, we utilize high-resolution geographic coastline boundary data obtained from the Global Self-consistent, Hierarchical, High-resolution Geography Database (GSHHG)\footnote{\url{https://www.ngdc.noaa.gov/mgg/shorelines/shorelines.html}} as our scene information.

\textit{1) Danish Straits Dataset:}
The original dataset encompasses approximately 100,000 preprocessed AIS messages, representing a 10\% fixed sample of 1,367 distinct journeys undertaken by cargo and tanker vessels. The data spans the period from January 01, 2019, to March 31, 2019. For the purposes of this research, the Region of Interest (ROI) was delineated as a rectangular area extending from coordinates (55.5°, 10.3°) to (58.0°, 13.0°), allowing for focused analysis within this specified maritime zone.

\textit{2) Baltic Sea Dataset:} 
The employed dataset comprises 154,760 preprocessed AIS records of vessels sailing in the Baltic Sea during June 2023. Specifically, the dataset comprises AIS messages transmitted by 2,395 vessels and encompasses an area delineated by the Region of Interest (ROI) extending from coordinates (54.0°, 12.0°) to (56.0°, 15.0°). In contrast to the previous dataset, it focuses on evaluating the complex interaction scenarios characterized by shortened time spans and increased vessel counts.
% In our study, we conduct evaluations using a substantial real-world AIS dataset, sourced from the Danish Maritime Authority (DMA). This dataset encompasses approximately 100,000 preprocessed AIS messages, representing a 10\% fixed sample from 1,367 distinct journeys undertaken by cargo and tanker vessels. The data span from January 01, 2019, to March 31, 2019. Additionally, we utilize high-resolution geographic coastline boundary data obtained from the Global Self-consistent, Hierarchical, High-resolution Geography Database (GSHHG). For the purposes of this research, the Region of Interest (ROI) was delineated as a rectangular area extending from coordinates \((55.5^{\circ}, 10.3^{\circ})\) to \((58.0^{\circ}, 13.0^{\circ})\), allowing for focused analysis within this specified maritime zone.

\begin{figure*}
    \centering
    \subfigure{
        \includegraphics[width=0.35\linewidth]{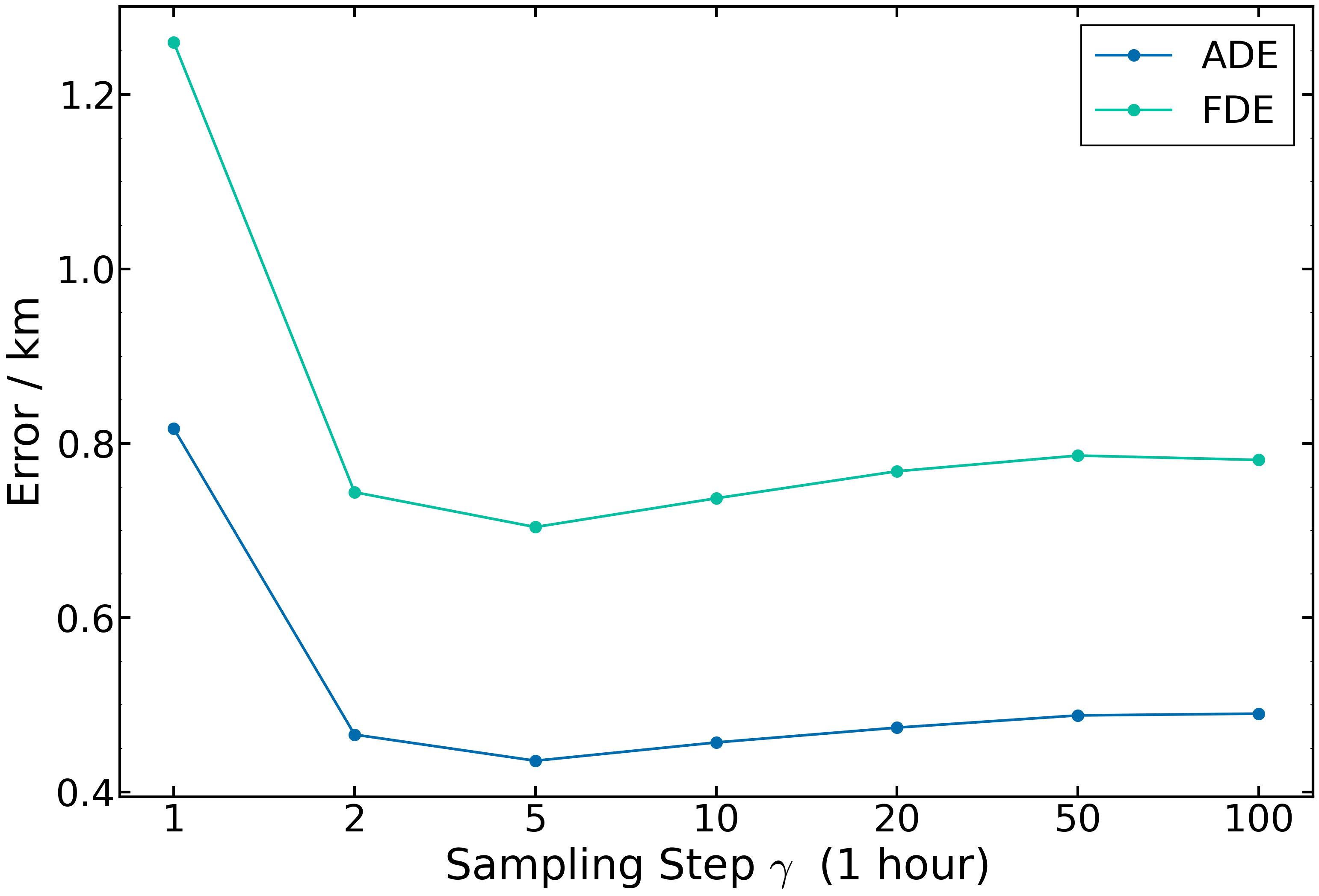} % 调整图片宽度
    }
    \hspace{80pt} % 添加水平空白，调整间距
    \subfigure{
        \includegraphics[width=0.35\linewidth]{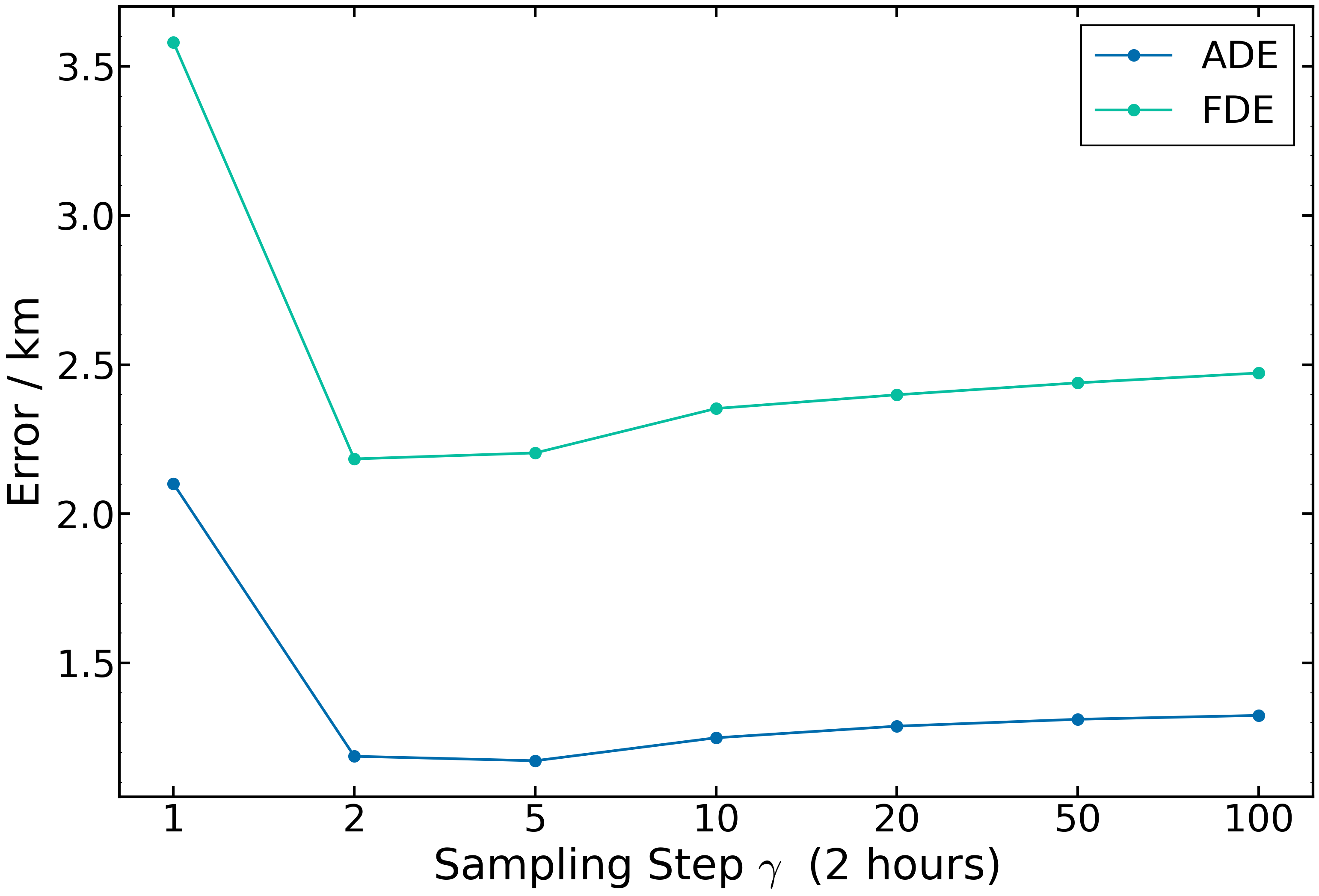} % 调整图片宽度
    } \\
    \subfigure{
        \includegraphics[width=0.35\linewidth]{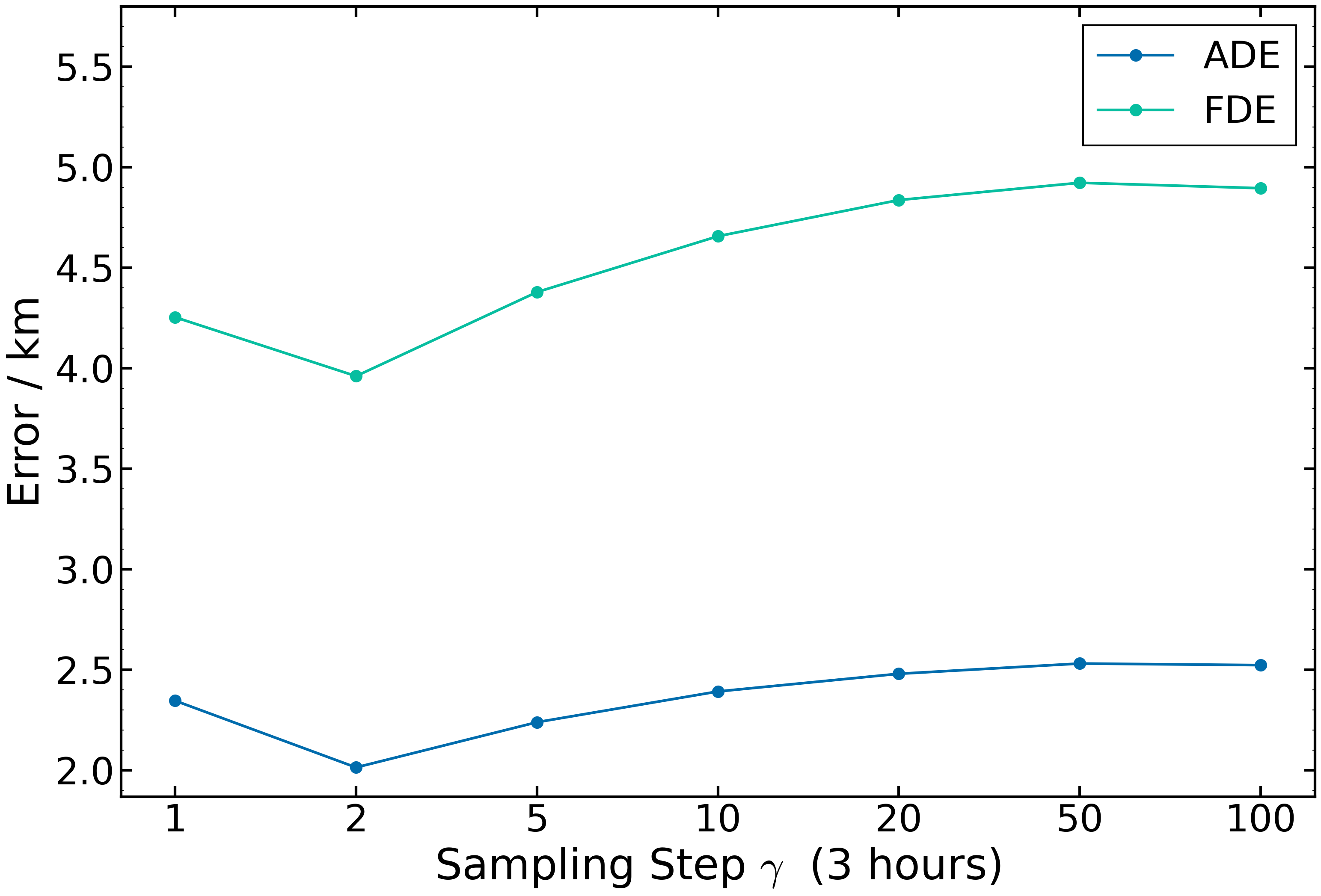} % 调整图片宽度
    }
    \hspace{80pt} % 添加水平空白，调整间距
    \subfigure{
        \includegraphics[width=0.35\linewidth]{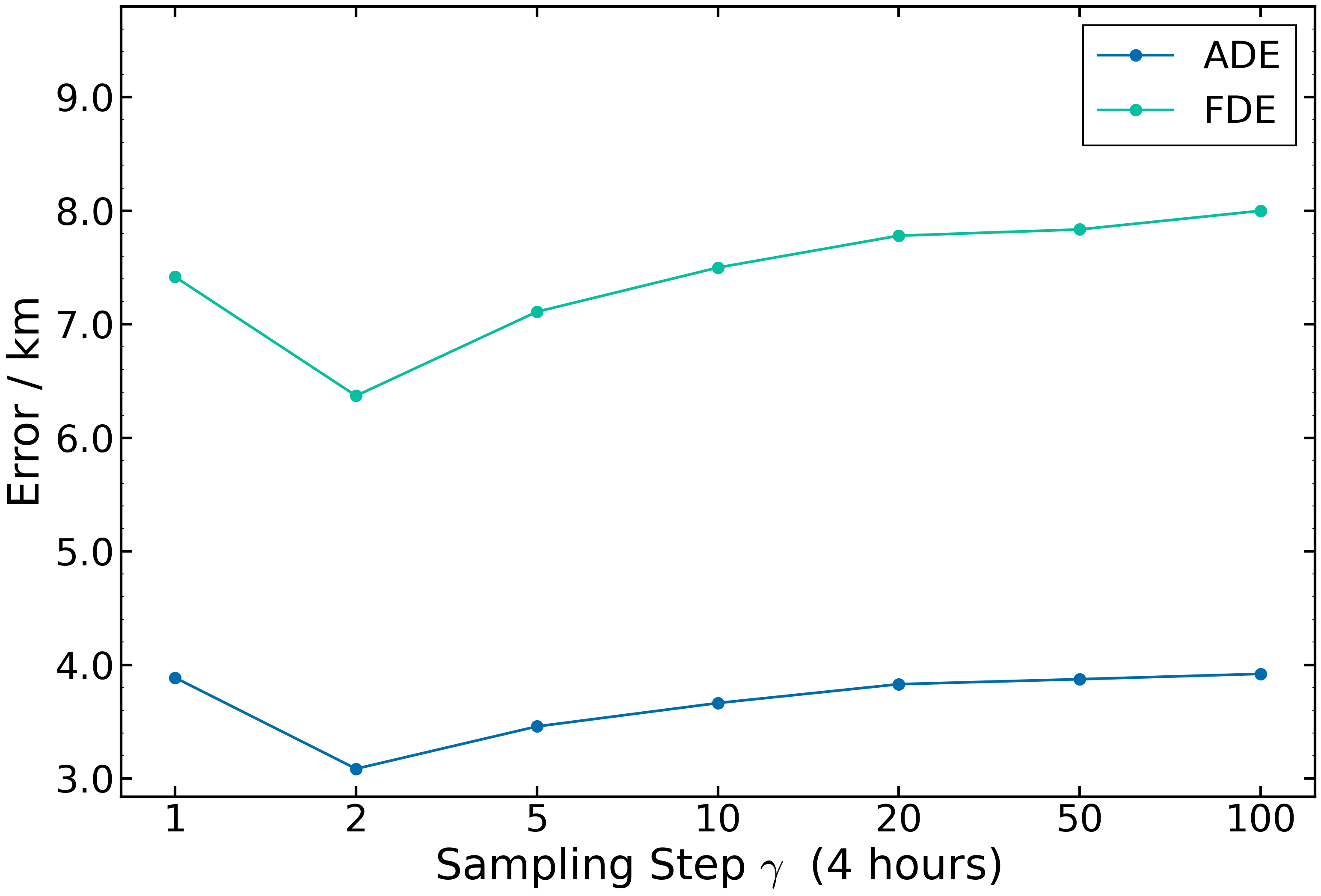} % 调整图片宽度
    }
    \caption{Prediction performance under different sampling steps $\gamma$ on the Danish Straits testing set over 4 hours. The trend of ADE is represented by blue lines, while the trend of FDE is represented by green lines.}
    \label{fig4}
\end{figure*}

Our pre-processing methodology is formulated as follows. Initially, we filter out invalid information, which included entries with missing values or MMSI values less than 9 digits. Subsequently, we eliminate implausible speed data, specifically, \(sog \geq 30\) and anchored vessel data with \(sog \leq 0.2\). Meanwhile, we address non-contiguous journeys with intervals exceeding 2 hours by dividing them into distinct groups of continuous journeys, each assigned a new MMSI. This facilitated the downsampling of AIS trajectories, utilizing a fixed-length sampling rate of \(\Delta=10\) minutes, and incorporating linear interpolation for missing or removed data between the minimum and maximum timestamps for each vessel. Moreover, we exclude journeys with a duration less than 36, and the Min-Max normalization method was applied to the filtered data. Consequently, we obtain continuous AIS data characterized by reduced signal acquisition noise, which was subsequently modeled as a directed scene graph \(G=(V, E)\) across timestamps.
\mbox{}\\
\textbf{Evaluation Metric} \quad In our evaluation, we utilize the Average Displacement Error (ADE) and Final Displacement Error (FDE) to assess the accuracy of our model. ADE calculates the average error between all the ground truth positions and the estimated positions in the trajectory, while FDE measures the displacement between the end points of ground truth and predicted trajectories. To enhance the precision of our displacement measurements and ensure relevance to real-world maritime scenarios, we implement the Haversine Formula in place of the traditional Euclidean distance. This adjustment accounts for the curvature of the Earth's surface, making it more suitable for geographical distance calculations. Considering the stochastic nature of our model, we adopt the Best-of-$N$ strategy \cite{bhattacharyya2018accurate} to compute the final ADE and FDE, with $N = 20$.
\mbox{}\\
\textbf{Baselines} \quad 
To validate the effectiveness of our proposed model, we compare our model with the following baselines: 

\begin{itemize}
    \item \textbf{Vanilla LSTM \cite{tang2022model}} utilizes RNN-based predictor utilizing LSTM networks to model trajectory points. Features are projected using input and output embedding layers with fully connected networks.
    
    \item \textbf{LSTM-Seq2Seq \cite{forti2020prediction}} extends traditional LSTM by incorporating an encoder-decoder architecture for trajectory prediction.

    \item \textbf{LSTM-Seq2Seq-Att \cite{capobianco2021deep}} integrates attention mechanisms to focus on specific parts of the input sequence during prediction.

    \item \textbf{TrAISformer \cite{nguyen2021traisformer}}: This is a Transformer-based architecture designed for trajectory prediction in an autoregressive manner. It incorporates speed over ground (SOG) and course over ground (COG) information and transforms regression problems into classification by constructing features as discrete high-dimensional vectors.

    \item \textbf{GANs \cite{gupta2018social}} utilizes generative adversarial networks to capture the future distribution, where both the generator and discriminator are based on LSTM networks.

    \item \textbf{CVAE \cite{han2023interaction}} exploits conditional variational auto-encoder to represent uncertainty for trajectory prediction, re-implemented based on the gated recurrent unit (GRU).
\end{itemize}
\mbox{}\\
\textbf{Implementation Details} \quad  We train our model using the Adam optimizer, with an initial learning rate of 0.0001 and a batch size of 256, following an exponential decay learning rate schedule. Our training assumes observation of $T=8$ timestamps (80 minutes) of a trajectory to predict subsequent future timestamps. The model's encoder features two LSTMs with 128 hidden dimensions and a three-layer CNN with 32d-64d-128d channels, producing a 64-dimensional embedding. The decoder incorporates a three-layer Transformer with 512 hidden dimensions and 4 attention heads. For the diffusion process, we set the total step size $K$ to 100 and the sampling step size $\gamma$ to 5. All experiments are conducted on an NVIDIA A100 Tensor Core GPU.

\begin{figure*}[t]
    \centering
    \begin{minipage}[b]{0.144\textwidth}
        \centering
        \hspace{-0.3mm} % 减小填充
        \includegraphics[width=\linewidth]{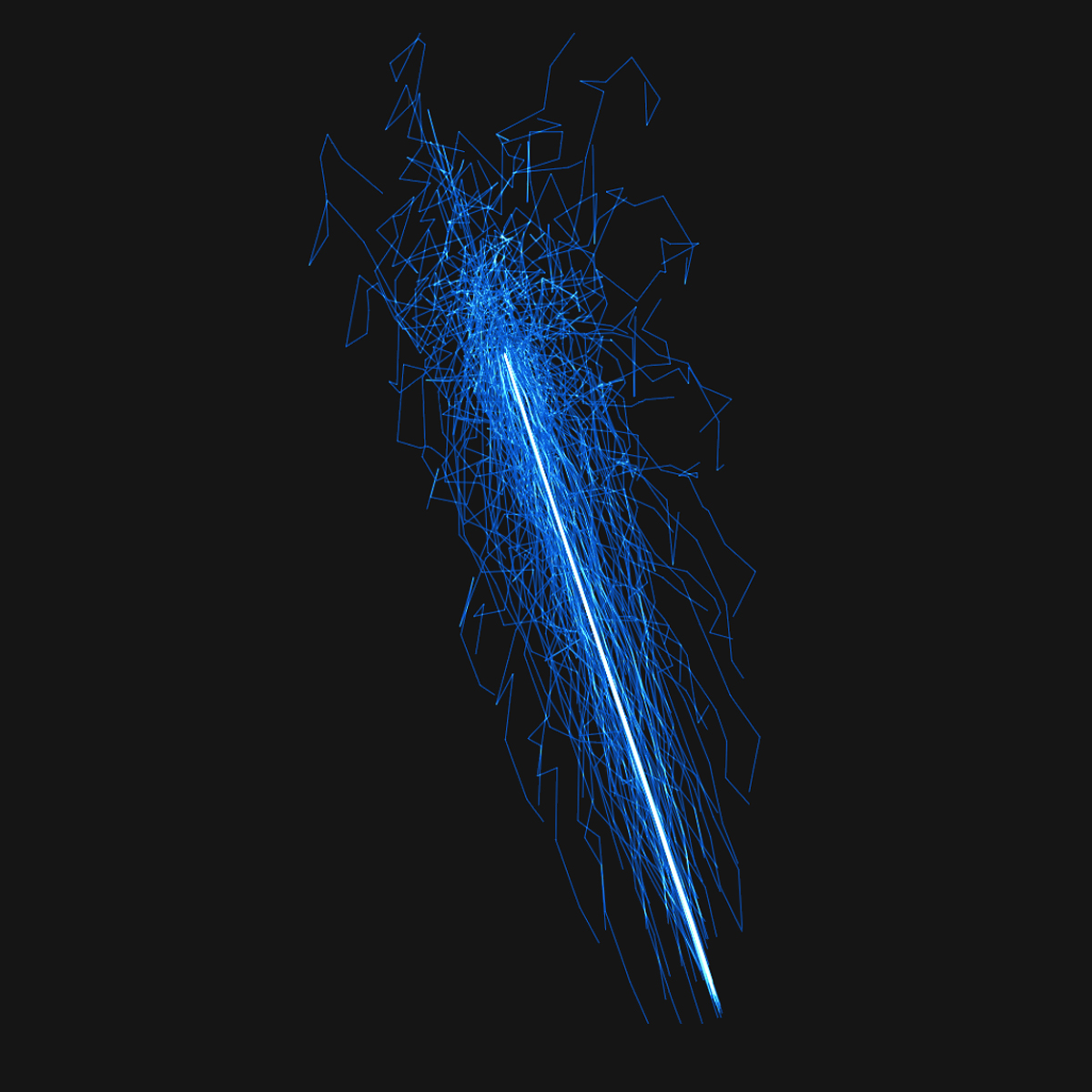}
        \subfigure{Overall}
        \label{fig:subfig1}
    \end{minipage}
    \hspace{-2.55mm}
    \begin{minipage}[b]{0.144\textwidth}
        \centering
        \includegraphics[width=\linewidth]{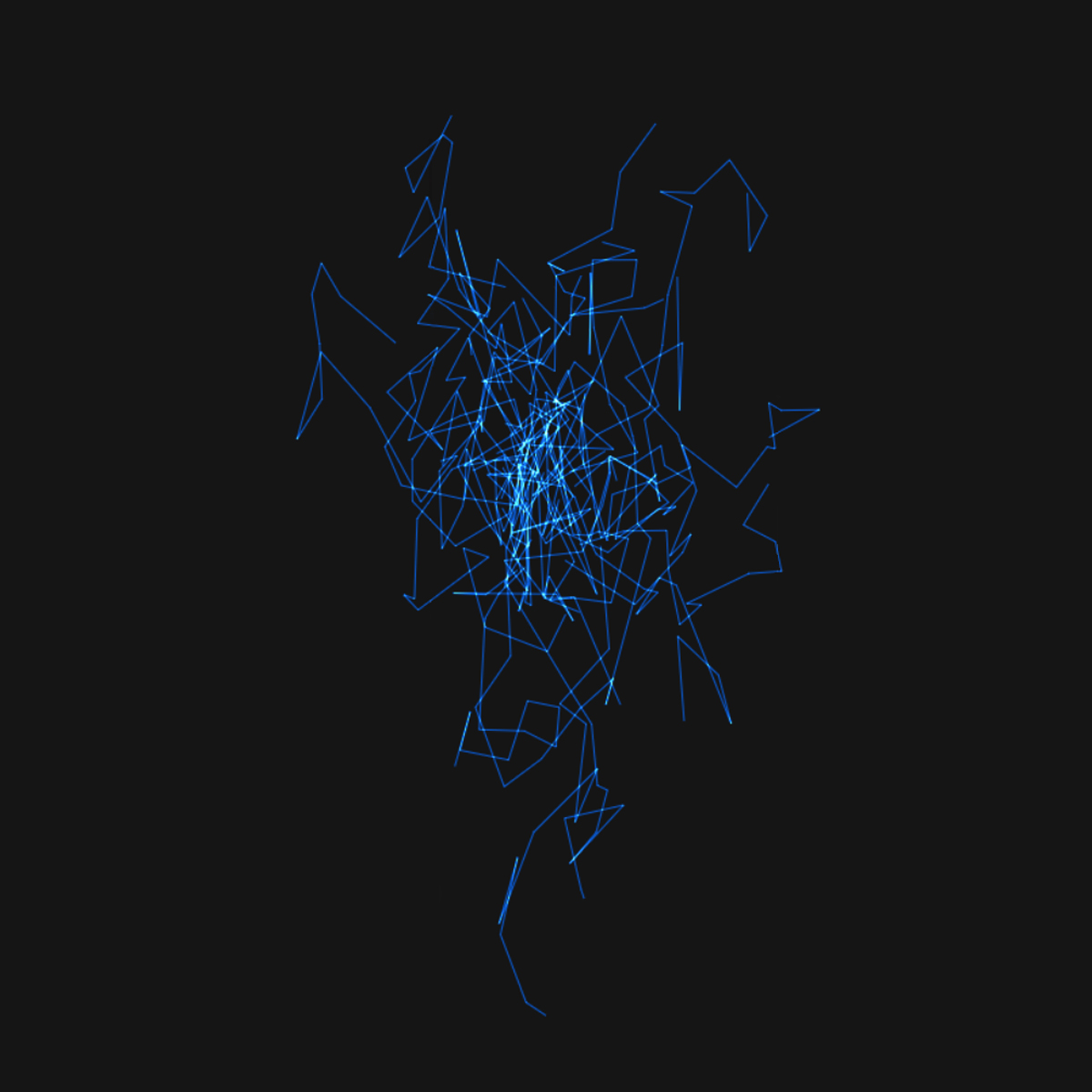}
        \subfigure{\textit{k} = 100}
        \label{fig:subfig2}
    \end{minipage}
    \hspace{-2.55mm}
    \begin{minipage}[b]{0.144\textwidth}
        \centering
        \includegraphics[width=\linewidth]{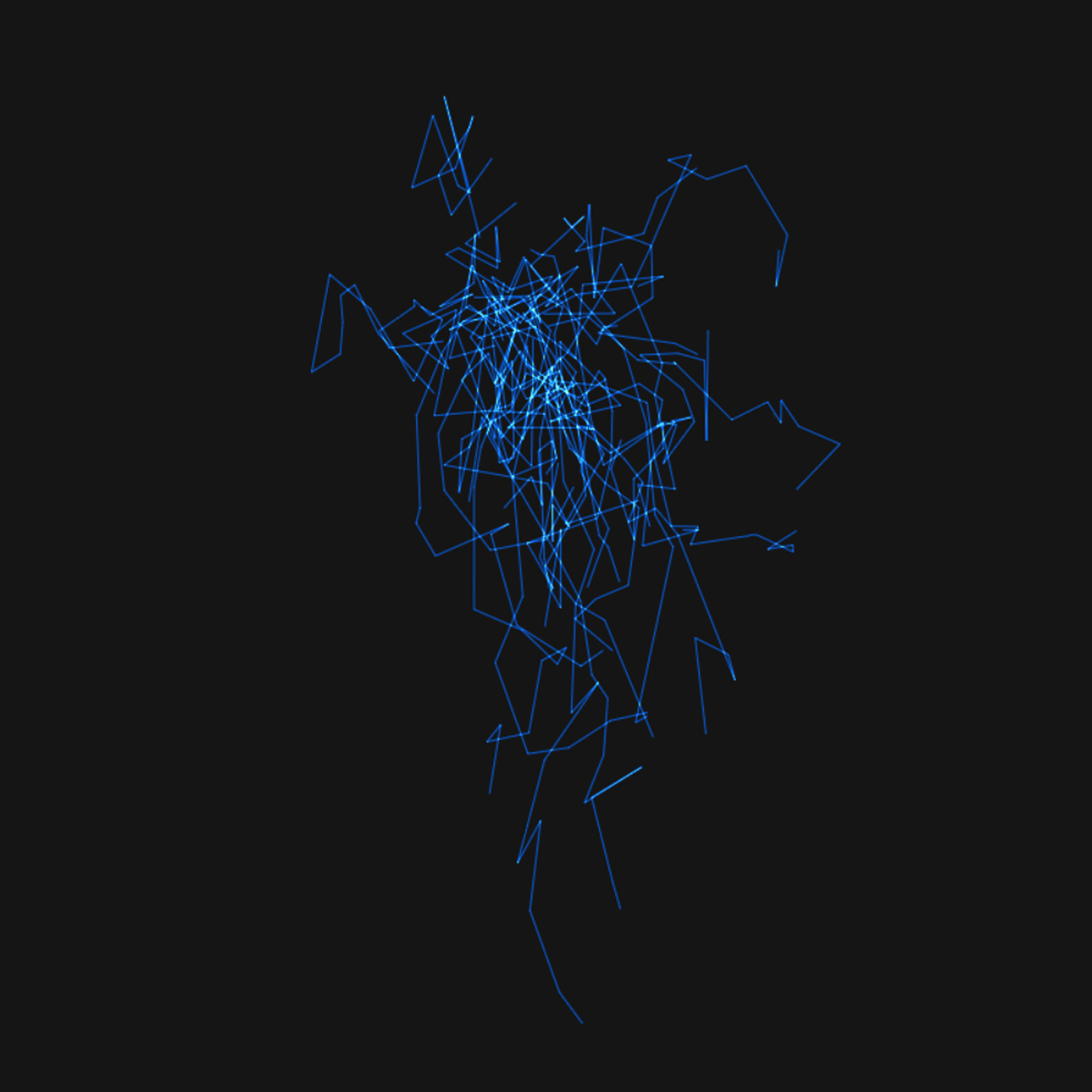}
        \subfigure{\textit{k} = 80}
        \label{fig:subfig3}
    \end{minipage}
    \hspace{-2.55mm}
    \begin{minipage}[b]{0.144\textwidth}
        \centering
        \includegraphics[width=\linewidth]{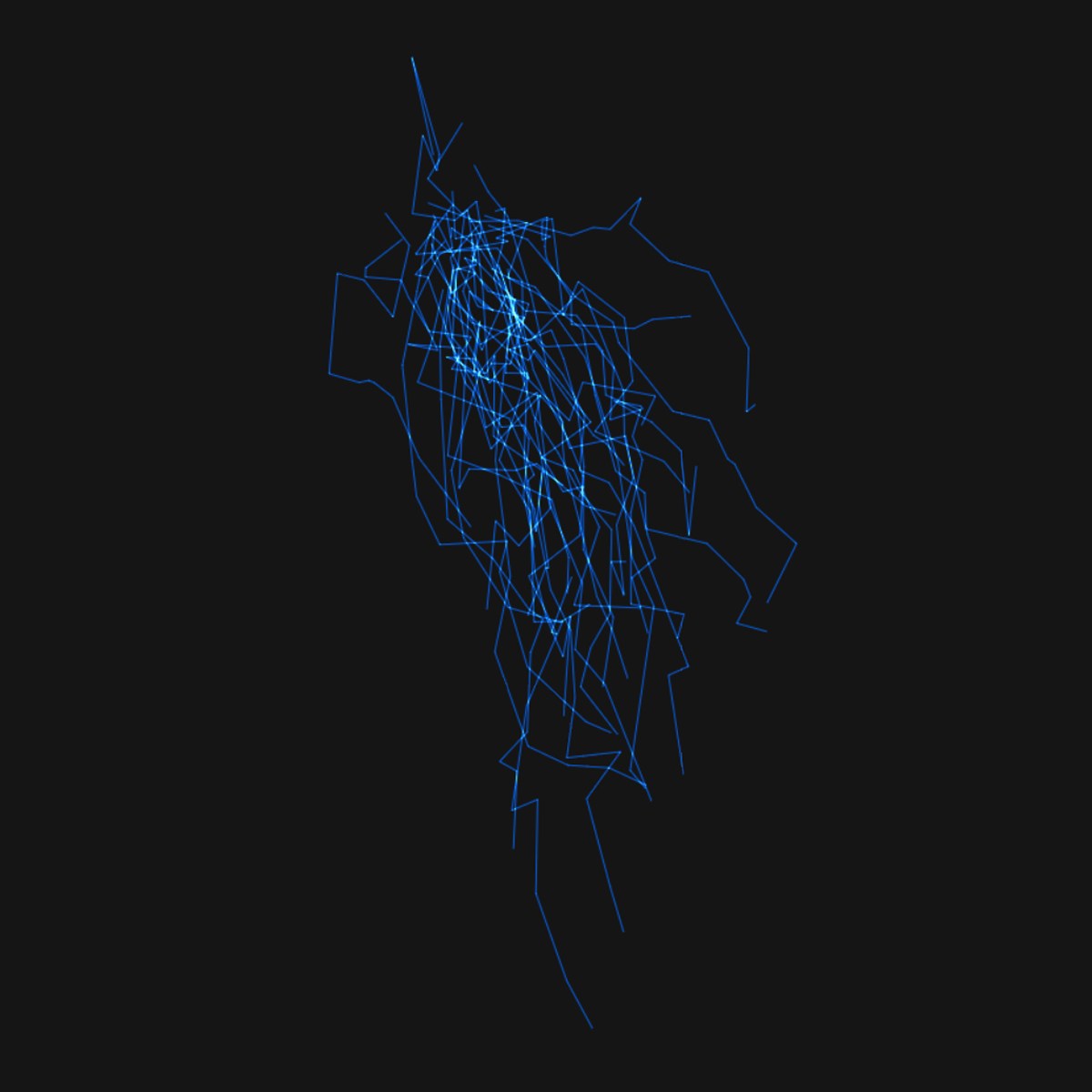}
        \subfigure{\textit{k} = 60}
        \label{fig:subfig4}
    \end{minipage}
    \hspace{-2.55mm}
    \begin{minipage}[b]{0.144\textwidth}
        \centering
        \includegraphics[width=\linewidth]{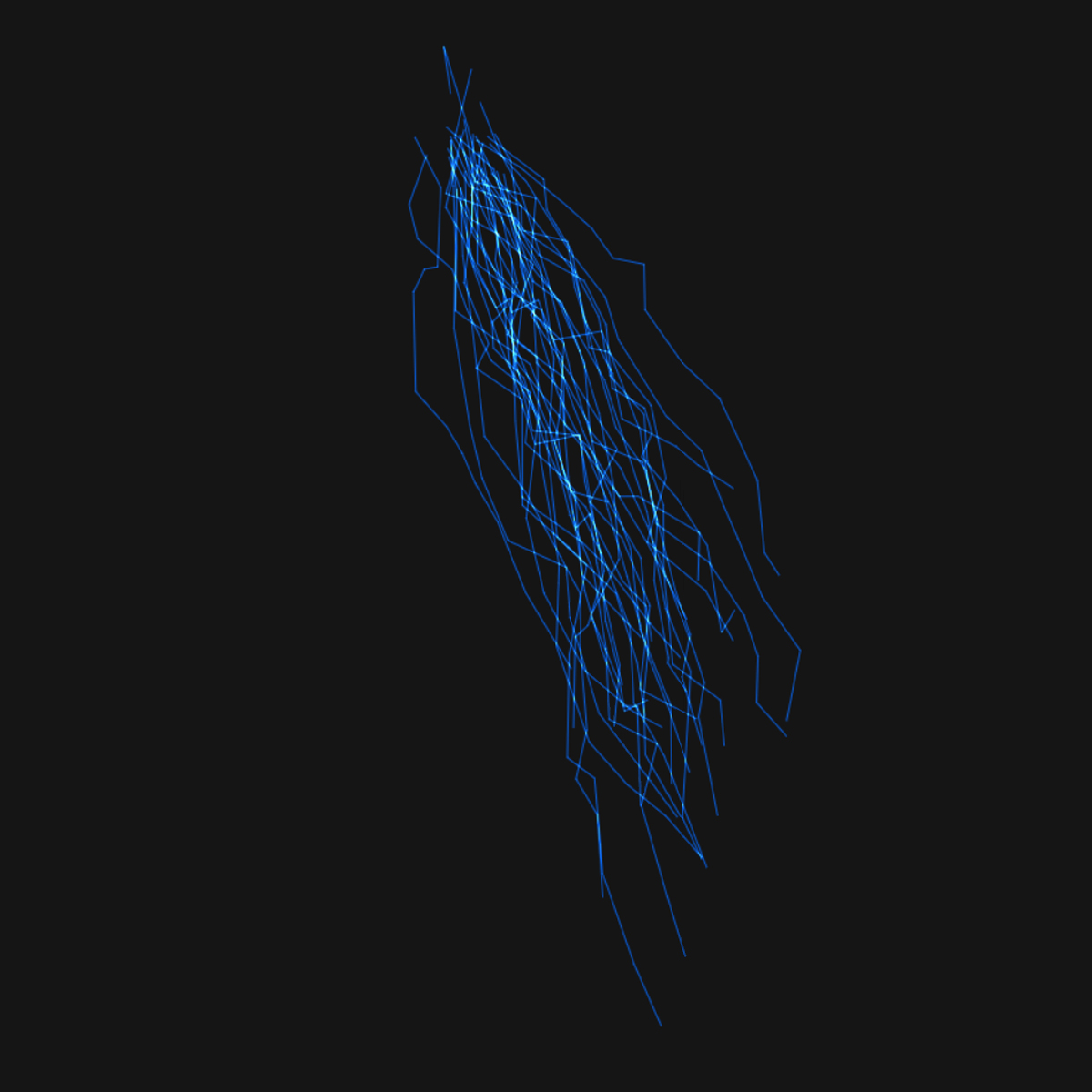}
        \subfigure{\textit{k} = 40}
        \label{fig:subfig5}
    \end{minipage}
    \hspace{-2.55mm}
    \begin{minipage}[b]{0.144\textwidth}
        \centering
        \includegraphics[width=\linewidth]{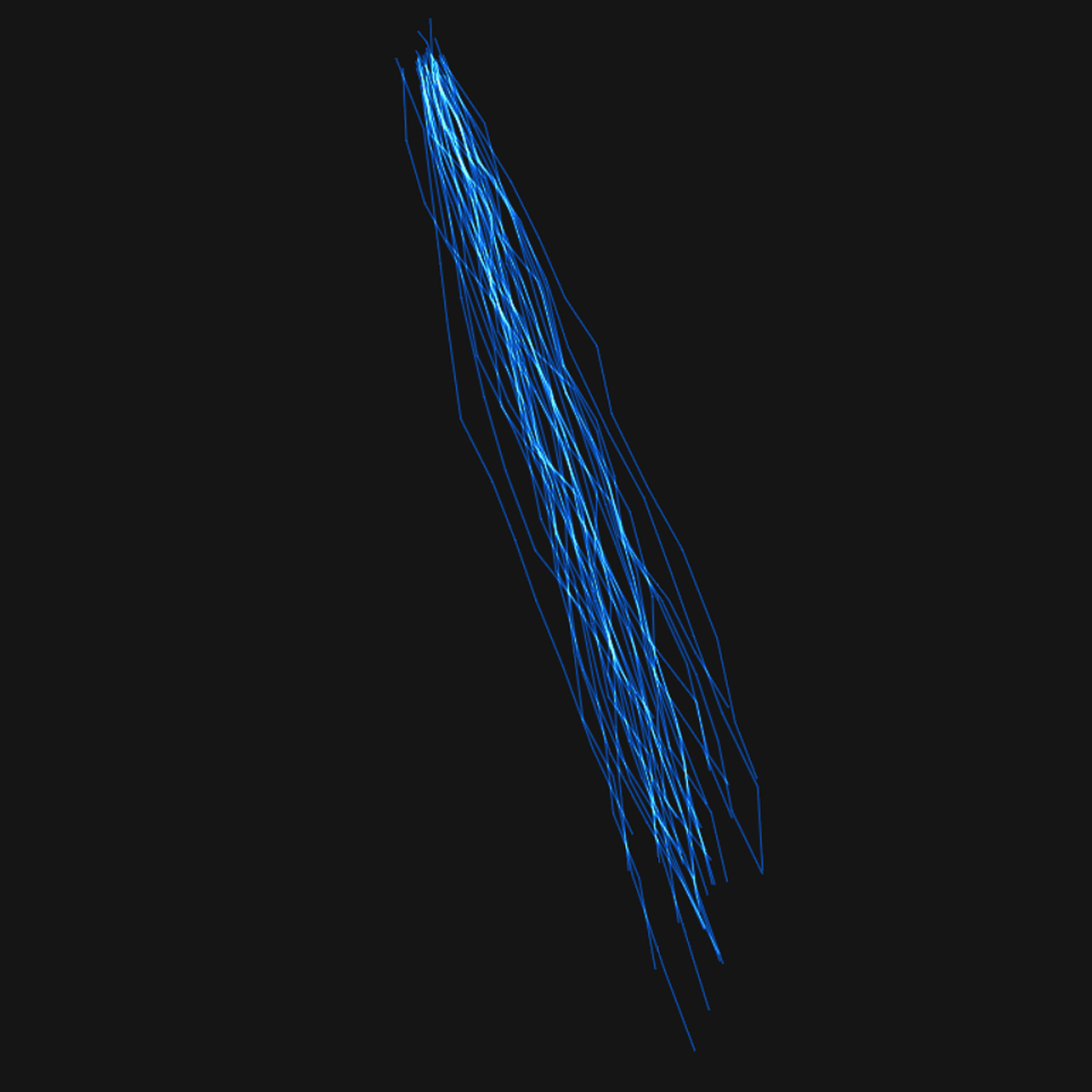}
        \subfigure{\textit{k} = 20}
        \label{fig:subfig6}
    \end{minipage}
    \hspace{-2.55mm}
    \begin{minipage}[b]{0.144\textwidth}
        \centering
        \includegraphics[width=\linewidth]{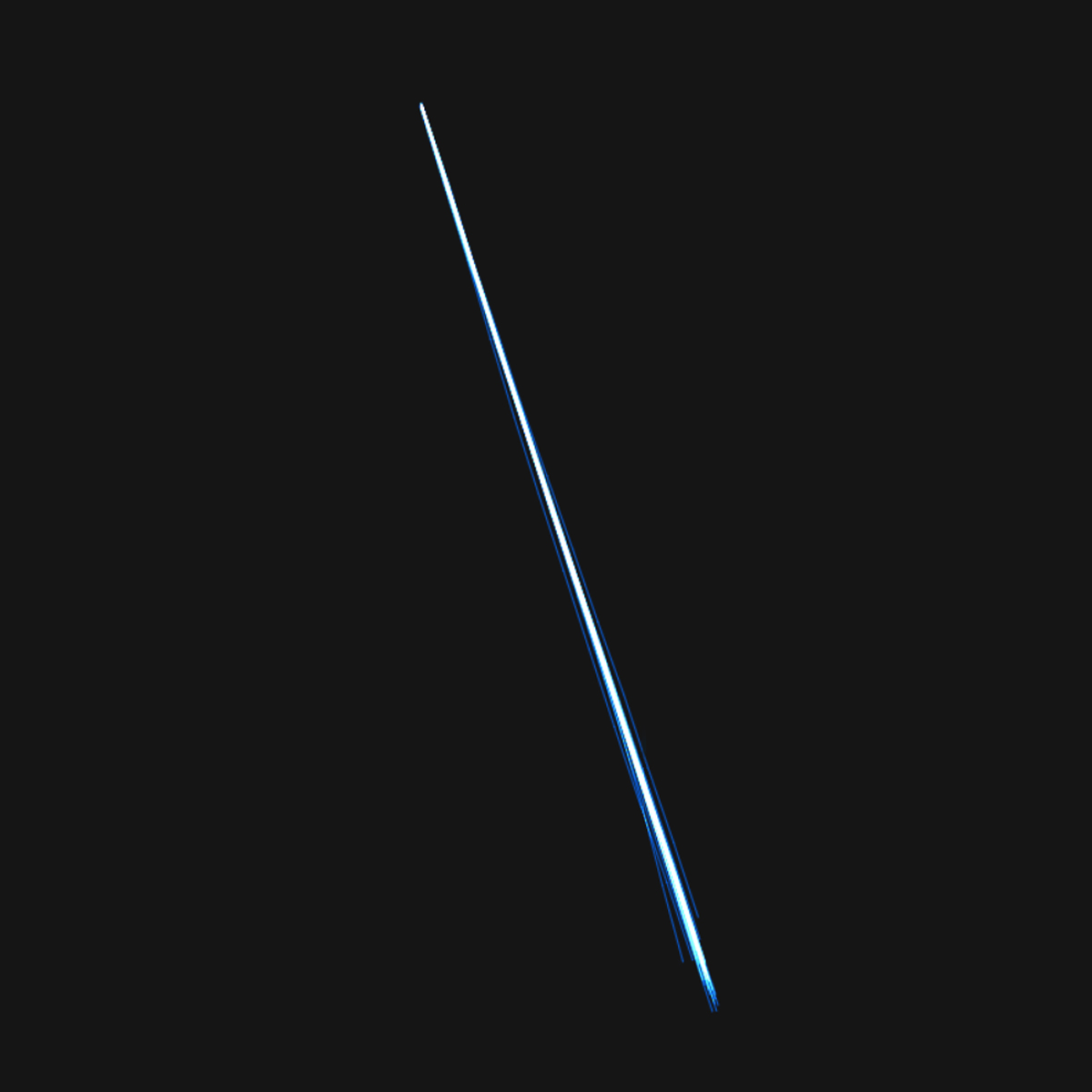}
        \subfigure{\textit{k} = 0}
        \label{fig:subfig7}
    \end{minipage}
    \caption{Visualization of denoised trajectories at each reverse diffusion step $k$. Our model progressively removes uncertainty and derives deterministic future trajectories through the learned reverse diffusion process. The trajectories start from a Gaussian noise distribution ($k$ = 100) and undergo a non-Markov denoising process to arrive at the predicted trajectories ($k$ = 0), successfully simulating the process from diversity to determinism.}
    \label{fig6}
\end{figure*}

\begin{figure}[b]
    \centering
    \includegraphics[scale=0.35]{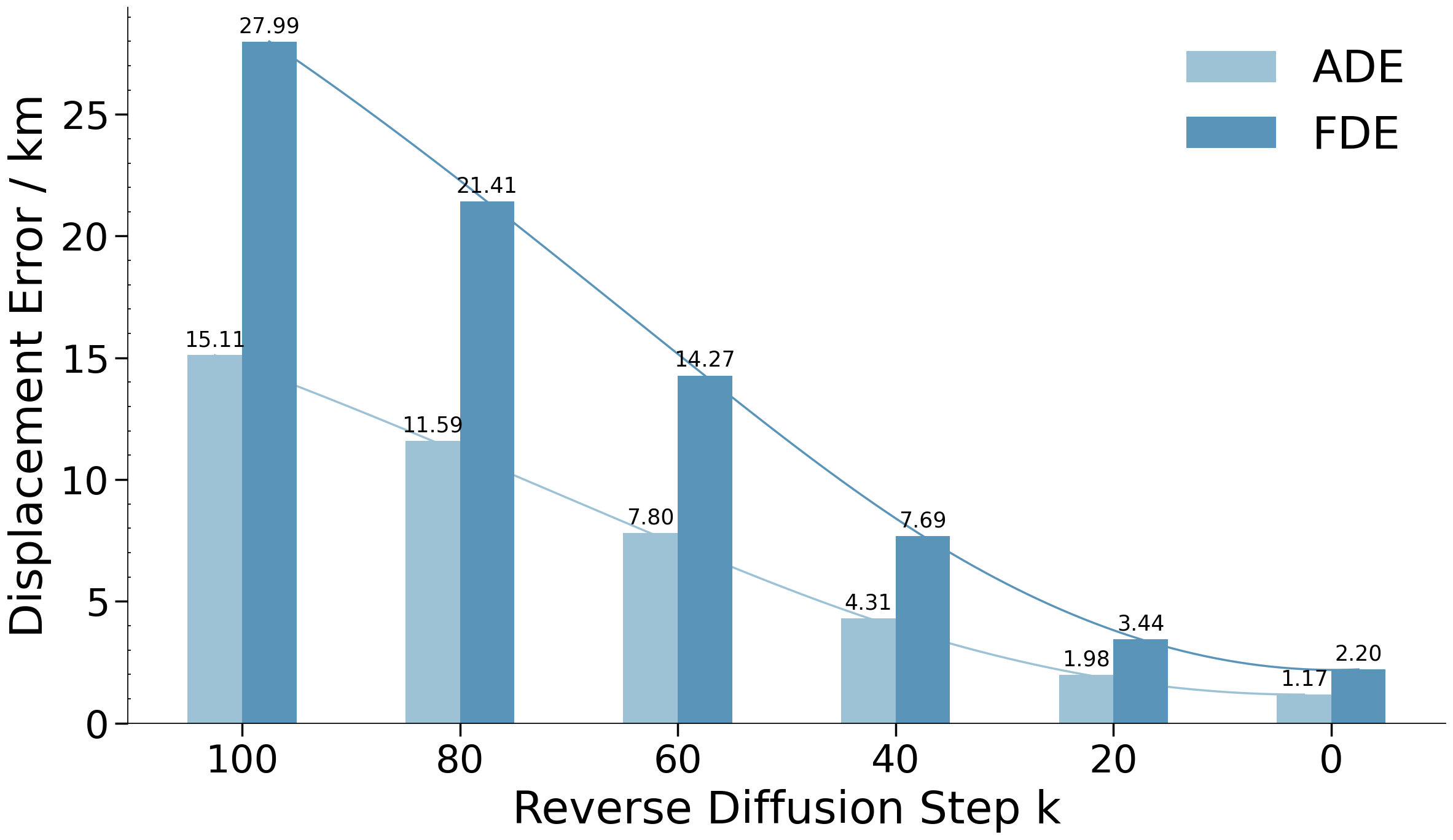}
    \caption{Trajectory forecast errors for different reverse diffusion steps at the 2-hour mark on the Danish Straits dataset. Starting from a Gaussian distribution ($k$ = 100) to the denoised trajectories ($k$ = 0).}
    \label{fig5}
\end{figure}

\subsection{Overall Performance}
In our study, we benchmark DiffuTraj against various baseline models using ADE and FDE. The results, detailing trajectory forecast errors for lead times from 0.5 to 4 hours (3 to 24 timestamps), are summarized in Table \ref{tab1}. Our key observations include:
\subsubsection{Comparison of Deterministic Predictive Models}
A comparison among LSTM-based models shows that the LSTM-Seq2Seq model significantly outperforms the Vanilla LSTM model, especially at longer forecast horizons (with a 37.8\% / 32.9\% reduction in ADE and a 40.4\% / 35.5\% reduction in FDE at the 4-hour mark in two datasets). In contrast, the performance difference between the LSTM-Seq2Seq and LSTM-Seq2Seq-Att models is minimal, with peak reductions of only 4.6\% / 3.8\% and 3.8\% / 4.2\% in ADE and FDE, respectively. This indicates the substantial performance gains achievable with Seq2Seq architectures over standard LSTM models, whereas the addition of an attention mechanism offers only marginal further improvement. Moreover, while the overall performance of other LSTM-based models declines in most cases when handling the Baltic Sea dataset, the Vanilla LSTM shows greater improvements with average reductions in ADE and FDE of 6.5\% and 12.8\%, respectively.
Previously leading in vessel trajectory prediction, TrAISformer, in our deterministic implementation, shows less proficiency in short-term predictions but gradually stabilizes in long-term forecasts. More specifically, in the Danish Straits dataset, TrAISformer performs worse than LSTM-Sqe2Seq-Att model for a time horizon up to 3 hours, but better for longer term prediction (up to 4 hours). This is likely due to its use of discrete high-dimensional embeddings, which effectively capture spatial dependencies for more stable long-term predictions. Regarding the Baltic Sea dataset, TrAISformer consistently outperforms the Vanilla LSTM in most cases, but fails to exceed the LSTM-Seq2Seq series models. This is due to the deterministic selection of the highest classification probability instead of random sampling.
\subsubsection{Comparison of Deep Generative Models} 
In the realm of deep generative models, CVAE  (proposed in \cite{han2023interaction}) consistently surpasses GANs (proposed in \cite{gupta2018social}) in performance for a time horizon up to 4 hours. Specifically, in terms of the best reduction and mean reduction of ADE and FDE in the Danish Straits dataset is 22.0\% / 17.6\% and 39.0\% / 30.6\%, respectively. Moreover, in the Baltic Sea dataset, the results improve to 31.2\% / 27.2\% and 47.4\% / 42.1\%, respectively. From a longitudinal comparison, the performance of GANs deteriorates comprehensively on the Baltic dataset, with average increases of 20.8\% in ADE and 20.1\% in FDE. This indicates GANs exhibits decreased performance in scenarios with shorter total duration and denser vessel counts. In contrast, the performance decline of CVAE is less pronounced, with average increases of 6.7\% in ADE and 0.7\% in FDE. Despite GANs exhibiting rapid error increases with heightened trajectory uncertainty, it outperforms deterministic models on average. Particularly, compared to the optimal deterministic prediction model, GANs provides higher accuracy, with average reductions of 13.5\% / 0.9\% in ADE and 16.5\% / 0.6\% in FDE across two datasets. This finding suggests that generative approaches, by modeling trajectory multimodality, are better suited to encapsulate vessel motion complexities and uncertainties. Moreover, limitations in autoregressive decoding restrict the capacity of deterministic models.
\subsubsection{Comparison of DiffuTraj with Existing Methods}
Our DiffuTraj model consistently outperforms other methods across both ADE and FDE metrics at varying timestamps in different datasets. Compared to the leading deterministic methods, DiffuTraj achieves significant improvements, reducing ADE and FDE by an average of \textbf{58.0\% / 55.7\%} and \textbf{61.1\% / 57.8\%}, and a maximum reduction of 75.6\% / 71.1\% and 82.2\% / 77.1\%, respectively. Against deep generative models, the reductions are \textbf{41.6\% / 38.6\%} in ADE and \textbf{34.1\% / 27.6\%} in FDE on average, with peaks at 60.0\% / 57.6\% and 57.9\% / 54.1\%, respectively. These enhancements are attributed to two key factors: (1) The explicit modeling of multi-modality in future motion states through a diffusion-based approach, which incrementally refines indeterminate predictions into more accurate trajectories. (2) The integration of scene context with spatial interactions among vessels, leading to more spatially sensitive and physically plausible trajectory predictions, where the condition exhibits an enhanced guiding effect as the forecast window increases.

\begin{figure*}
\centering
\begin{minipage}[b]{0.01\linewidth}
\rotatebox{90}{\scriptsize{Ours}}\\ \\  \\ \\ \\ \vspace{7pt}
\rotatebox{90}{\scriptsize{TrAISformer}} \\ \\ \\  \\ \\
\vspace{1pt}
\rotatebox{90}{\scriptsize{CVAE}}\\ \\ \\ \\ \vspace{11pt}
\rotatebox{90}{\scriptsize{LSTM-Seq2Seq}}\\ \vspace{11pt}
\end{minipage}
\subfigure[\scriptsize{Island Restriction}]{
\begin{minipage}[b]{0.16\linewidth}
\includegraphics[width=1.05\linewidth]{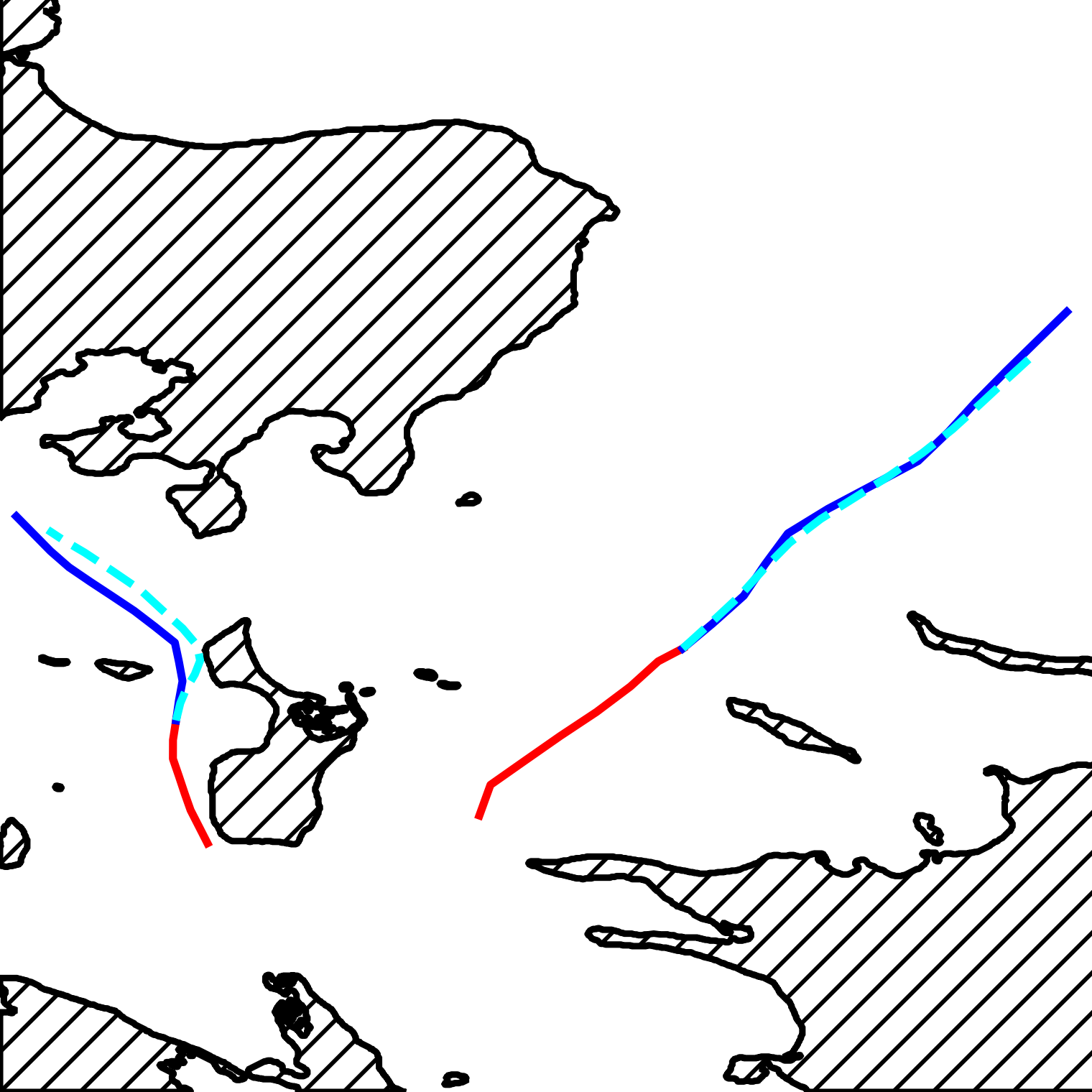}\vspace{4pt}
\includegraphics[width=1.05\linewidth]{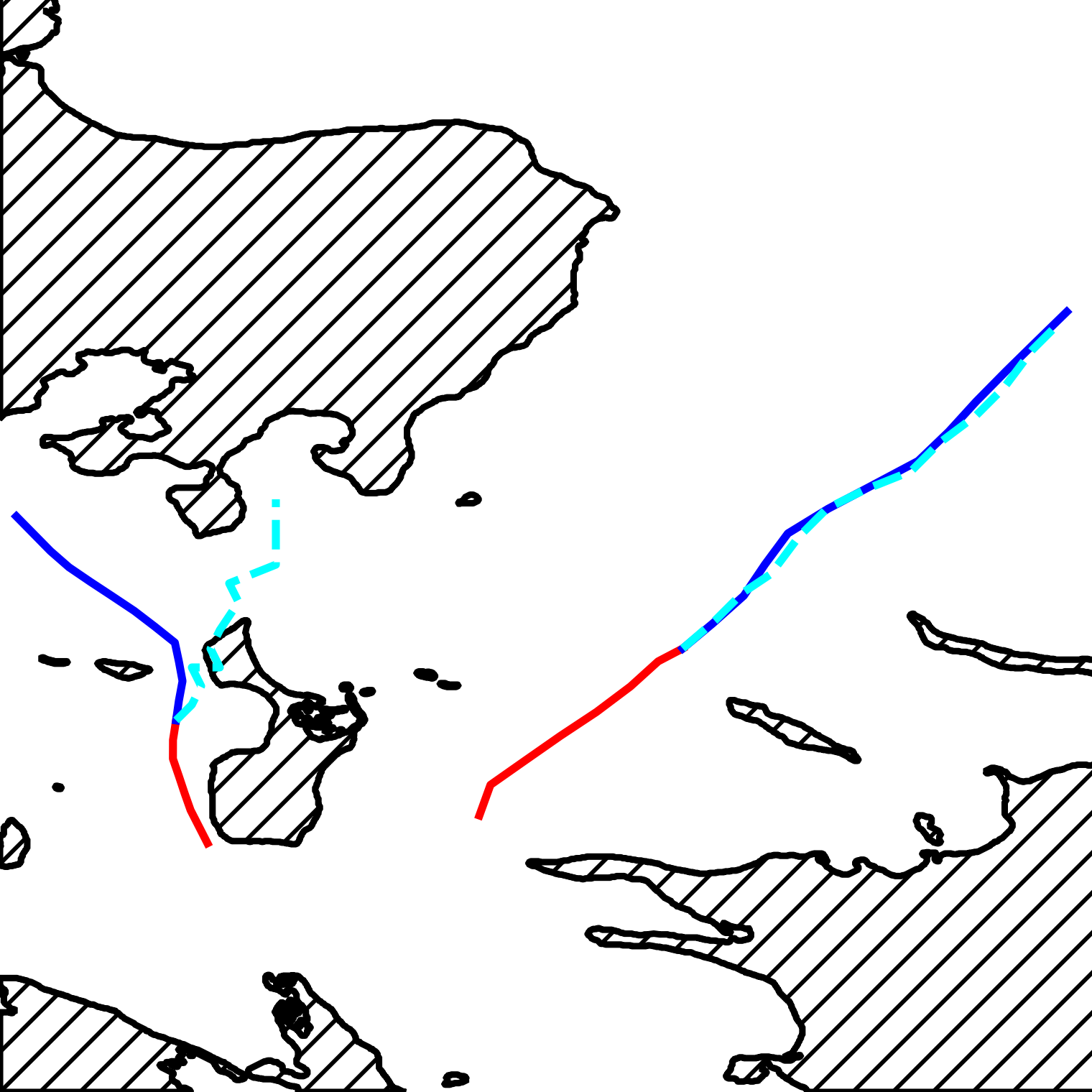}\vspace{4pt}
\includegraphics[width=1.05\linewidth]{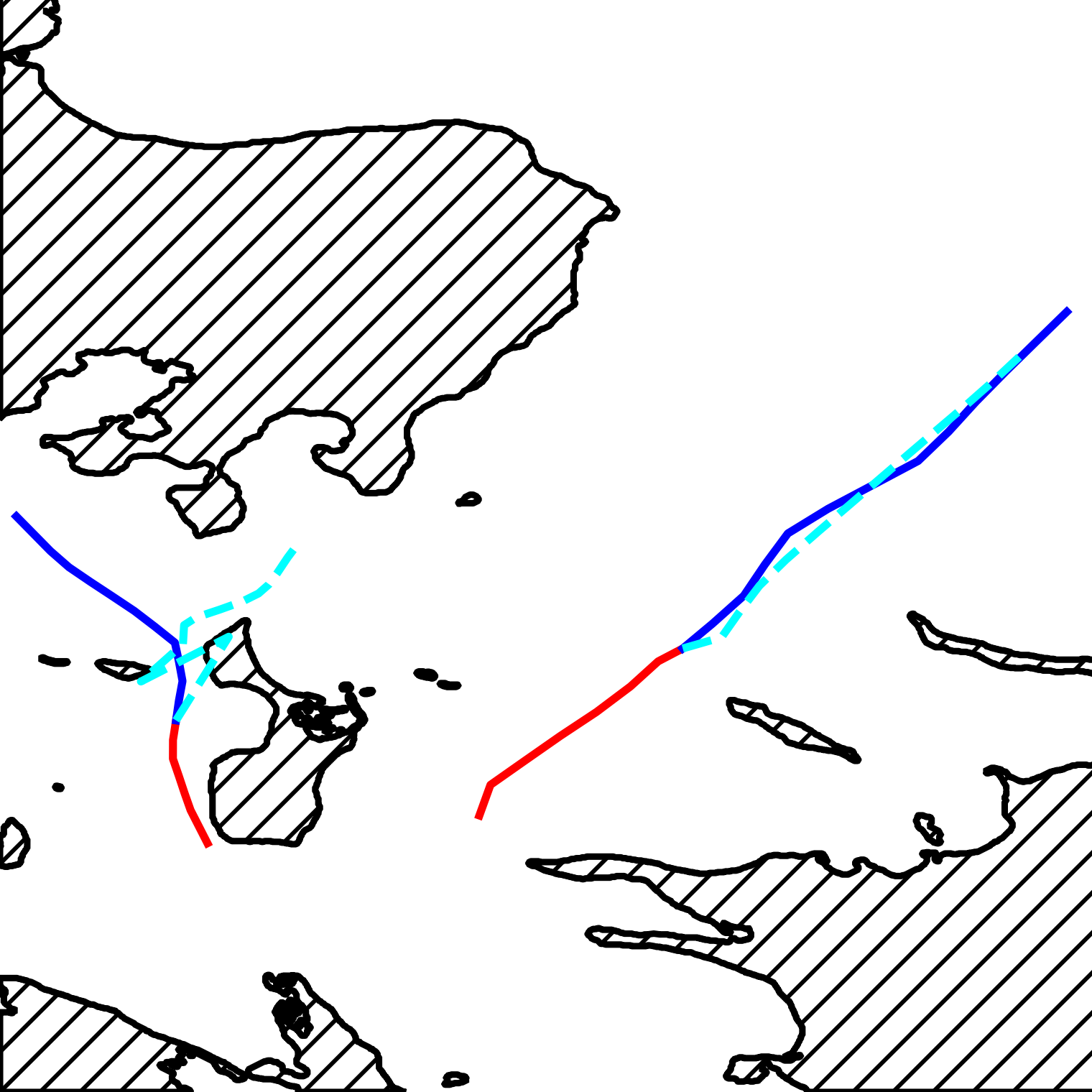}\vspace{4pt}
\includegraphics[width=1.05\linewidth]{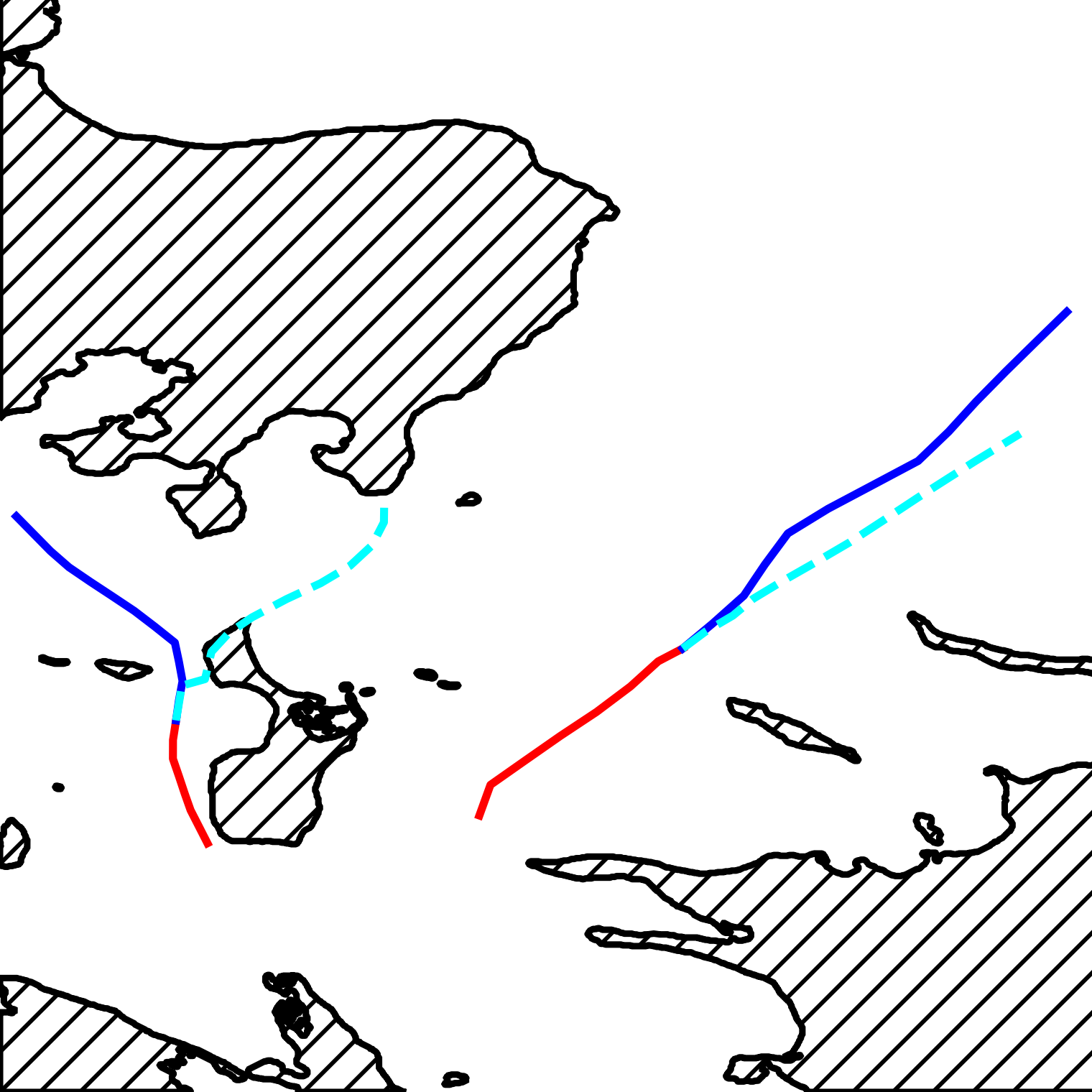}
\end{minipage}}
\hspace{8pt}
\subfigure[\scriptsize{Narrow Waterway}]{
\begin{minipage}[b]{0.16\linewidth}
\includegraphics[width=1.05\linewidth]{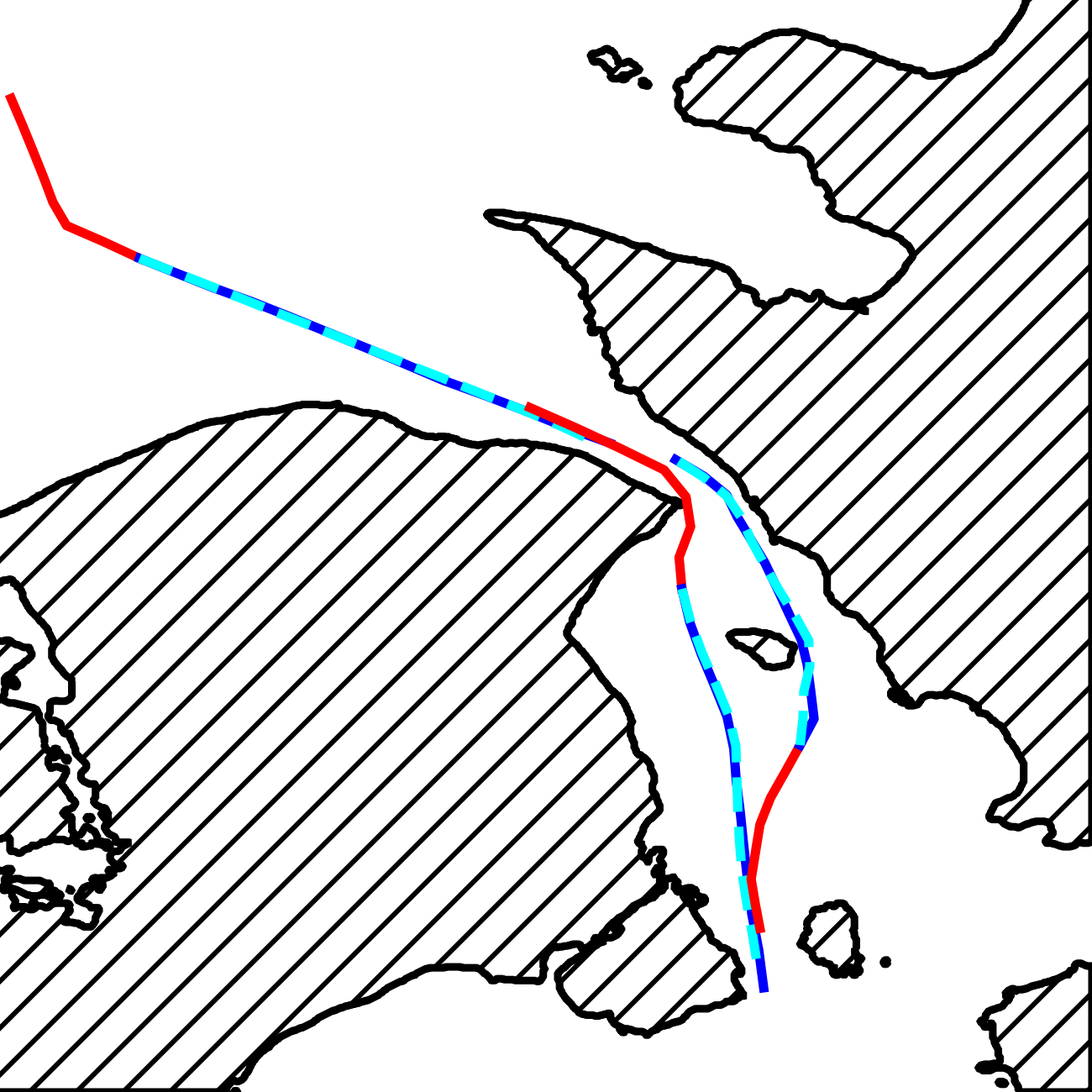}\vspace{4pt}
\includegraphics[width=1.05\linewidth]{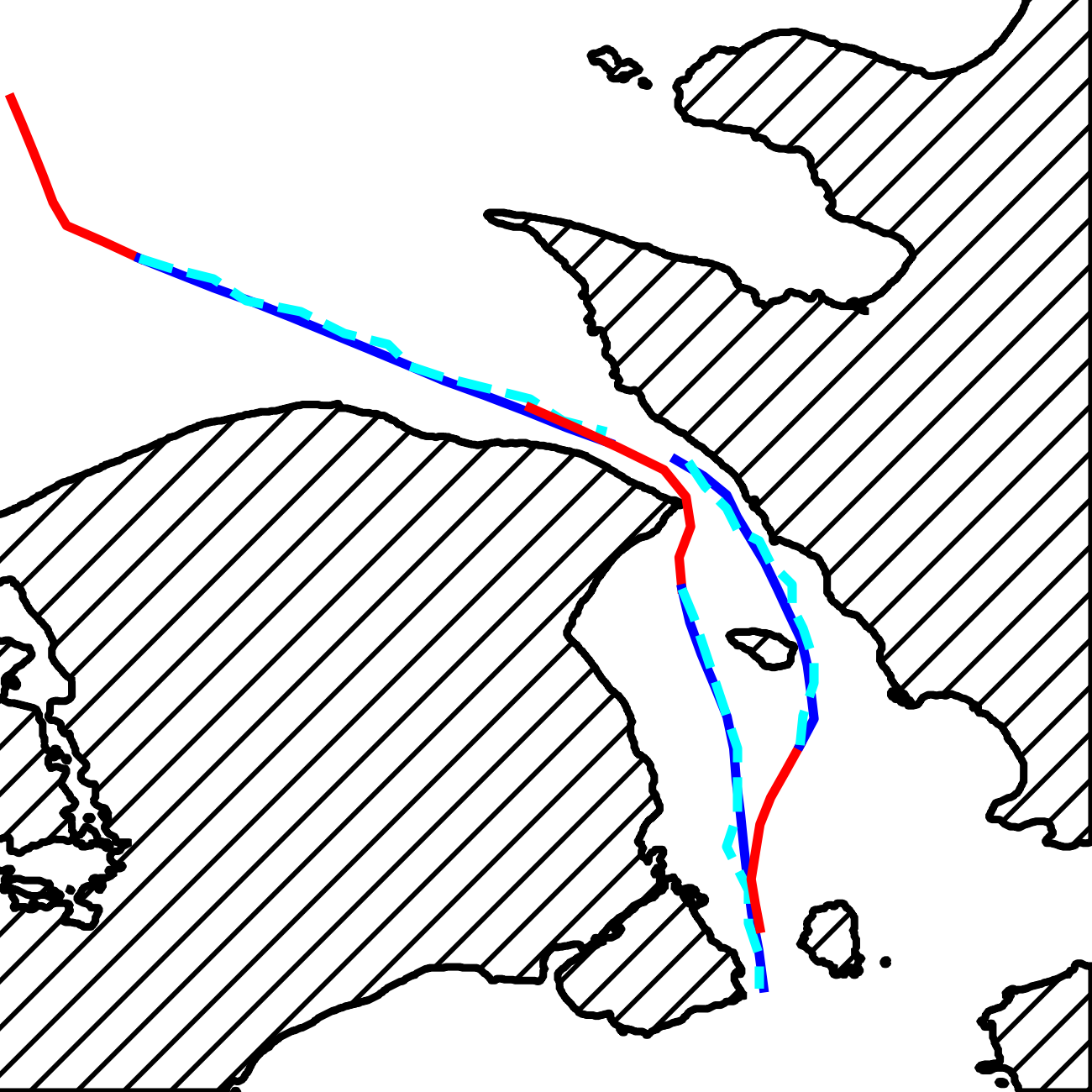}\vspace{4pt}
\includegraphics[width=1.05\linewidth]{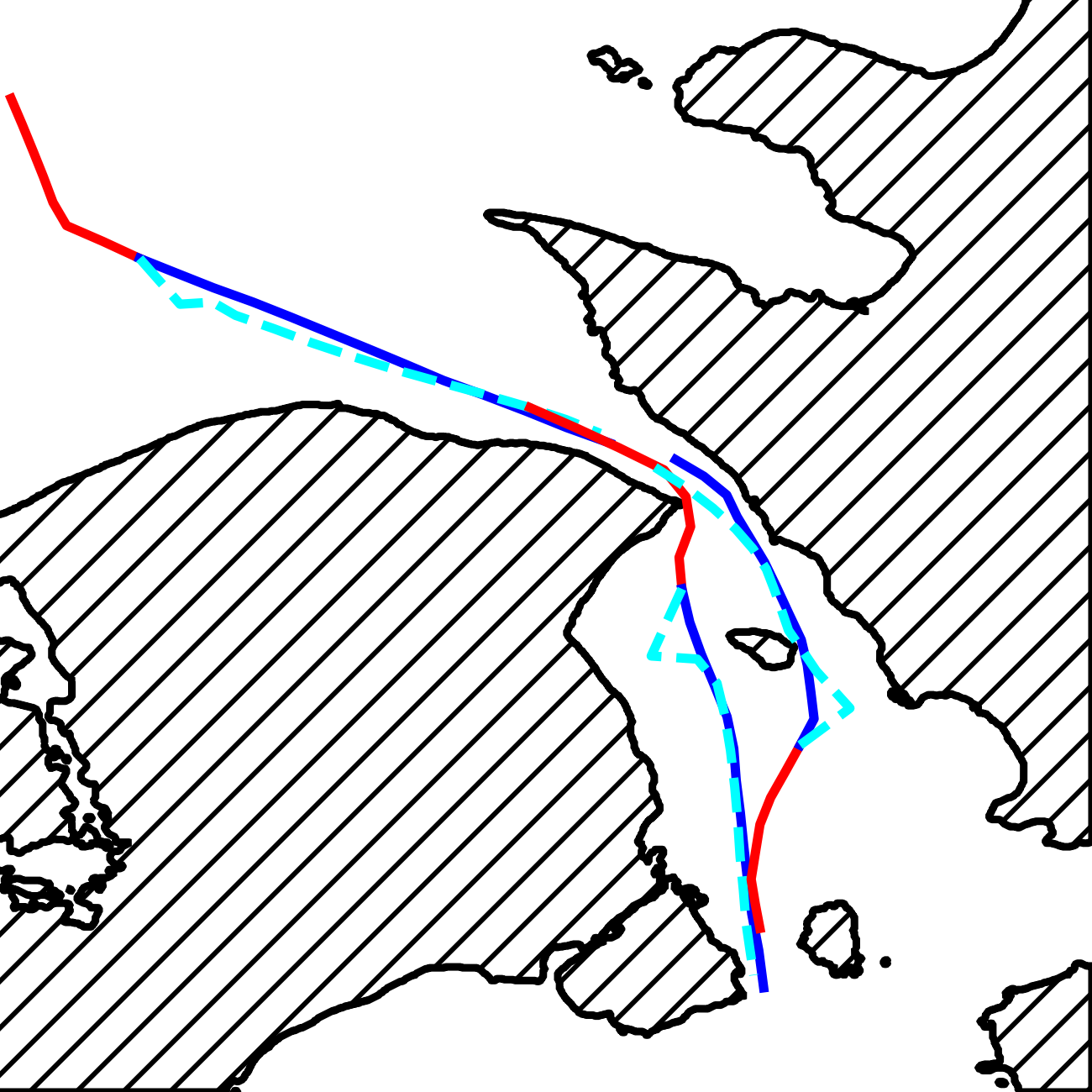}\vspace{4pt}
\includegraphics[width=1.05\linewidth]{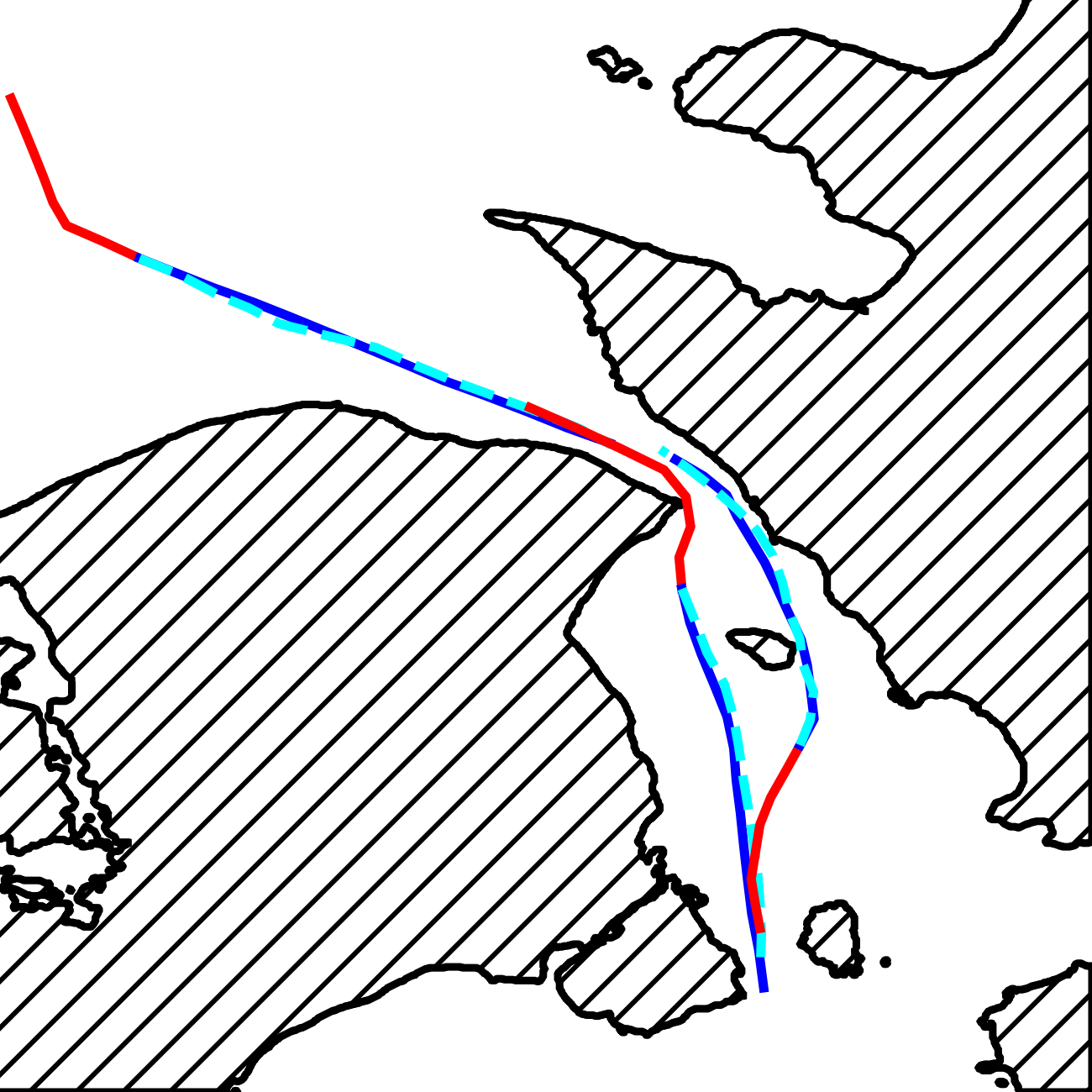}
\end{minipage}}
\hspace{8pt}
\subfigure[\scriptsize{Multi-Vessel Encounter}]{
\begin{minipage}[b]{0.16\linewidth}
\includegraphics[width=1.05\linewidth]{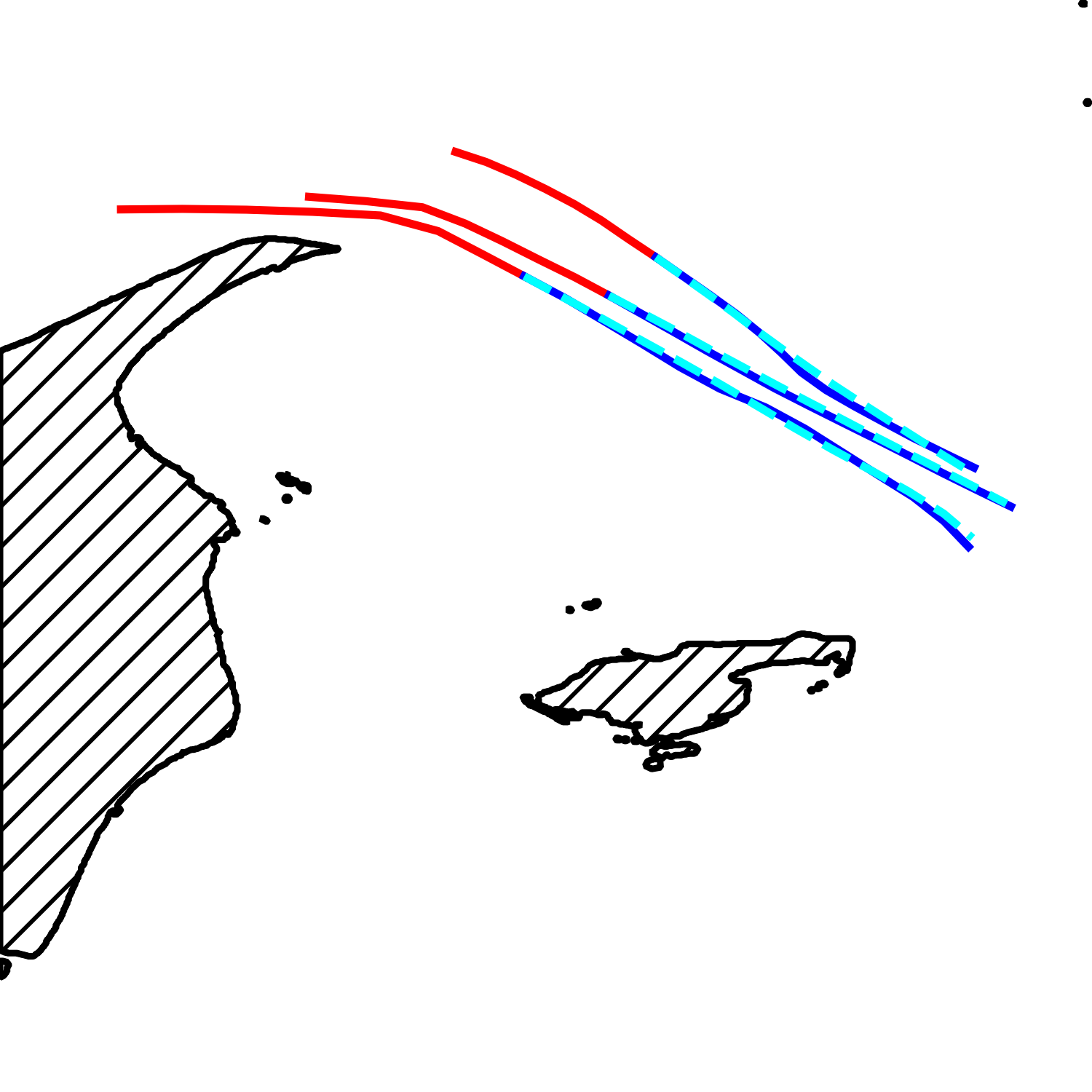}\vspace{4pt}
\includegraphics[width=1.05\linewidth]{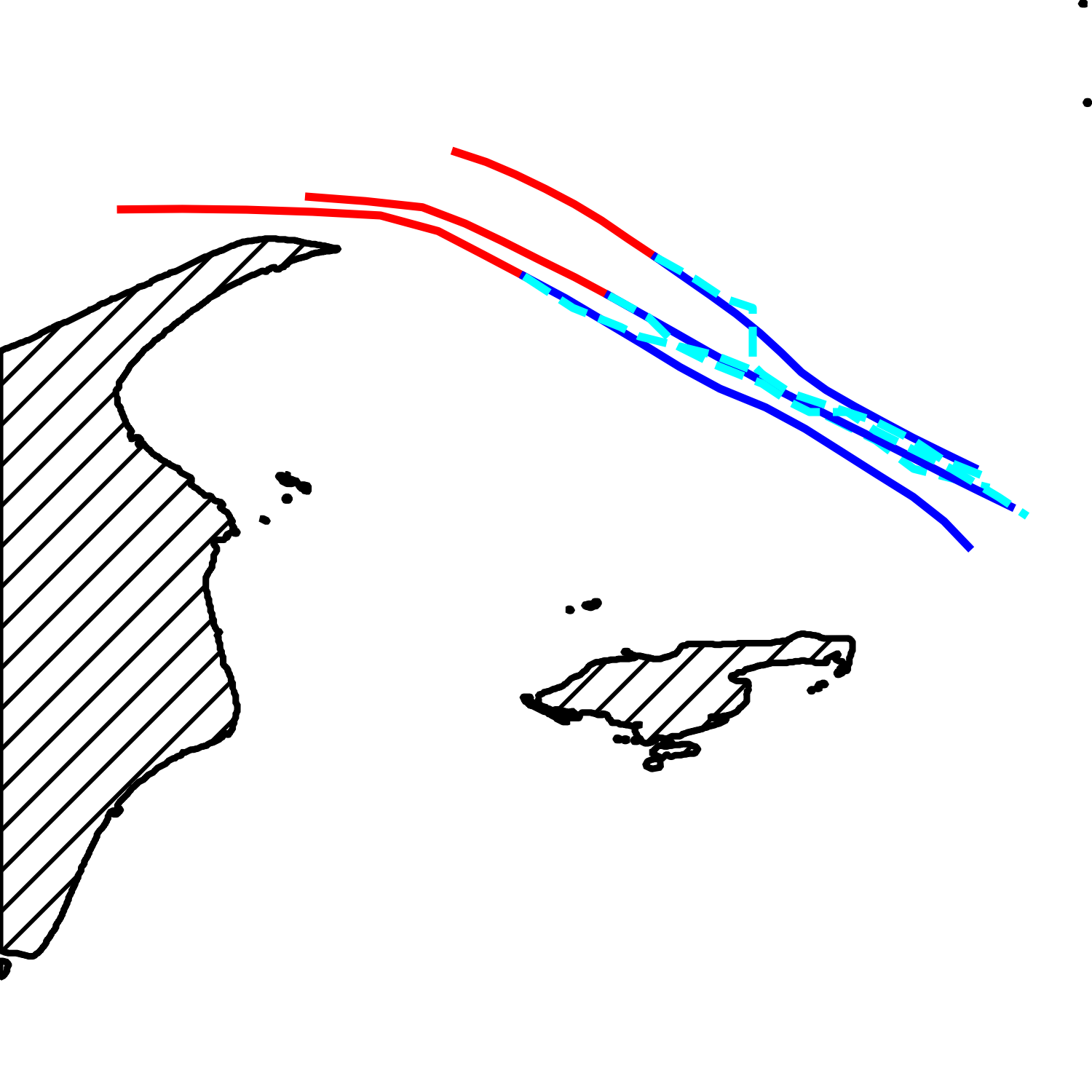}\vspace{4pt}
\includegraphics[width=1.05\linewidth]{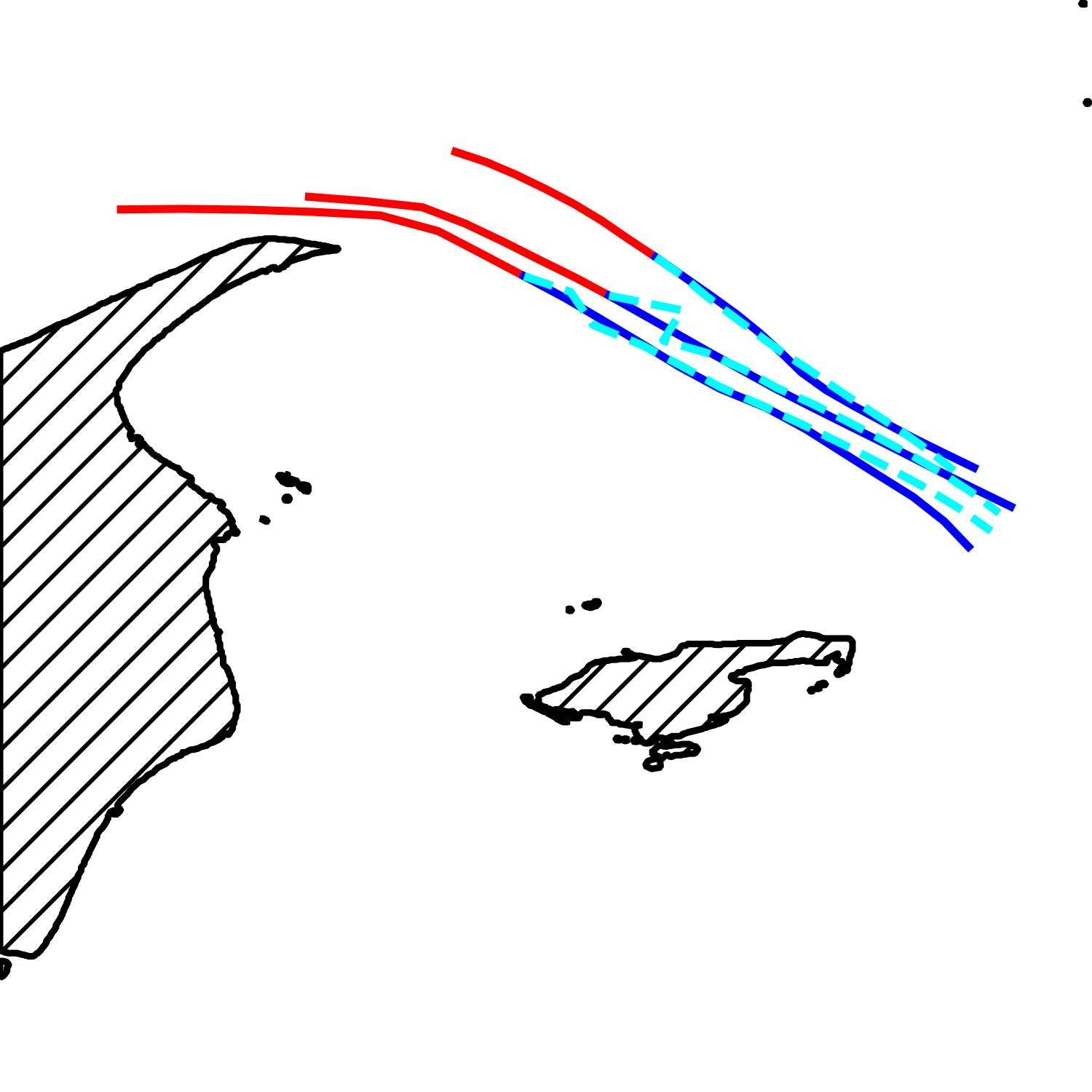}\vspace{4pt}
\includegraphics[width=1.05\linewidth]{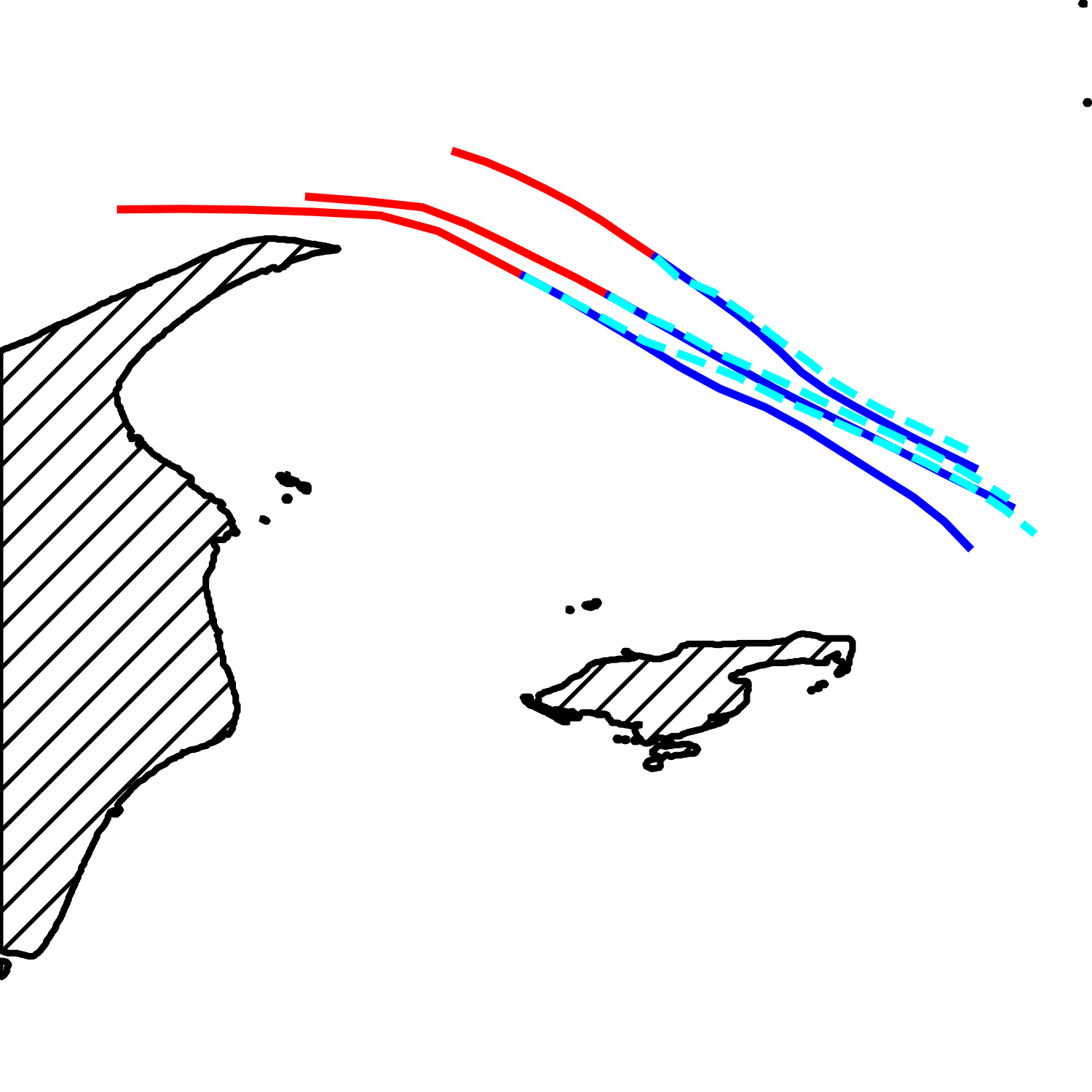}
\end{minipage}}
\hspace{8pt}
\subfigure[\scriptsize{Sudden Turn}]{
\begin{minipage}[b]{0.16\linewidth}
\includegraphics[width=1.05\linewidth]{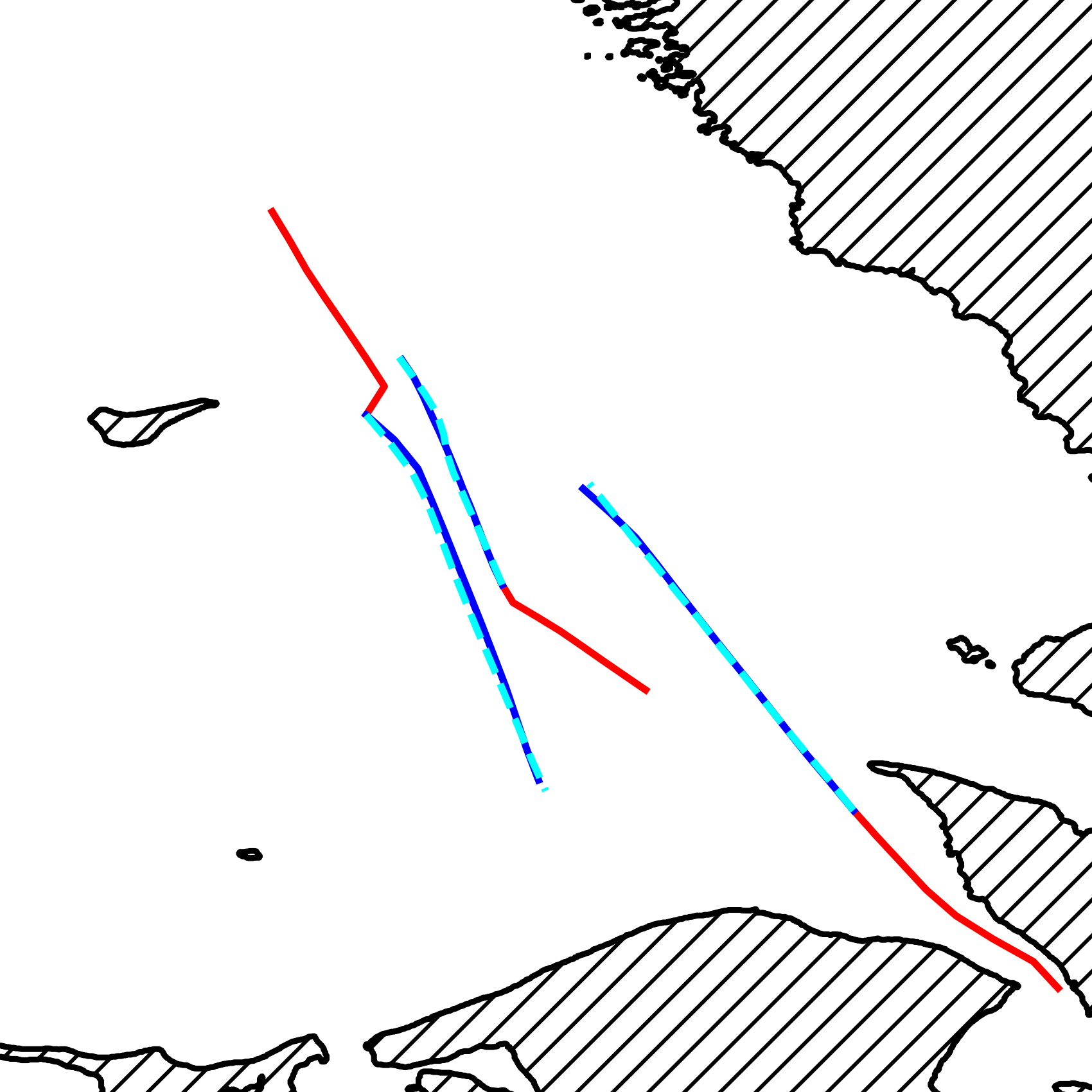}\vspace{4pt}
\includegraphics[width=1.05\linewidth]{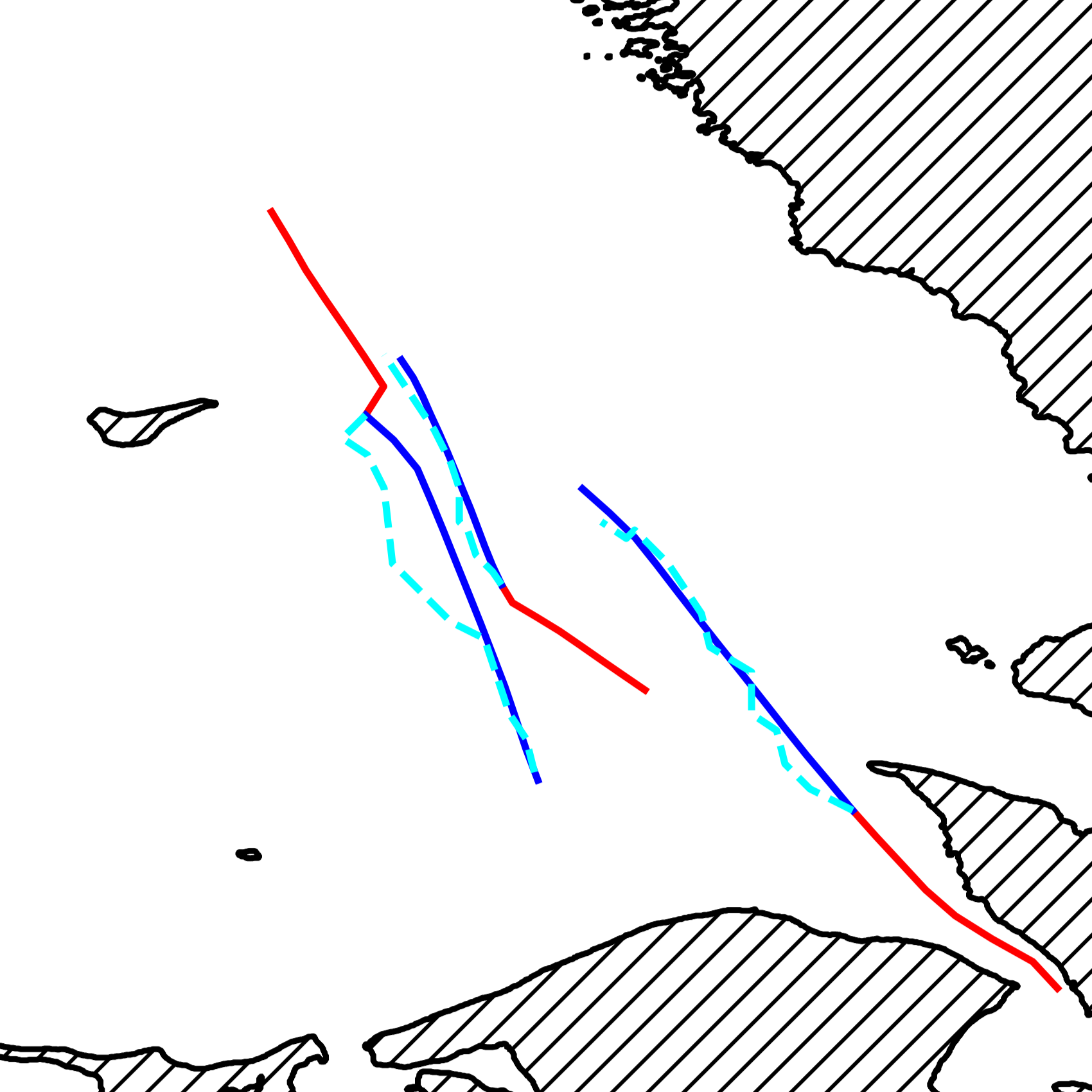}\vspace{4pt}
\includegraphics[width=1.05\linewidth]{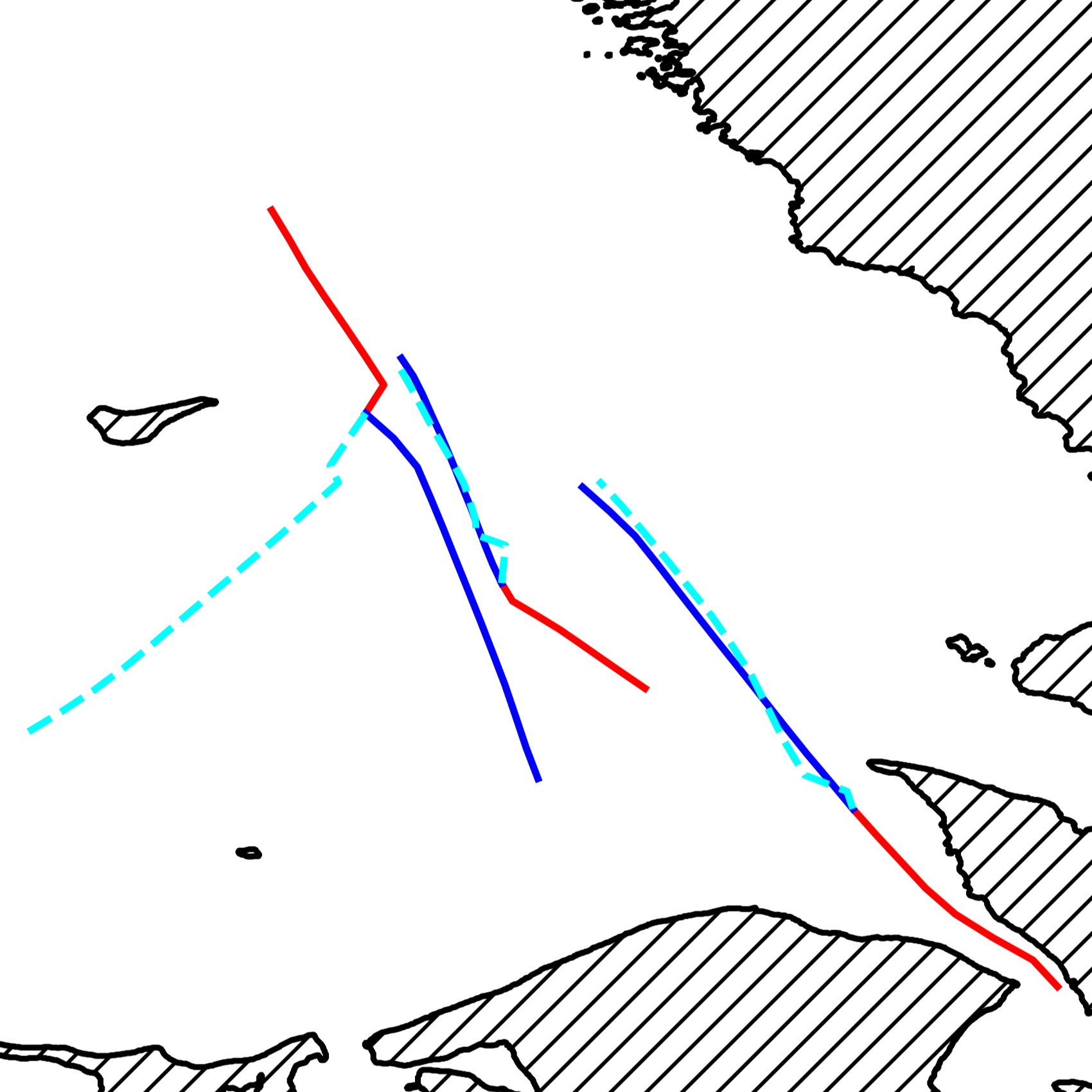}\vspace{4pt}
\includegraphics[width=1.05\linewidth]{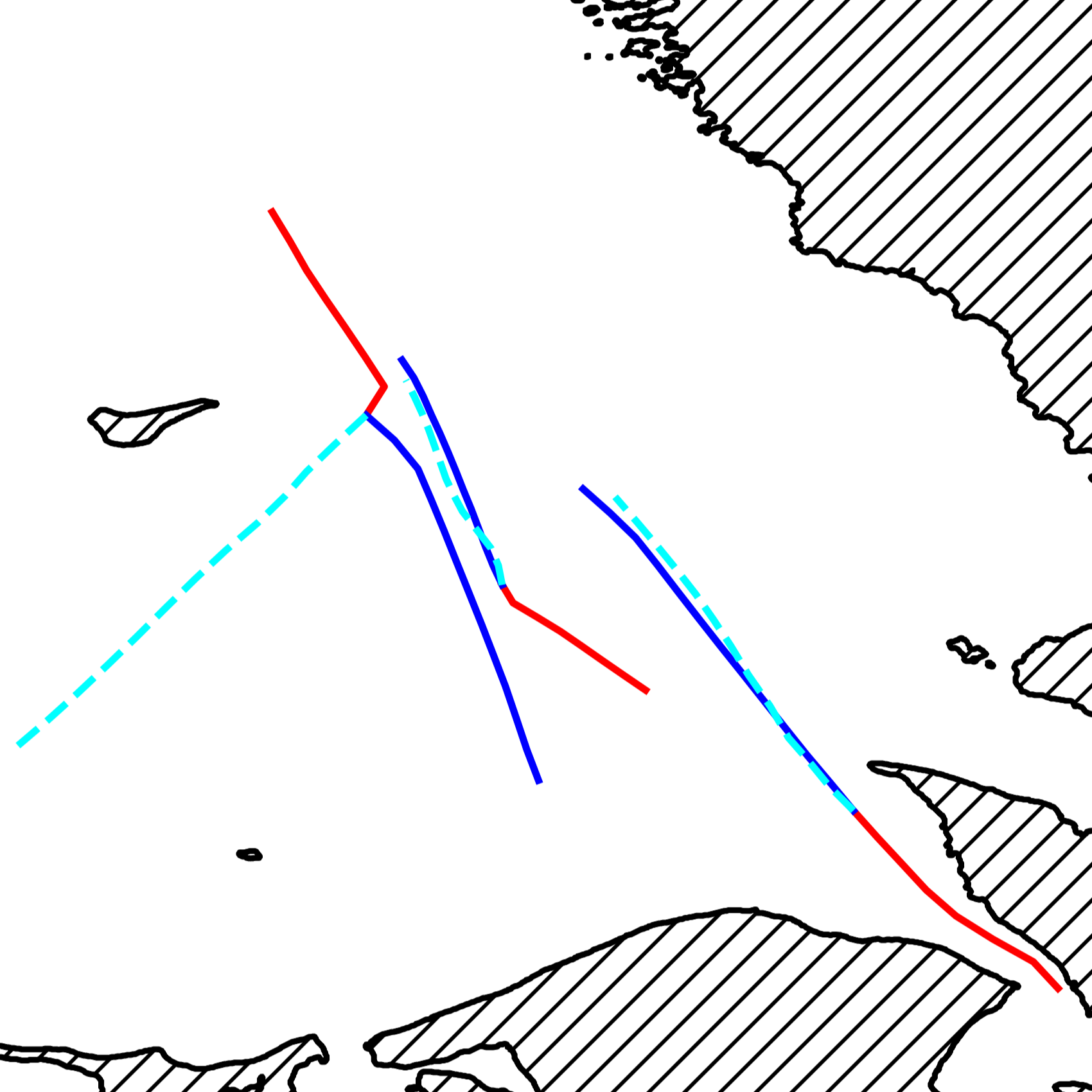}
\end{minipage}}
\hspace{8pt}
\subfigure[\scriptsize{Encounter+Turn}]{
\begin{minipage}[b]{0.16\linewidth}
\includegraphics[width=1.05\linewidth]{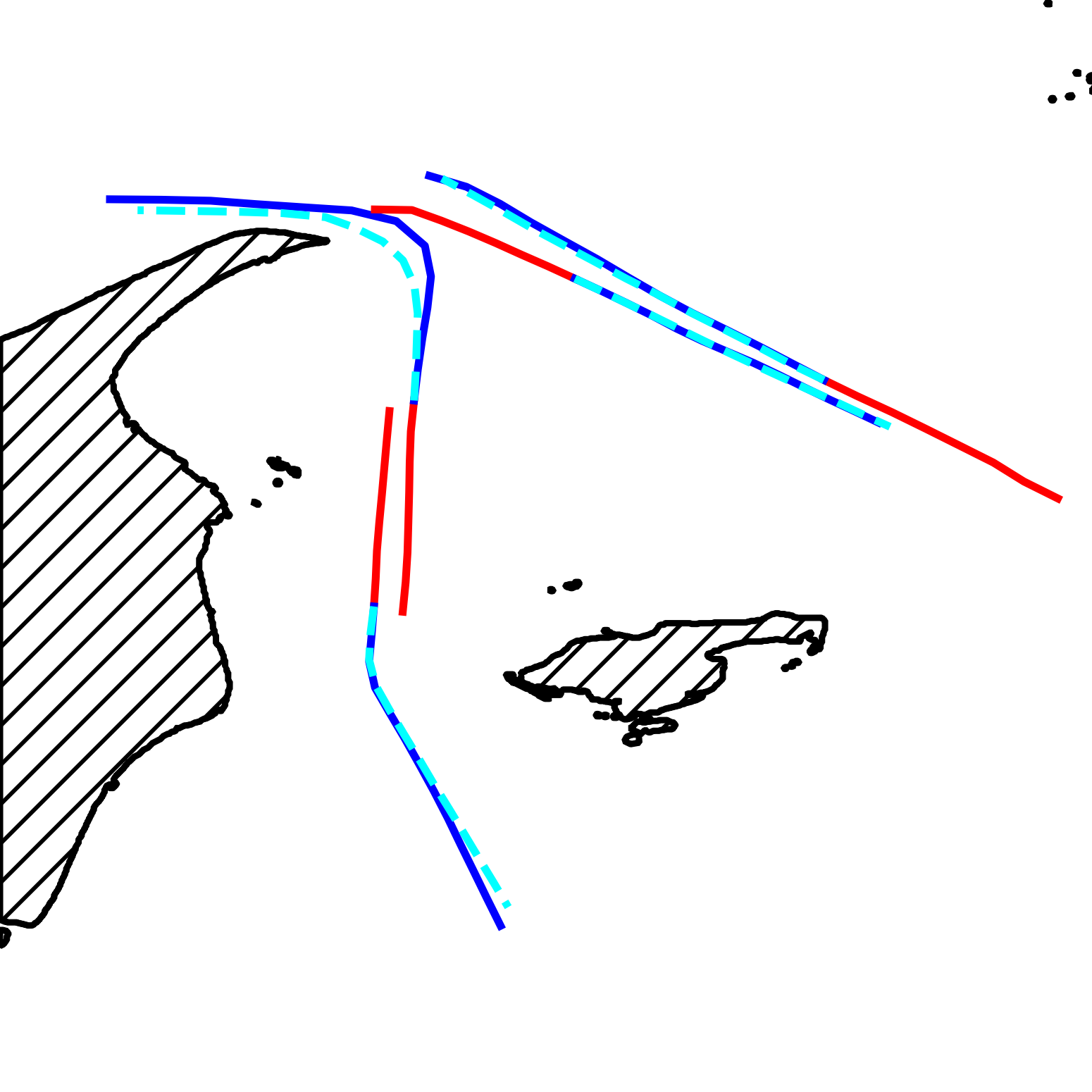}\vspace{4pt}
\includegraphics[width=1.05\linewidth]{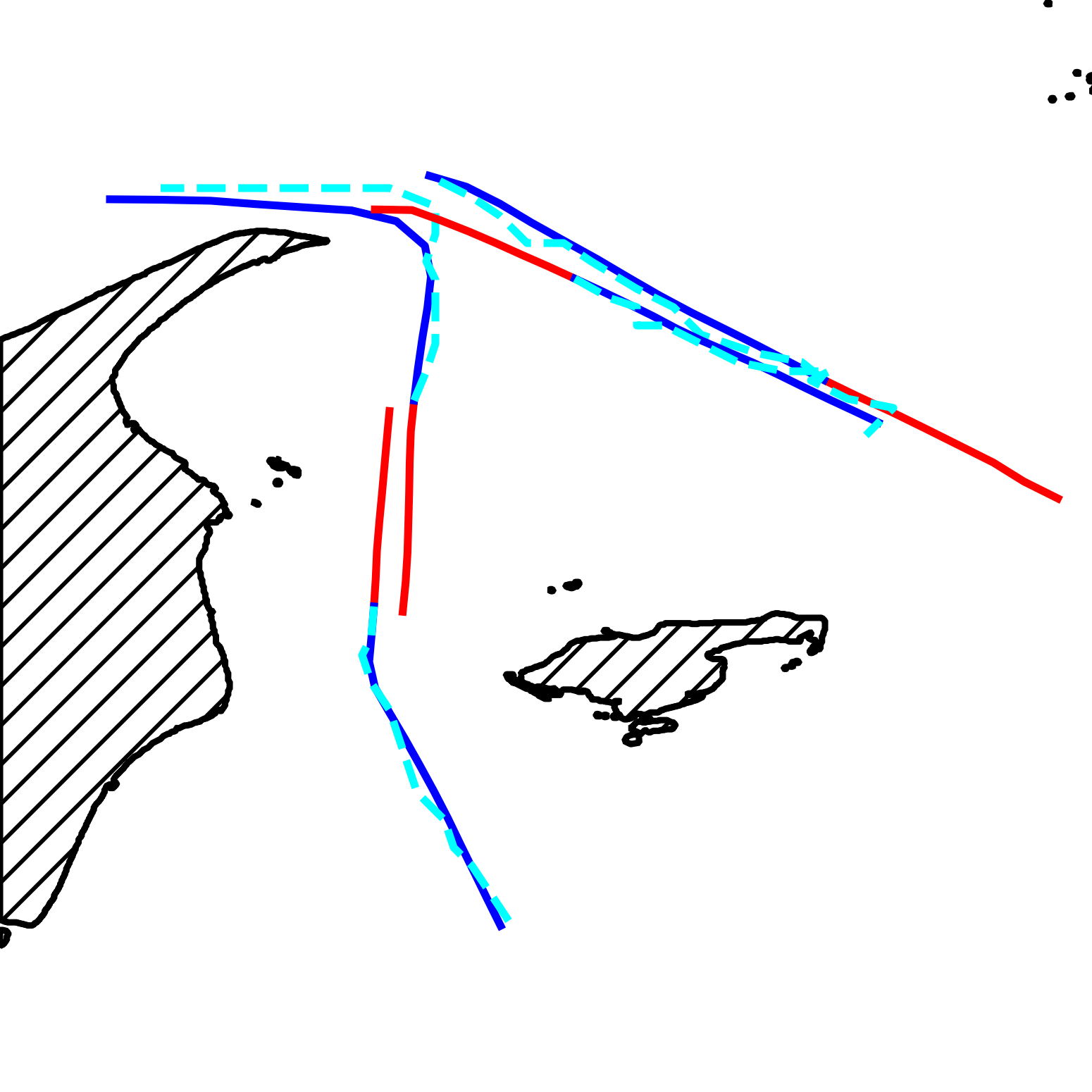}\vspace{4pt}
\includegraphics[width=1.05\linewidth]{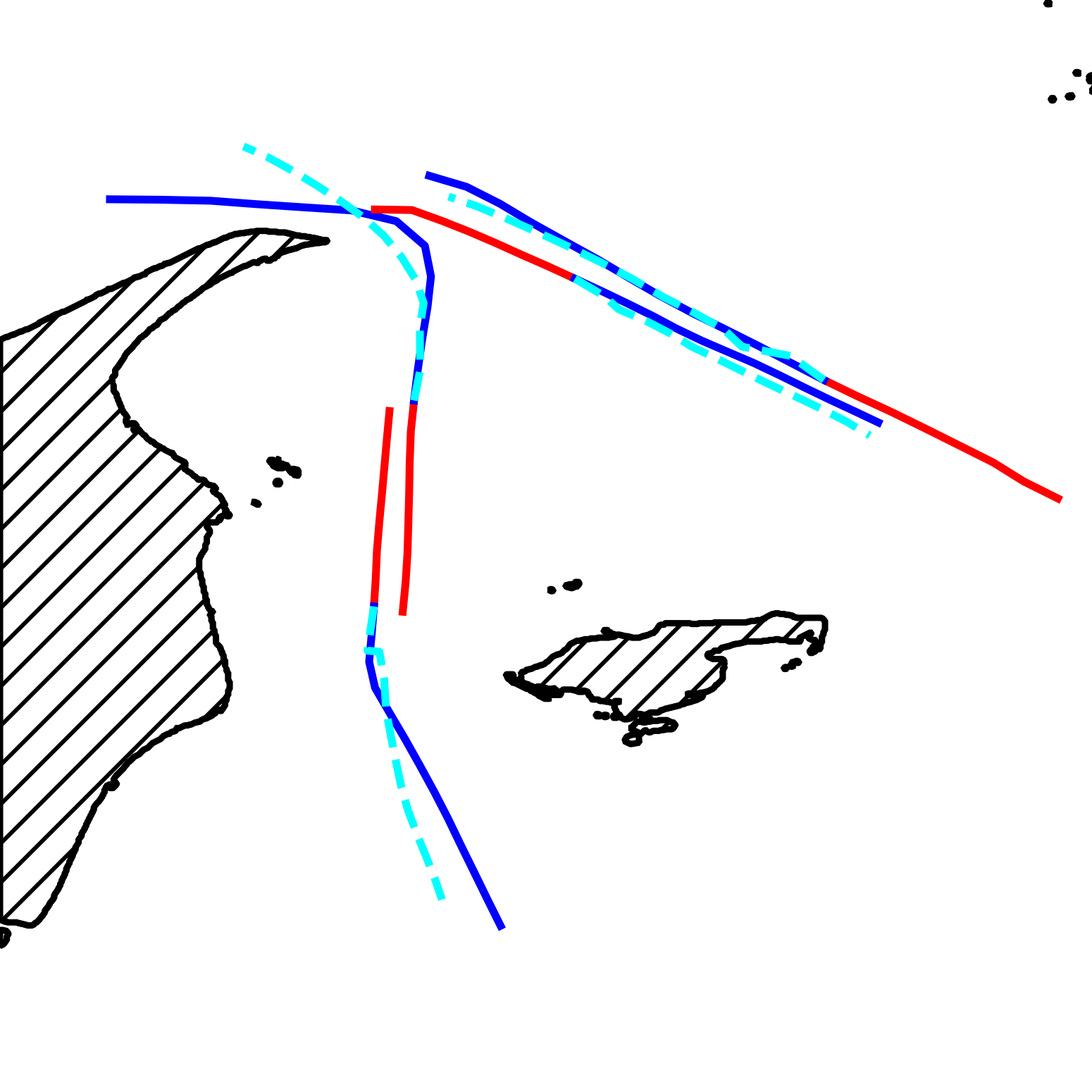}\vspace{4pt}
\includegraphics[width=1.05\linewidth]{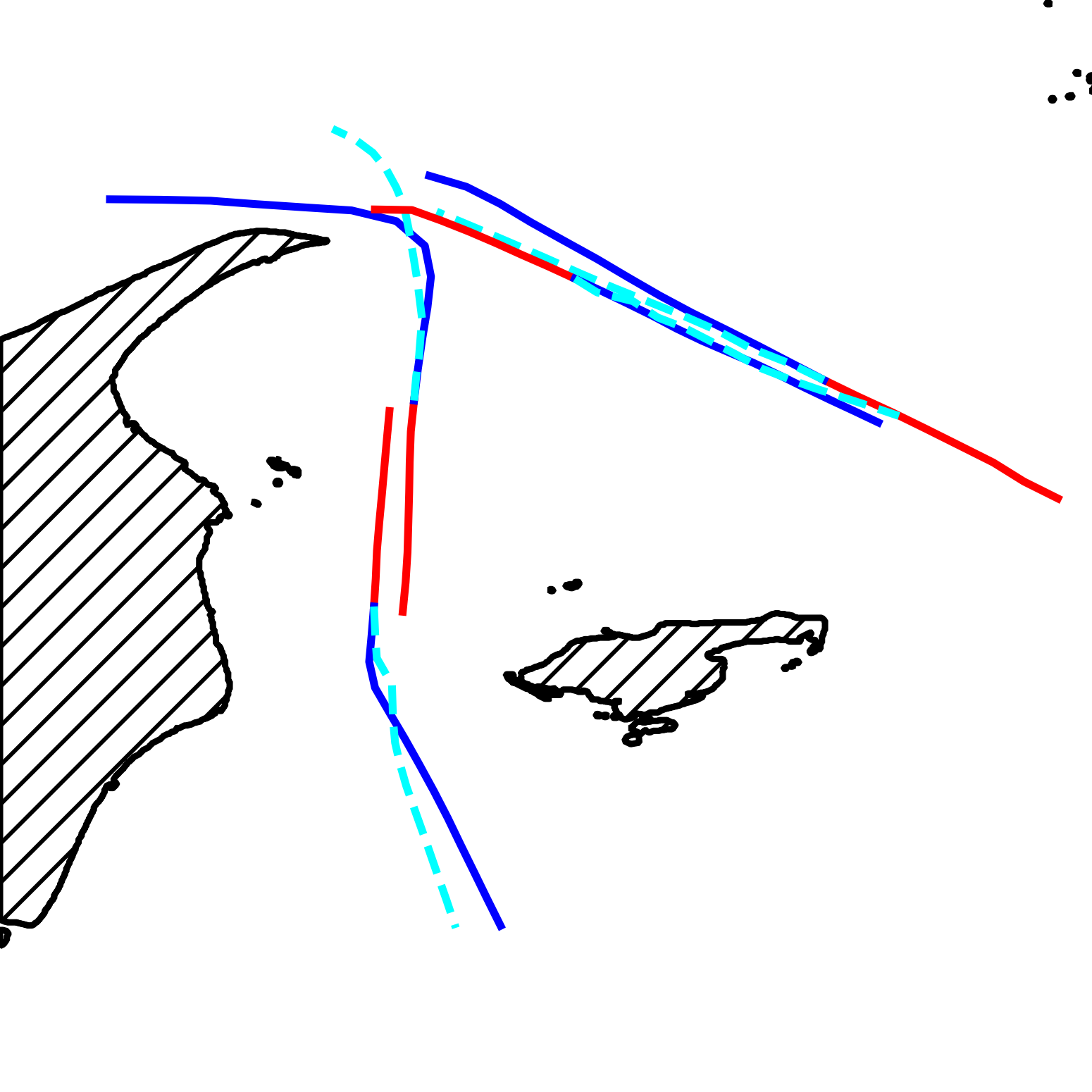}
\end{minipage}}
\caption{Qualitative results on five different scenes on the Danish Straits dataset. Given the observed trajectories (red), we illustrate the ground truth paths (blue) and predicted trajectories (dashed cyan) for different competitive baselines. Our method produces more realistic and natural results across different complex scenarios. Best viewed in color and Zoom-in for enhanced detail.}
\label{fig7}
\end{figure*}

\begin{figure*}
\centering
\begin{minipage}[b]{0.01\linewidth}
\rotatebox{90}{\scriptsize{Ours}}\\ \\  \\ \\ \\ \vspace{7pt}
\rotatebox{90}{\scriptsize{TrAISformer}} \\ \\ \\  \\ \\
\vspace{1pt}
\rotatebox{90}{\scriptsize{CVAE}}\\ \\ \\ \\ \vspace{11pt}
\rotatebox{90}{\scriptsize{LSTM-Seq2Seq}}\\ \vspace{14pt}
\end{minipage}
\subfigure[\scriptsize{Island Restriction}]{
\begin{minipage}[b]{0.16\linewidth}
\includegraphics[width=1.05\linewidth]{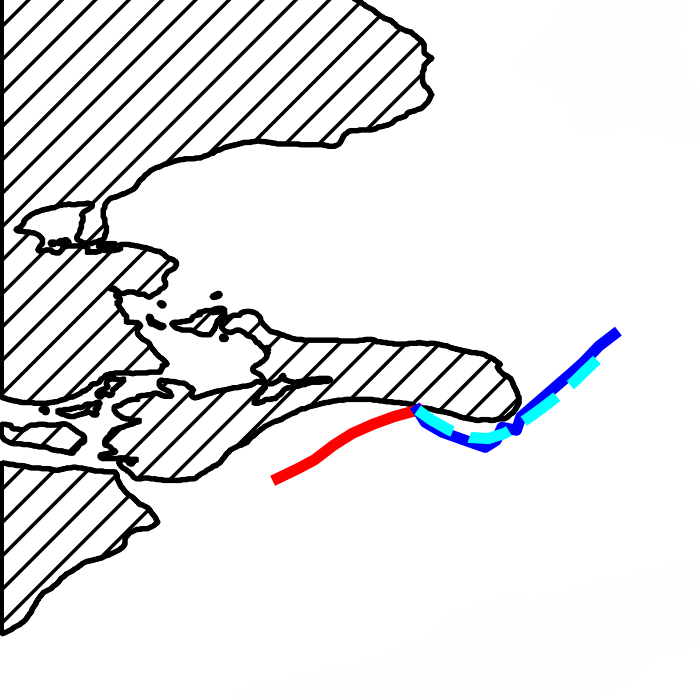}\vspace{4pt}
\includegraphics[width=1.05\linewidth]{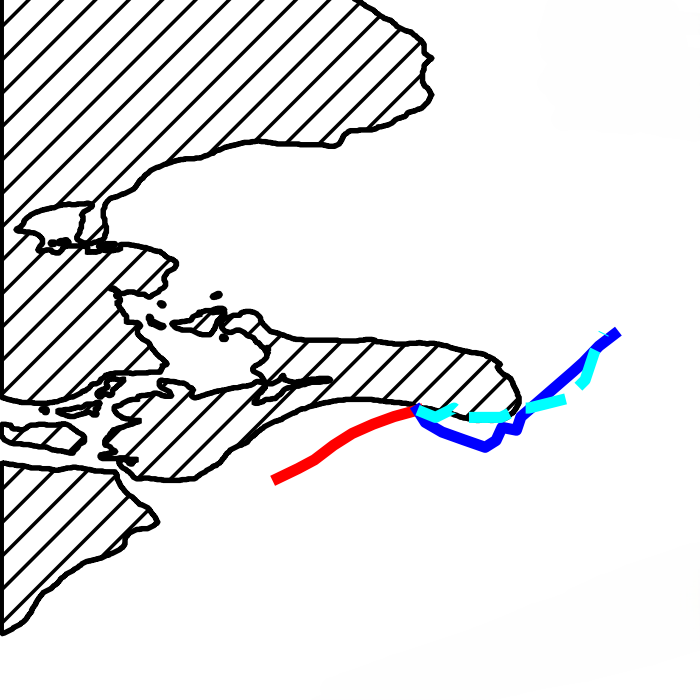}\vspace{4pt}
\includegraphics[width=1.05\linewidth]{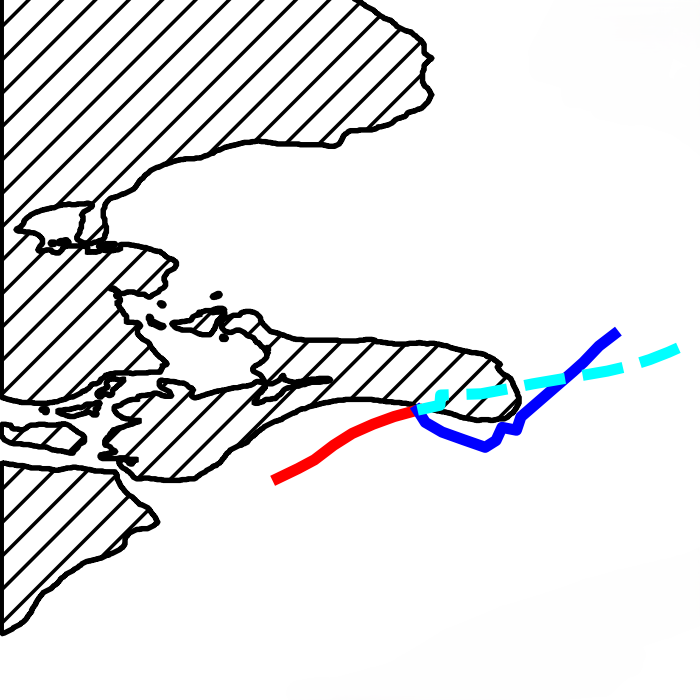}\vspace{4pt}
\includegraphics[width=1.05\linewidth]{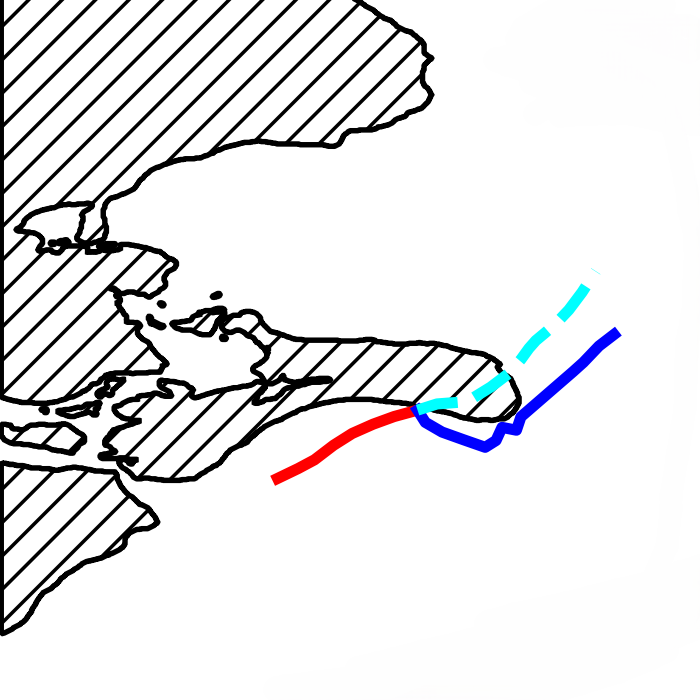}
\end{minipage}}
\hspace{8pt}
\subfigure[\scriptsize{Narrow Waterway}]{
\begin{minipage}[b]{0.16\linewidth}
\includegraphics[width=1.05\linewidth]{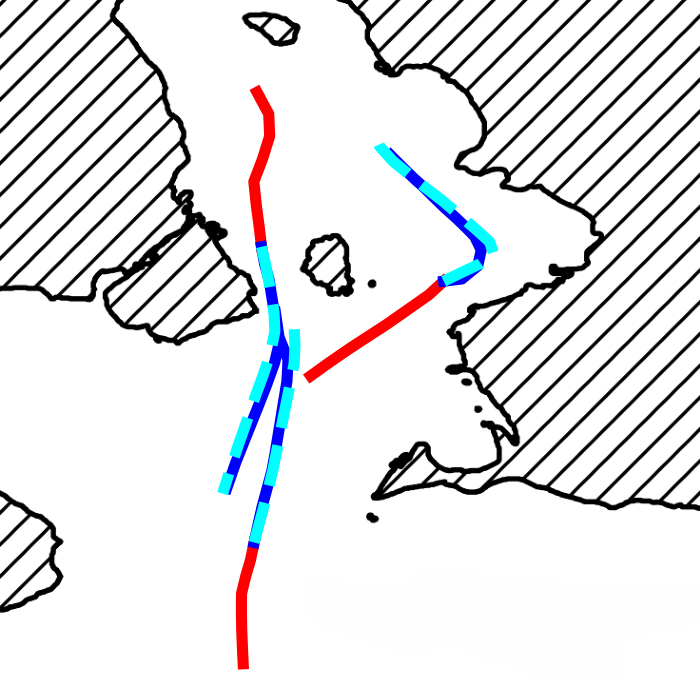}\vspace{4pt}
\includegraphics[width=1.05\linewidth]{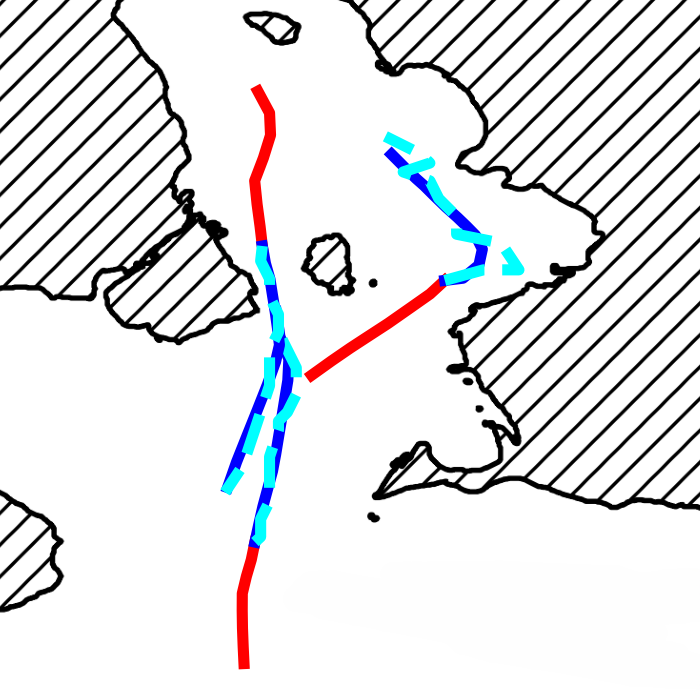}\vspace{4pt}
\includegraphics[width=1.05\linewidth]{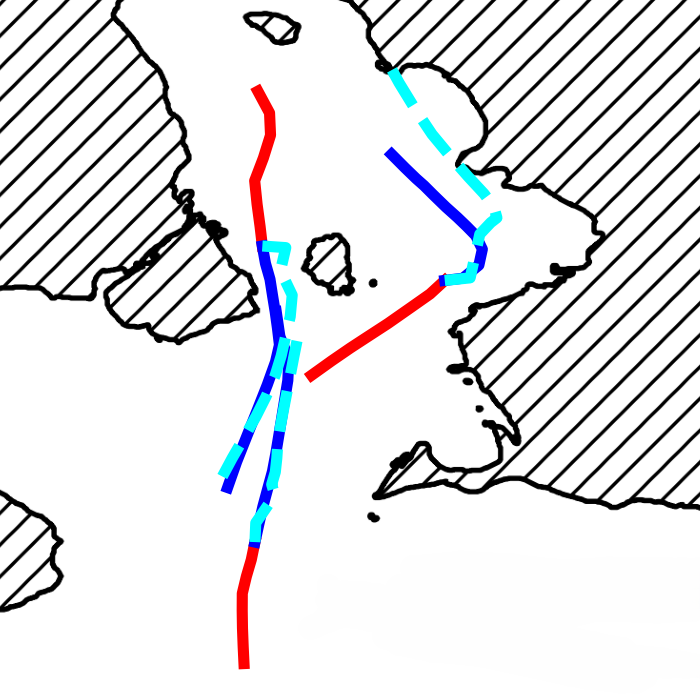}\vspace{4pt}
\includegraphics[width=1.05\linewidth]{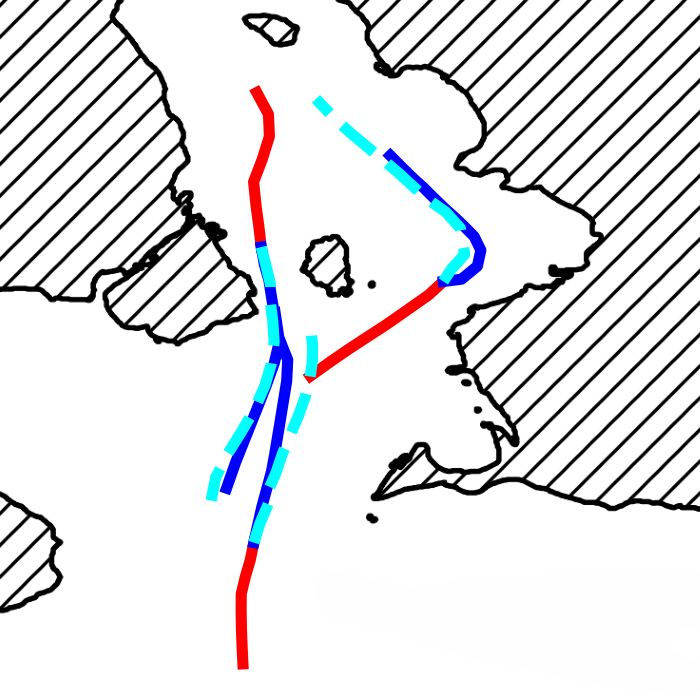}
\end{minipage}}
\hspace{8pt}
\subfigure[\scriptsize{Multi-Vessel Encounter}]{
\begin{minipage}[b]{0.16\linewidth}
\includegraphics[width=1.05\linewidth]{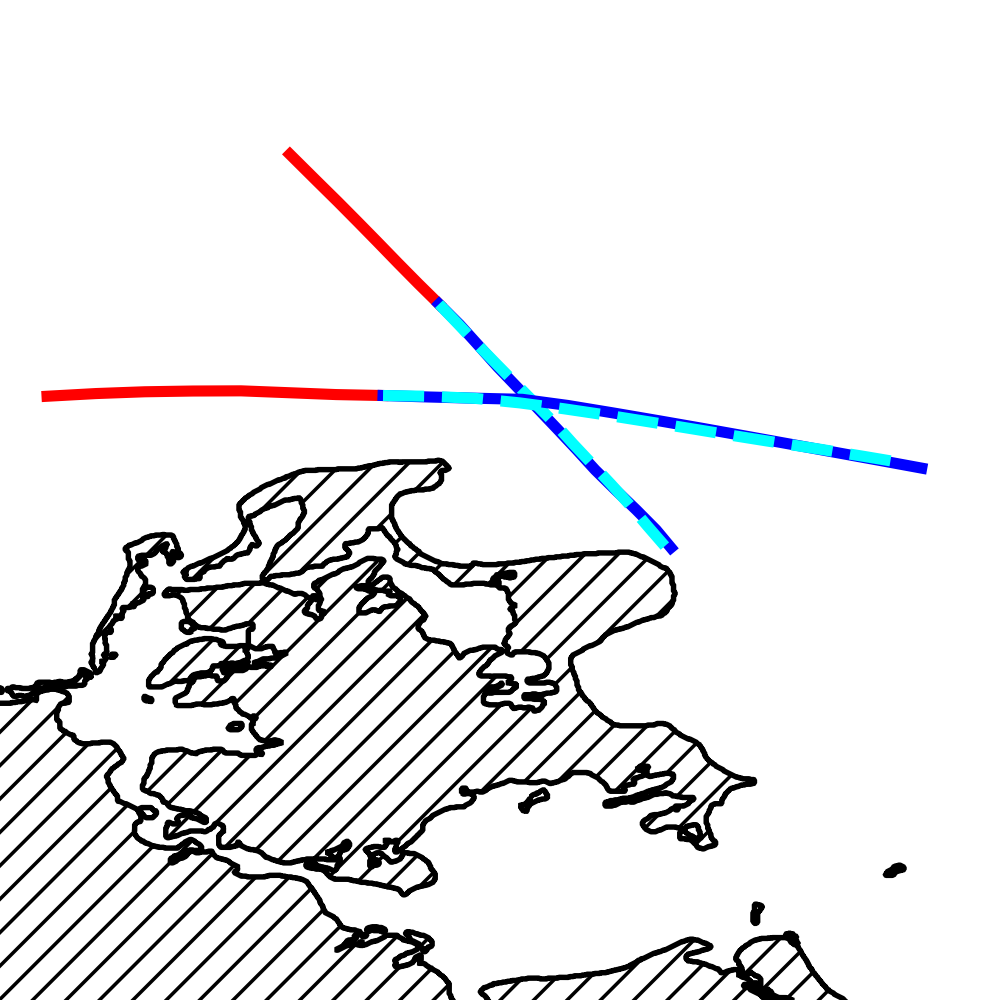}\vspace{4pt}
\includegraphics[width=1.05\linewidth]{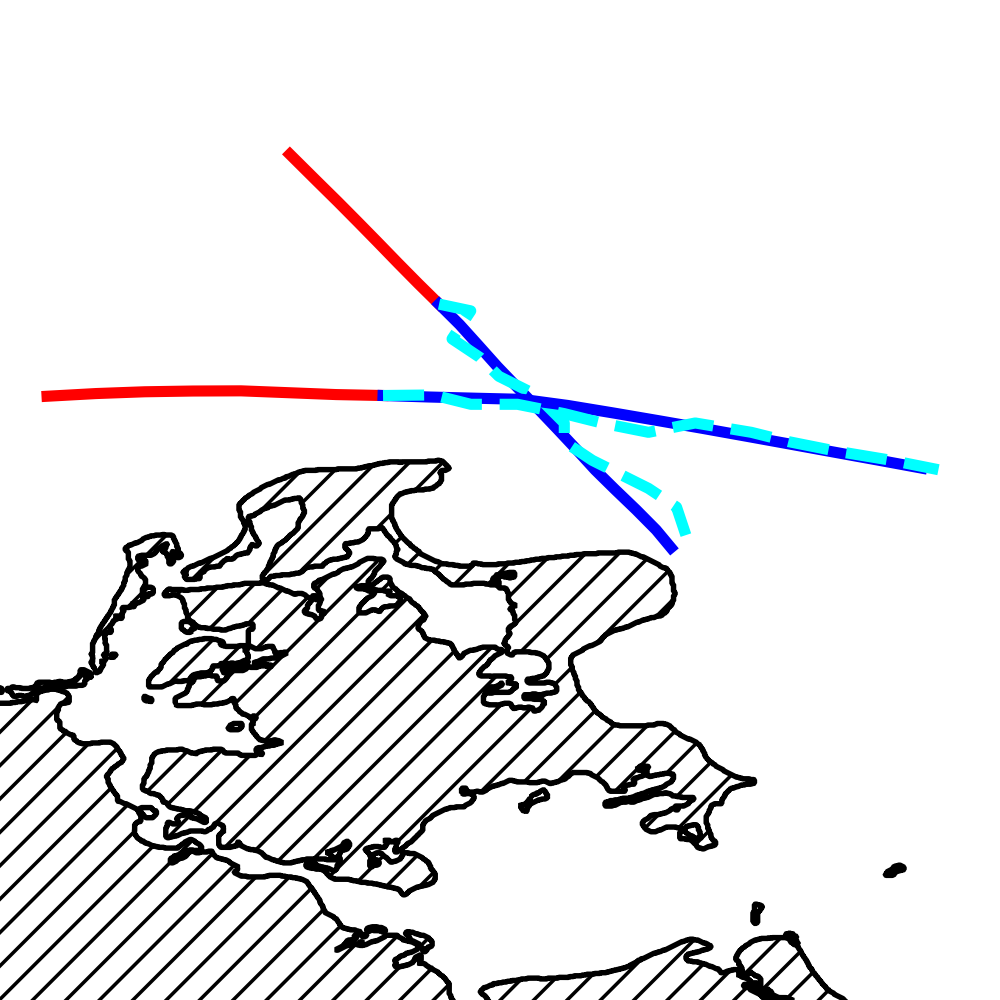}\vspace{4pt}
\includegraphics[width=1.05\linewidth]{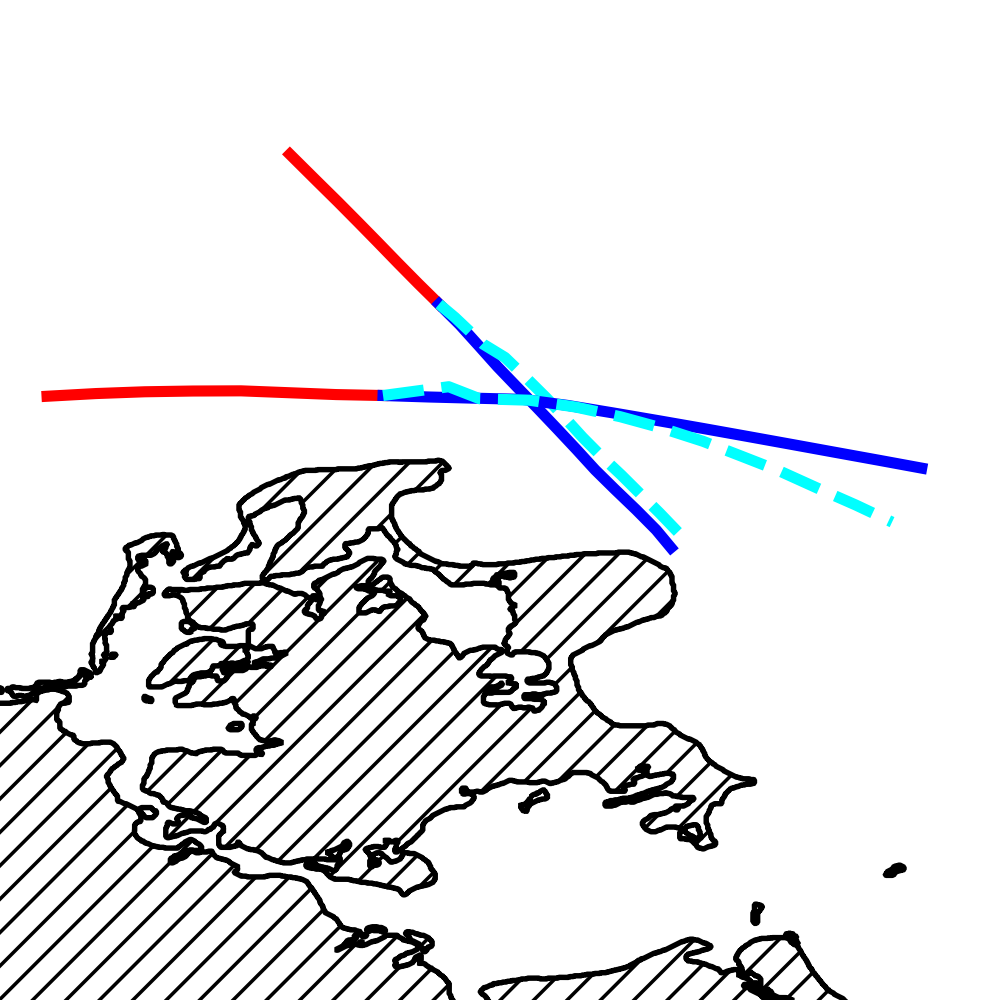}\vspace{4pt}
\includegraphics[width=1.05\linewidth]{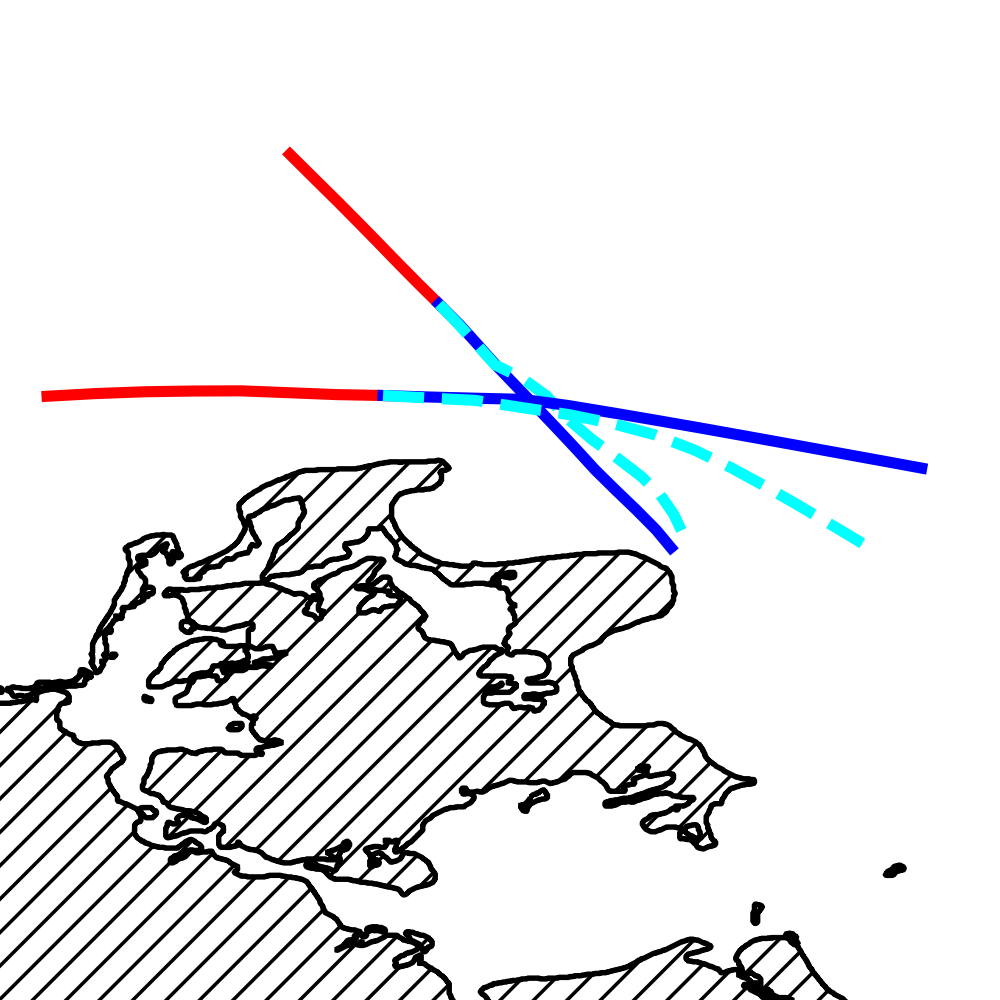}
\end{minipage}}
\hspace{8pt}
\subfigure[\scriptsize{Sudden Turn}]{
\begin{minipage}[b]{0.16\linewidth}
\includegraphics[width=1.05\linewidth]{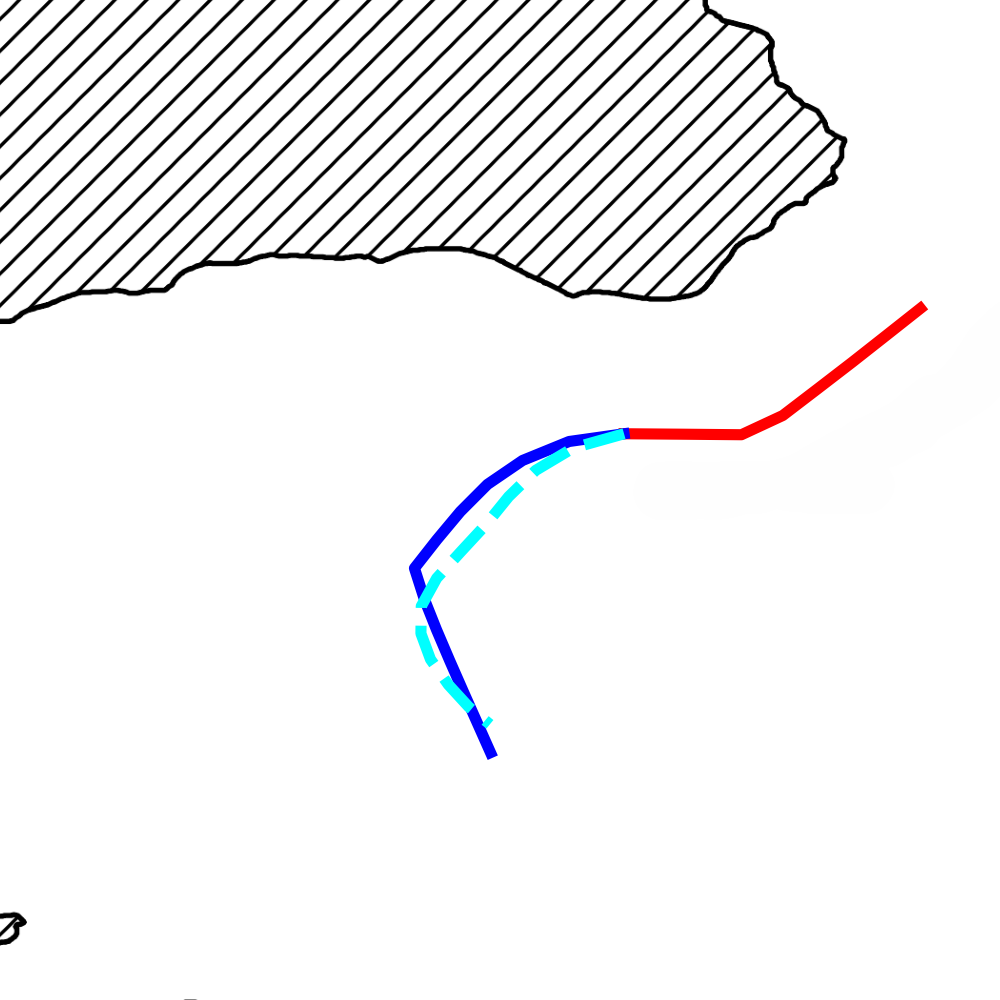}\vspace{4pt}
\includegraphics[width=1.05\linewidth]{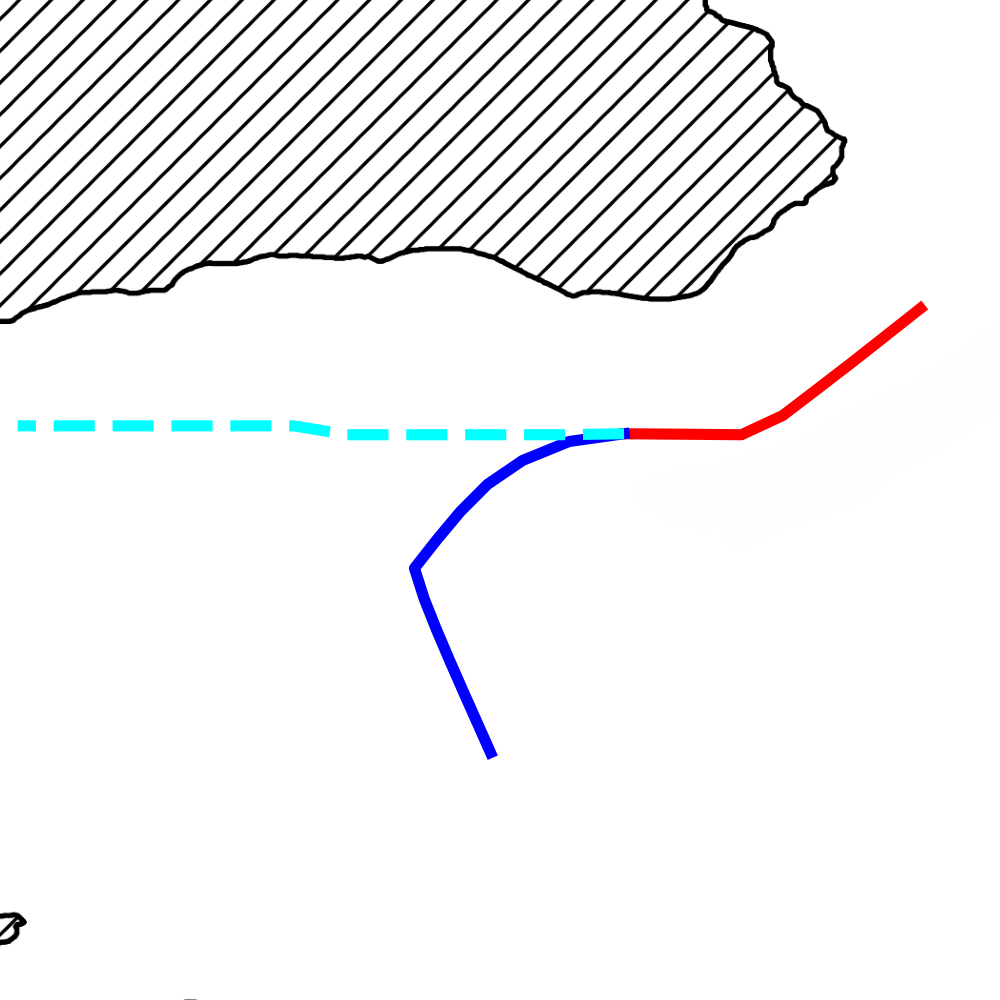}\vspace{4pt}
\includegraphics[width=1.05\linewidth]{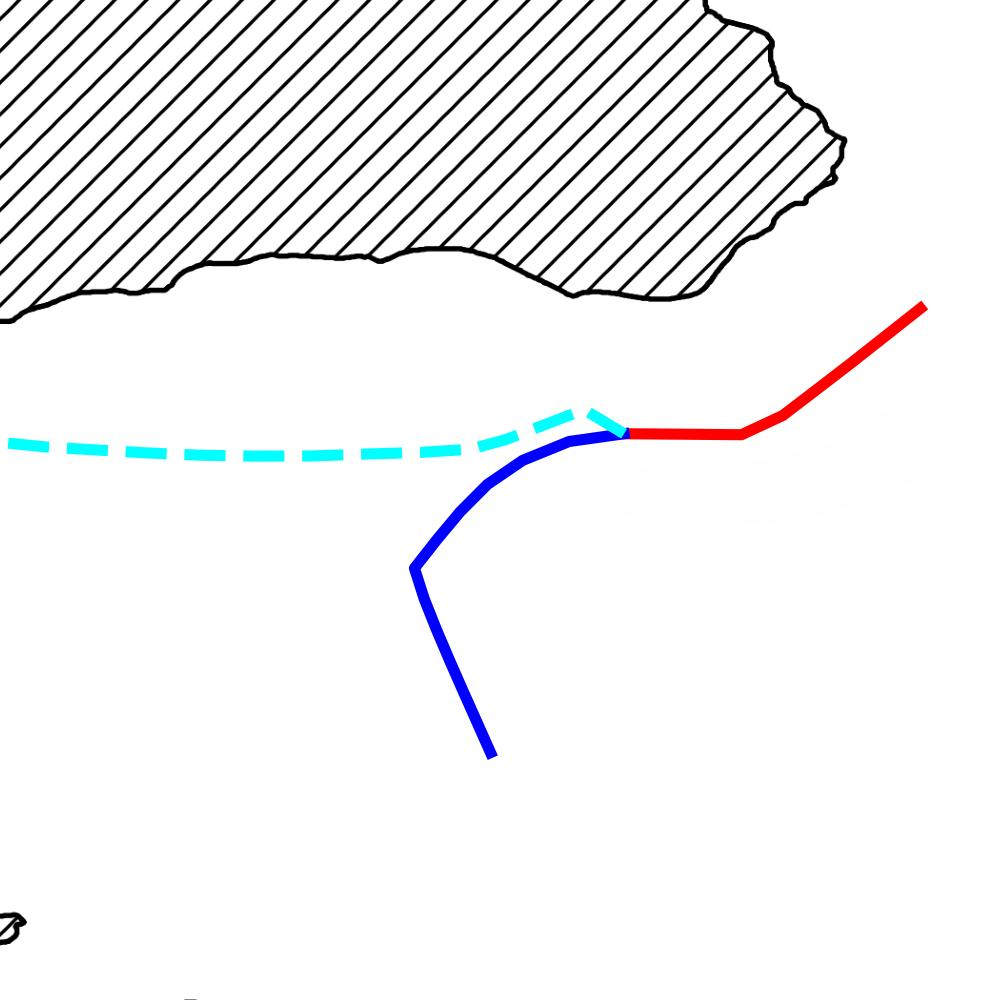}\vspace{4pt}
\includegraphics[width=1.05\linewidth]{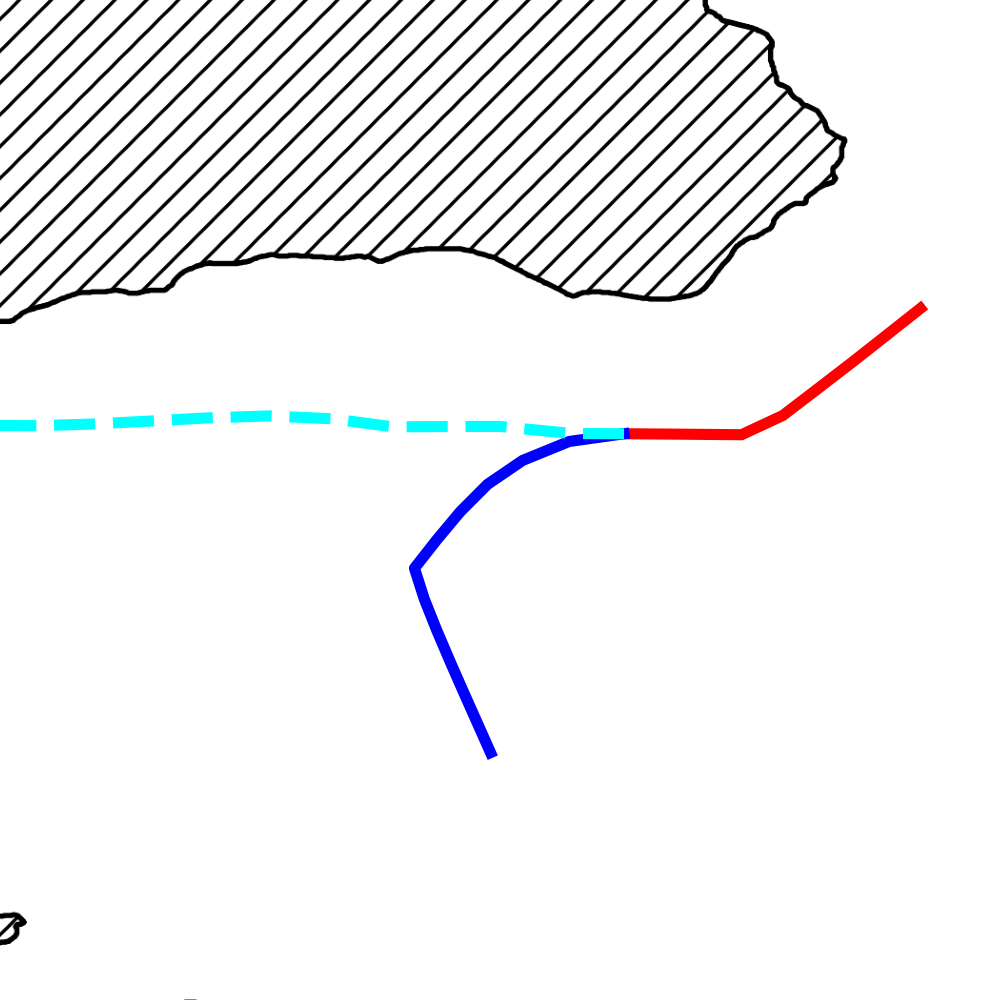}
\end{minipage}}
\hspace{8pt}
\subfigure[\scriptsize{Encounter+Turn}]{
\begin{minipage}[b]{0.16\linewidth}
\includegraphics[width=1.05\linewidth]{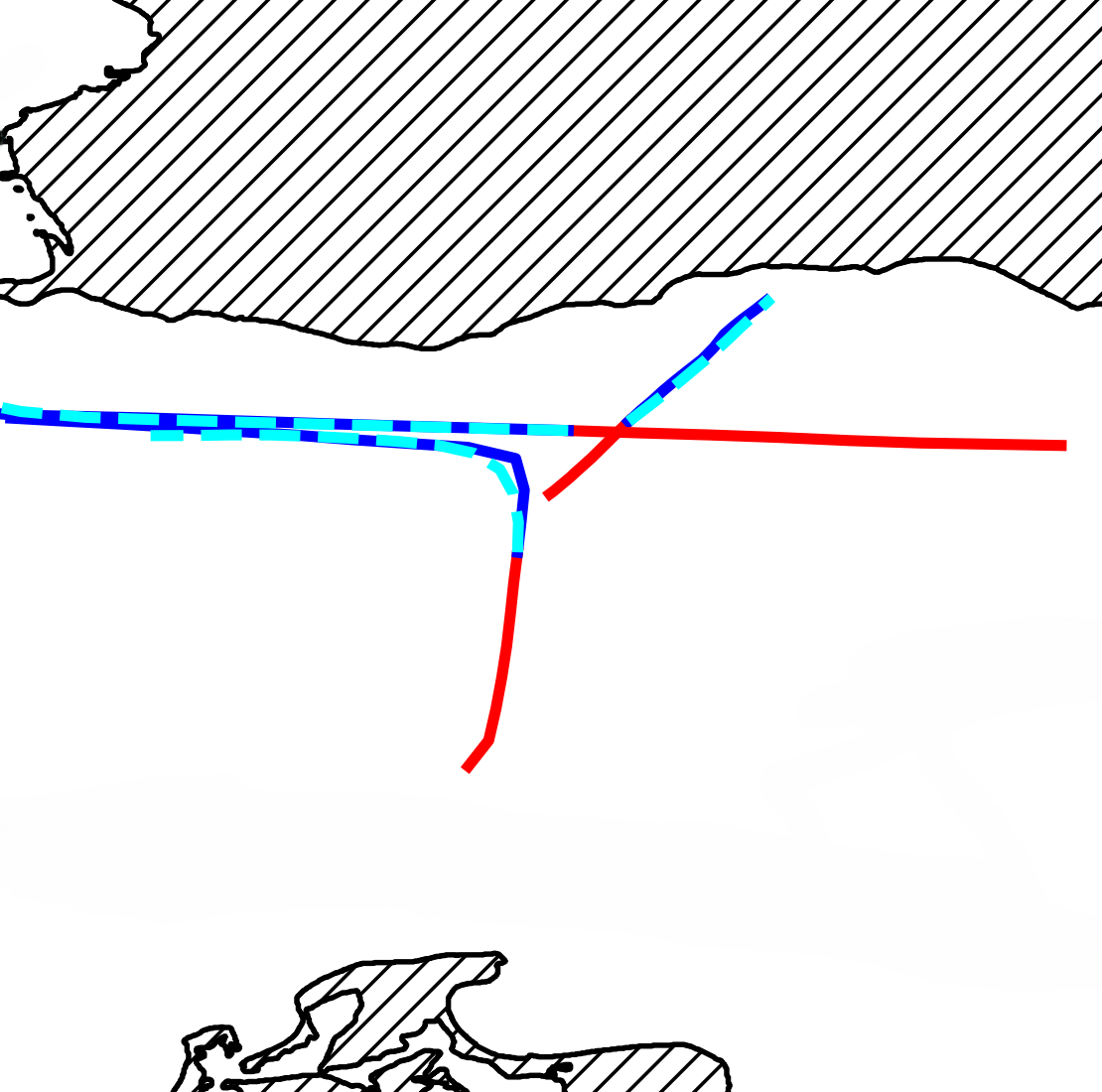}\vspace{4pt}
\includegraphics[width=1.05\linewidth]{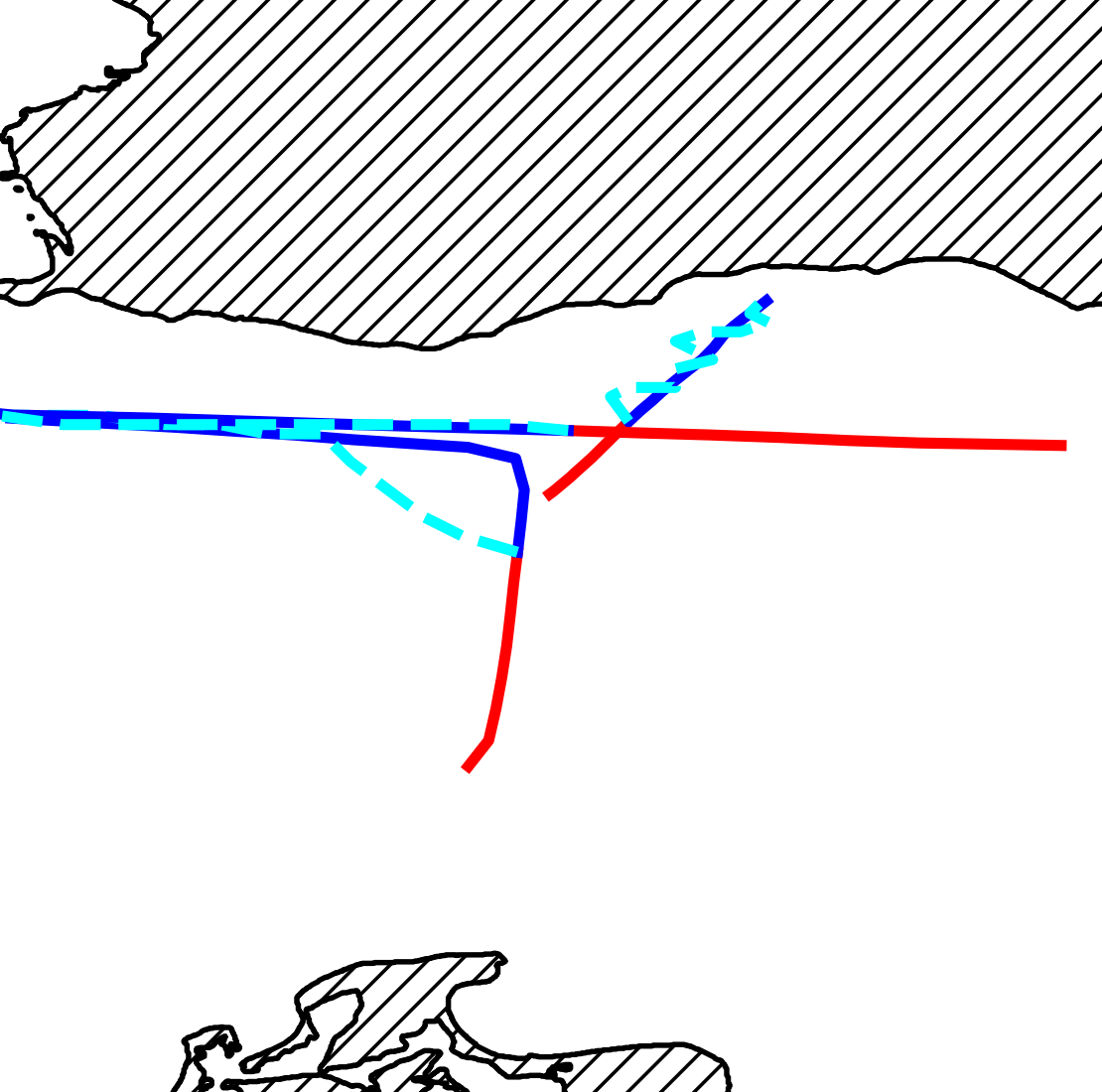}\vspace{4pt}
\includegraphics[width=1.05\linewidth]{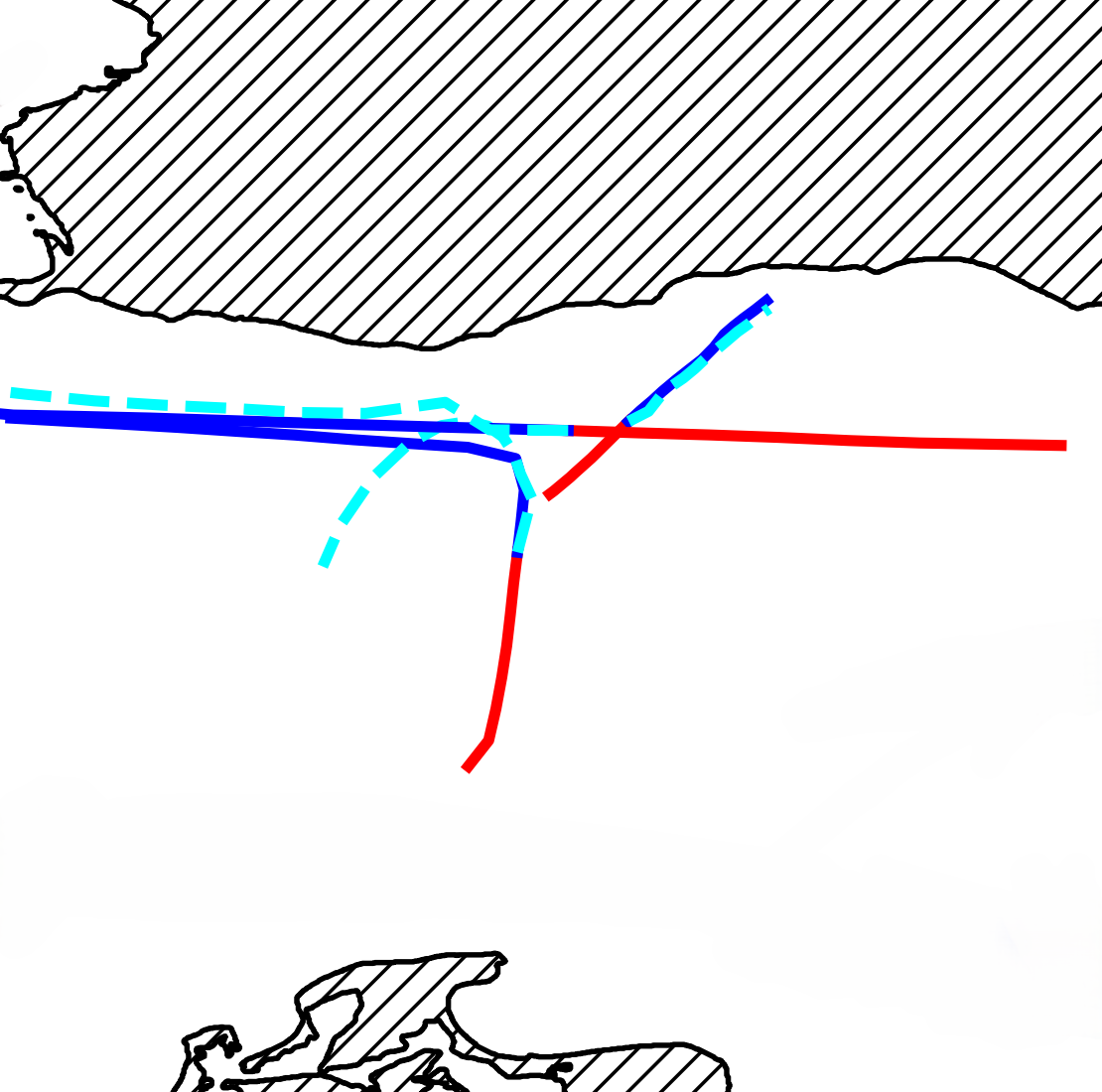}\vspace{4pt}
\includegraphics[width=1.05\linewidth]{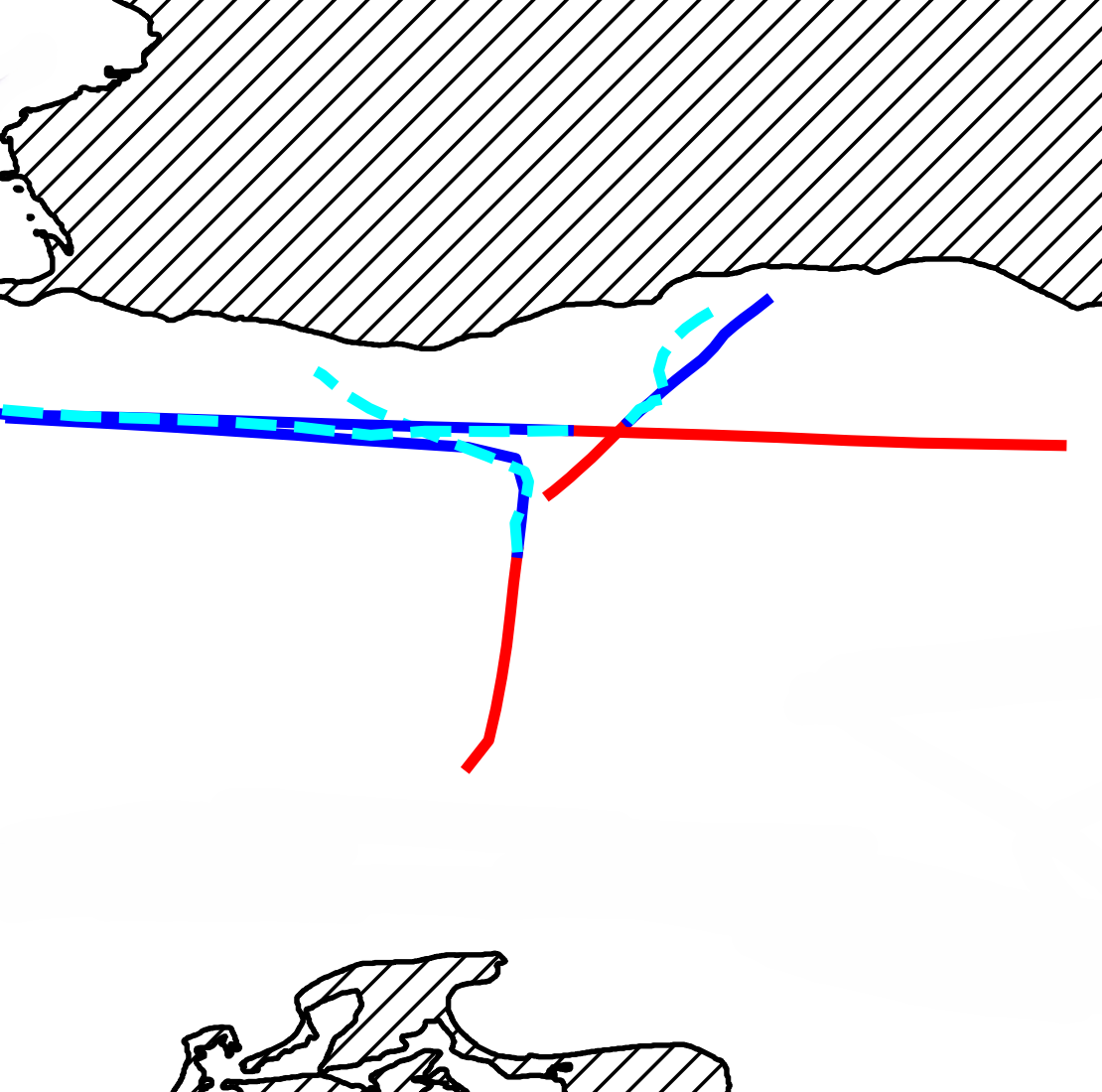}
\end{minipage}}
\caption{Qualitative results on five different scenes on the Baltic Sea dataset. Given the observed trajectories (red), we illustrate the ground truth paths (blue) and predicted trajectories (dashed cyan) for different competitive baselines. Our method produces more realistic and natural results across different complex scenarios. Best viewed in color and Zoom-in for enhanced detail.}
\label{fig9}
\end{figure*}

\begin{figure*}
\hspace{8pt}
	\centering
	\subfigure{
			\includegraphics[width=1.16in]{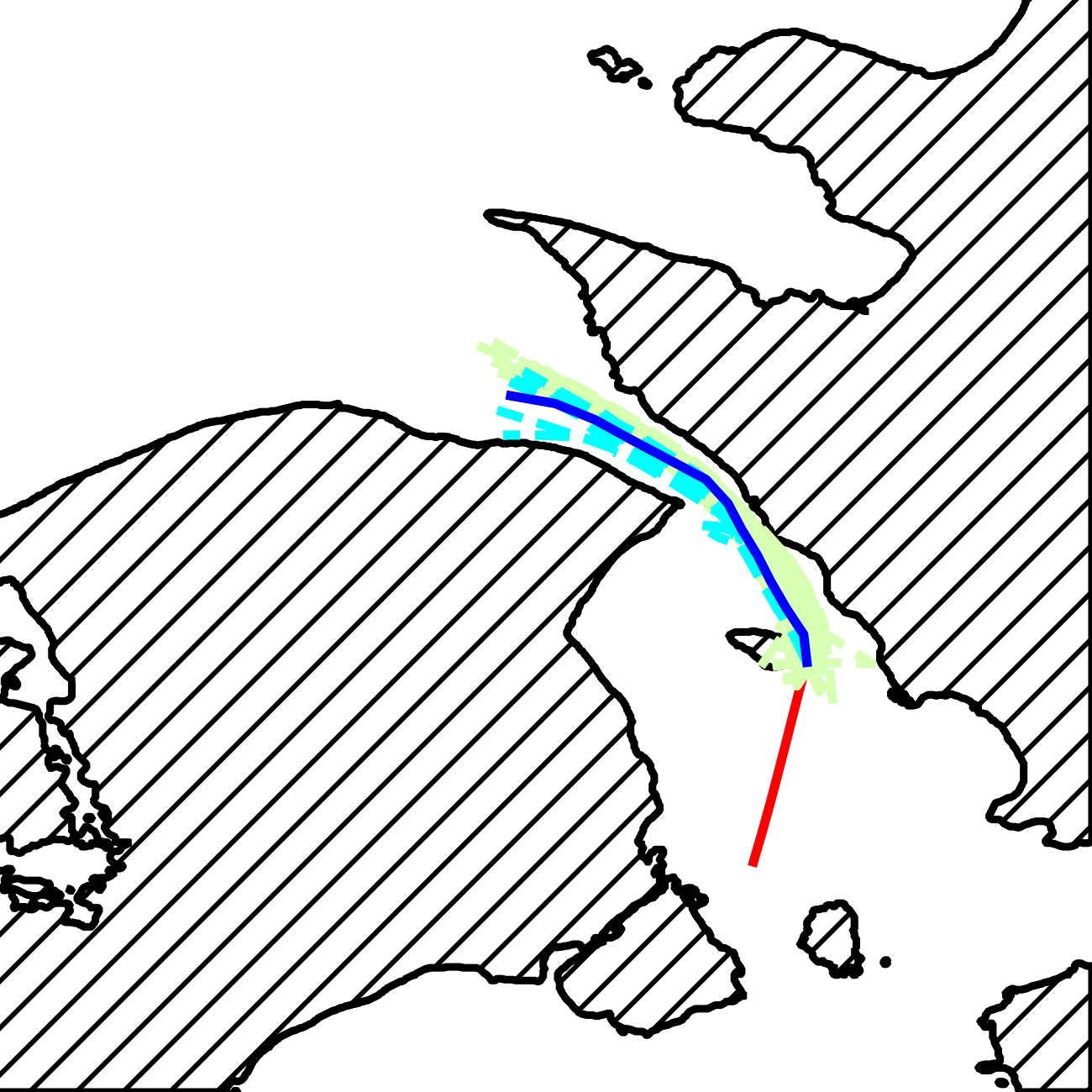}
	}%
\hspace{8pt}
	\subfigure{		
			\includegraphics[width=1.16in]{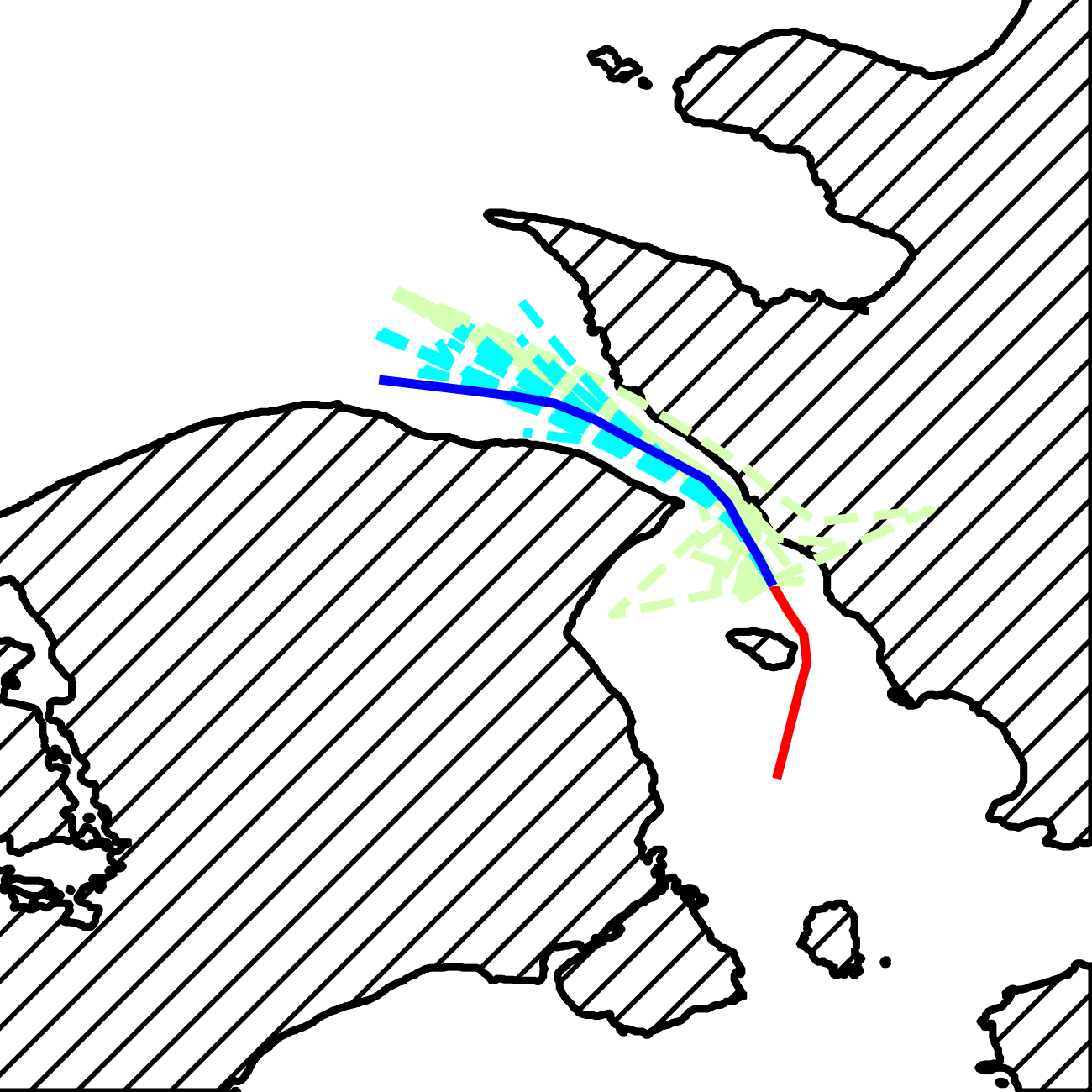}
	}%
\hspace{8pt}
        \subfigure{
			\includegraphics[width=1.16in]{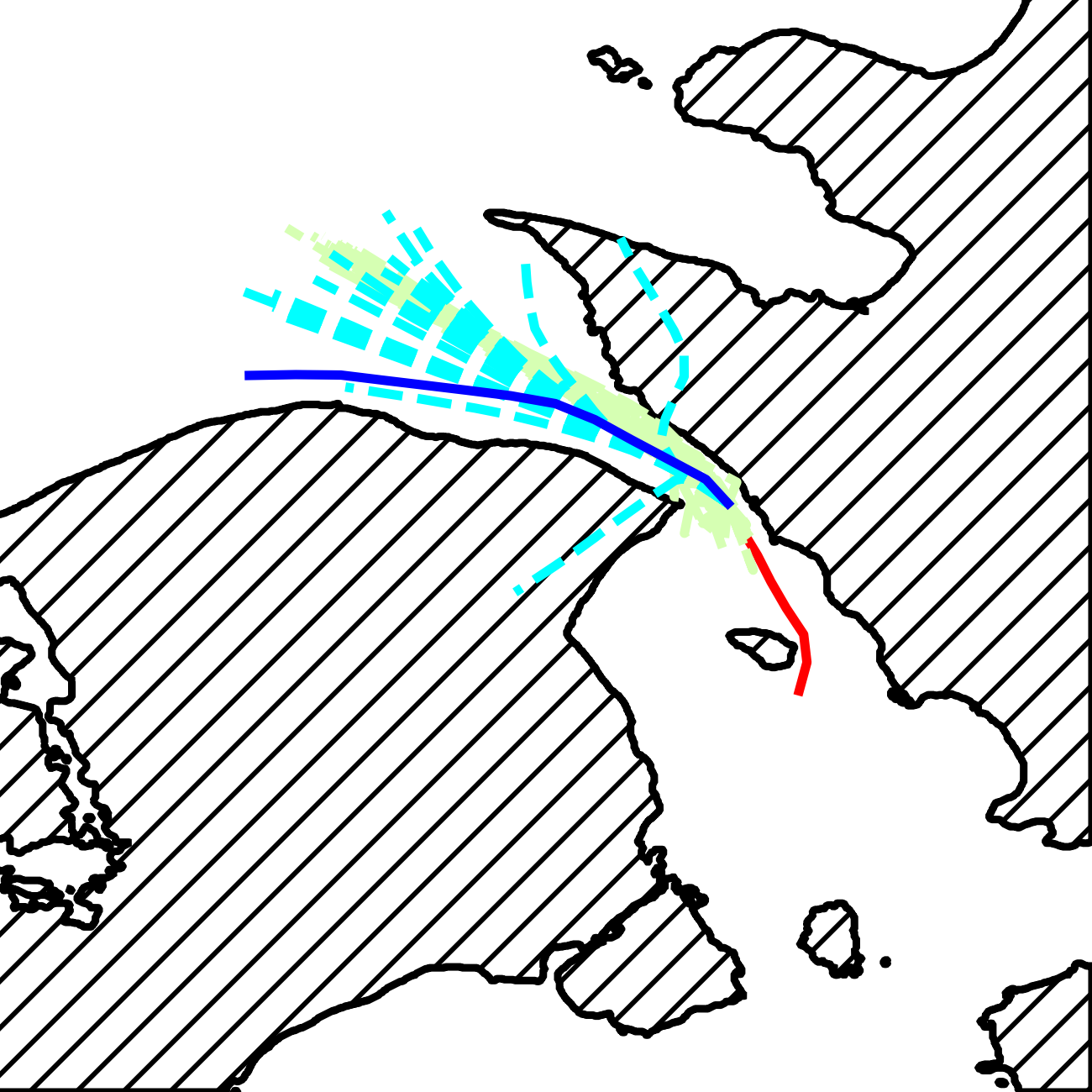}
	}%
\hspace{8pt}
	\subfigure{		
			\includegraphics[width=1.16in]{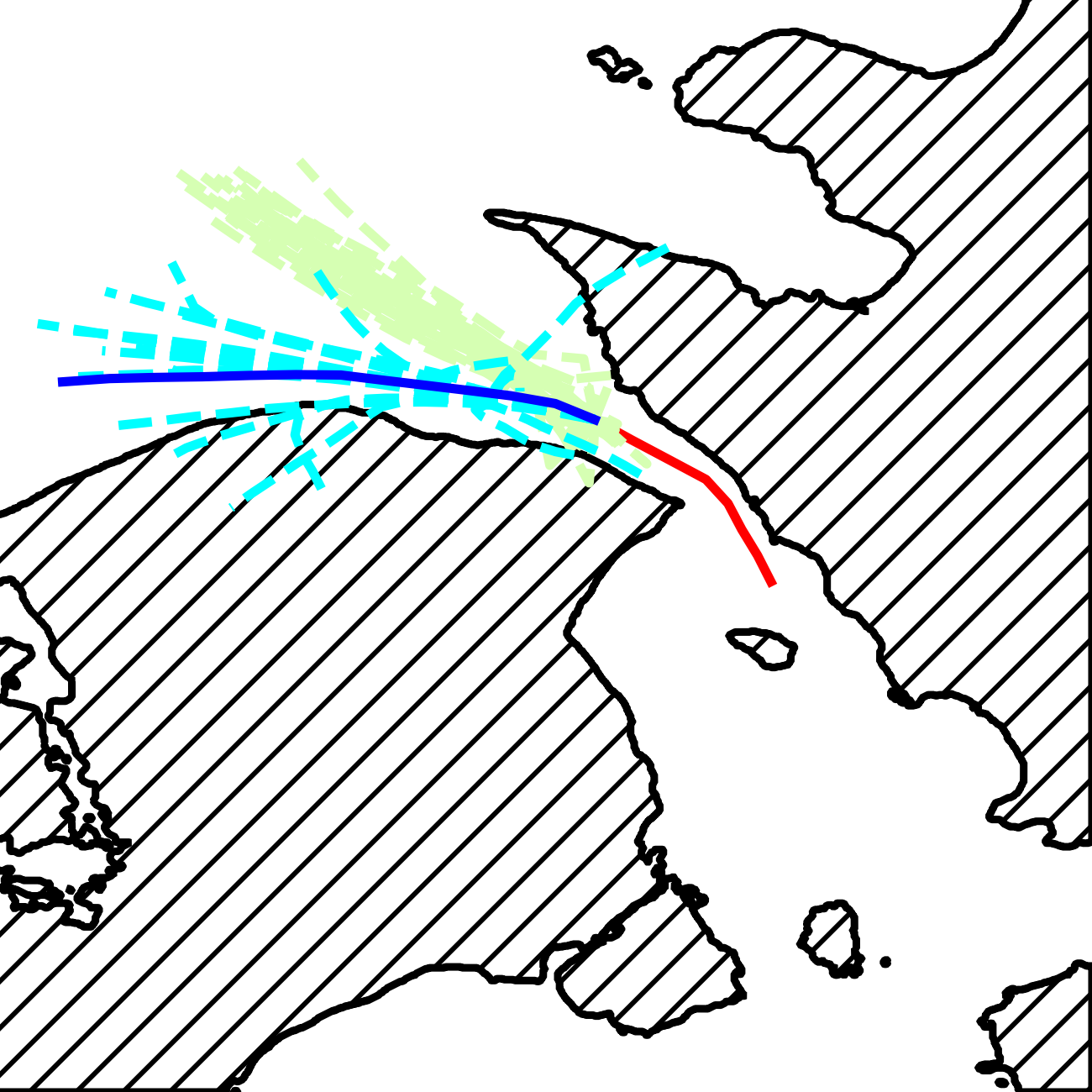}
	}%
\hspace{8pt}
	\subfigure{
			\includegraphics[width=1.16in]{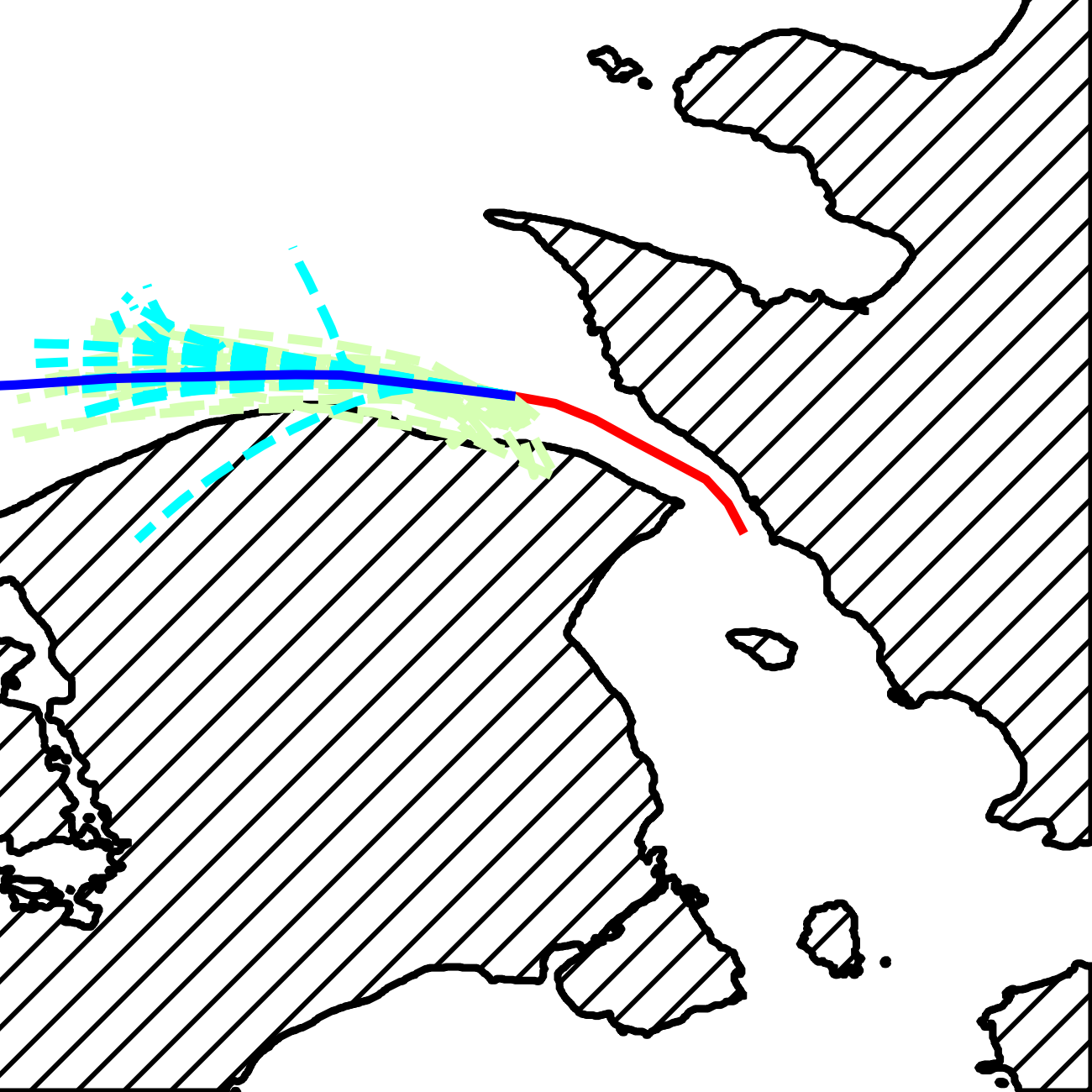}
	}%
 
	\subfigure{
			\includegraphics[width=1.16in]{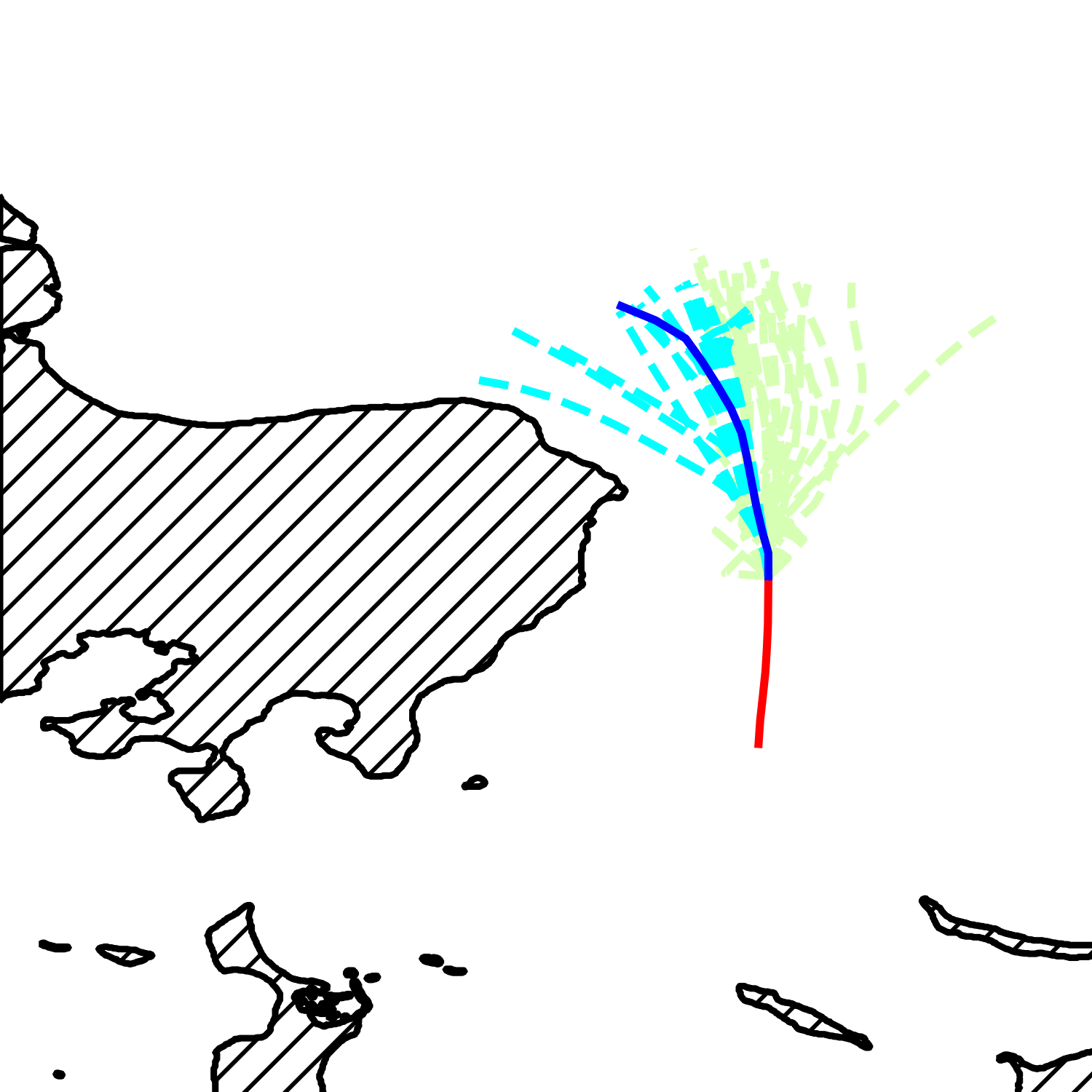}
	}%
\hspace{8pt}
        \subfigure{
			\includegraphics[width=1.16in]{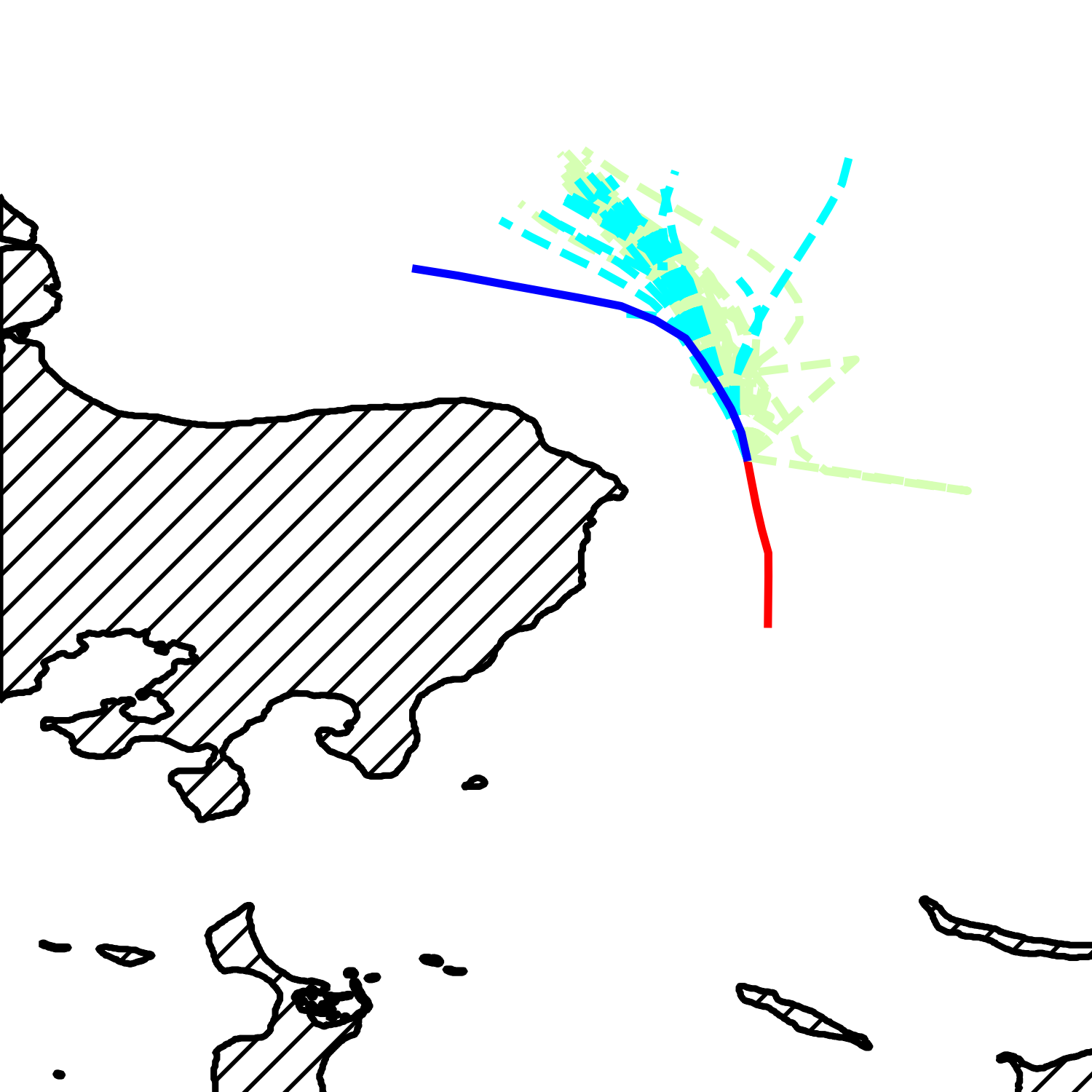}
	}%
\hspace{8pt}
         \subfigure{
			\includegraphics[width=1.16in]{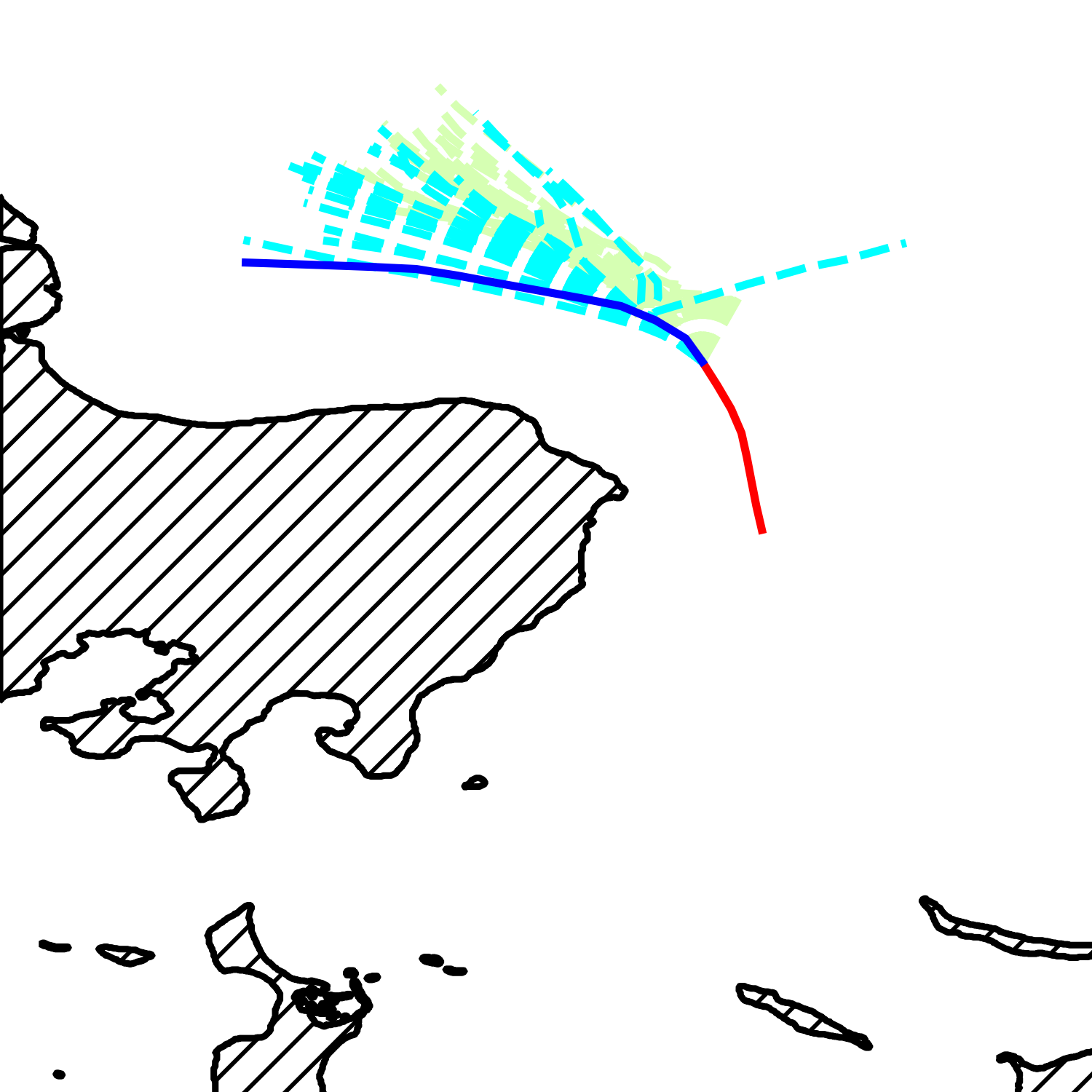}
\hspace{8pt}
        \subfigure{
			\includegraphics[width=1.16in]{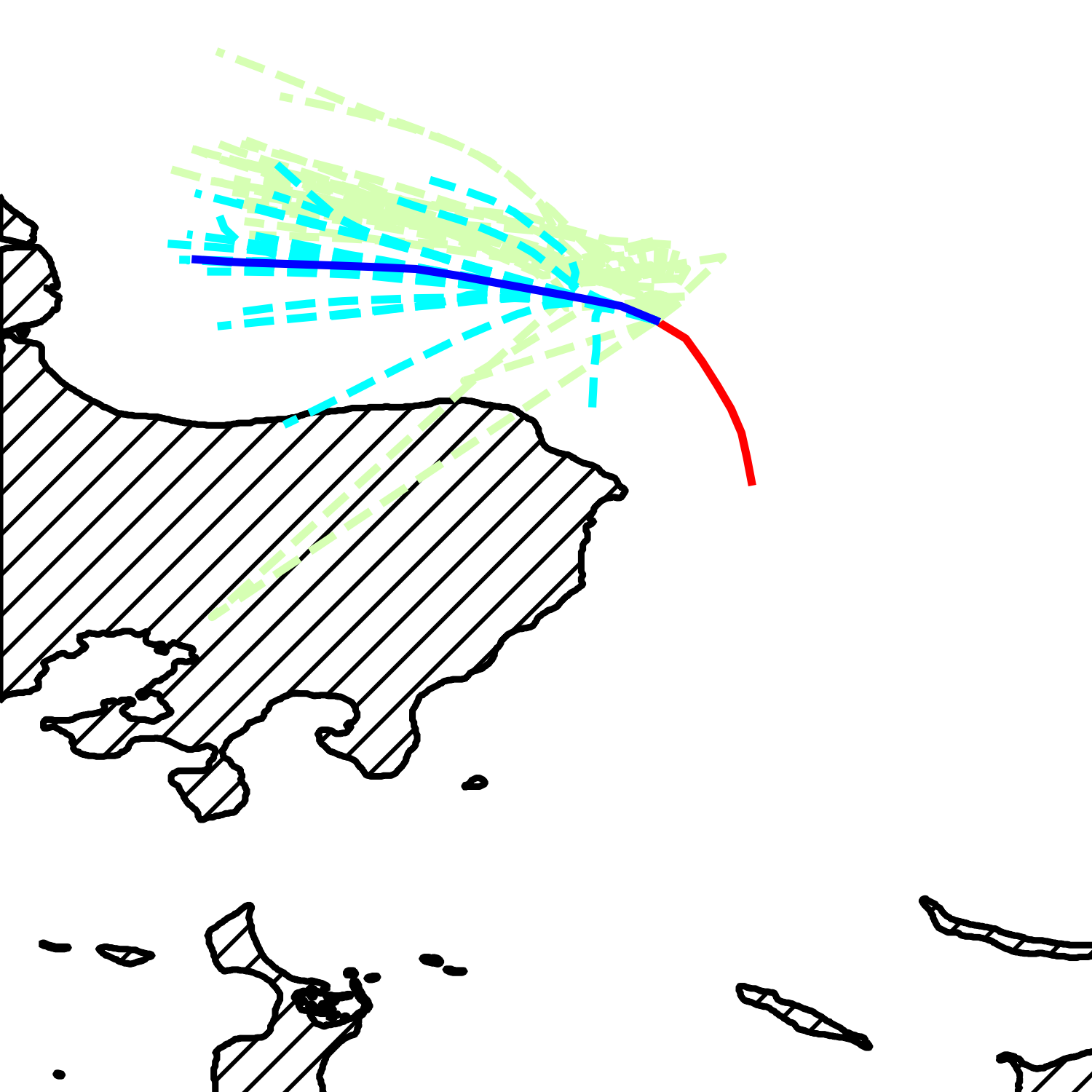}
\hspace{8pt}
	}%
        \subfigure{
			\includegraphics[width=1.16in]{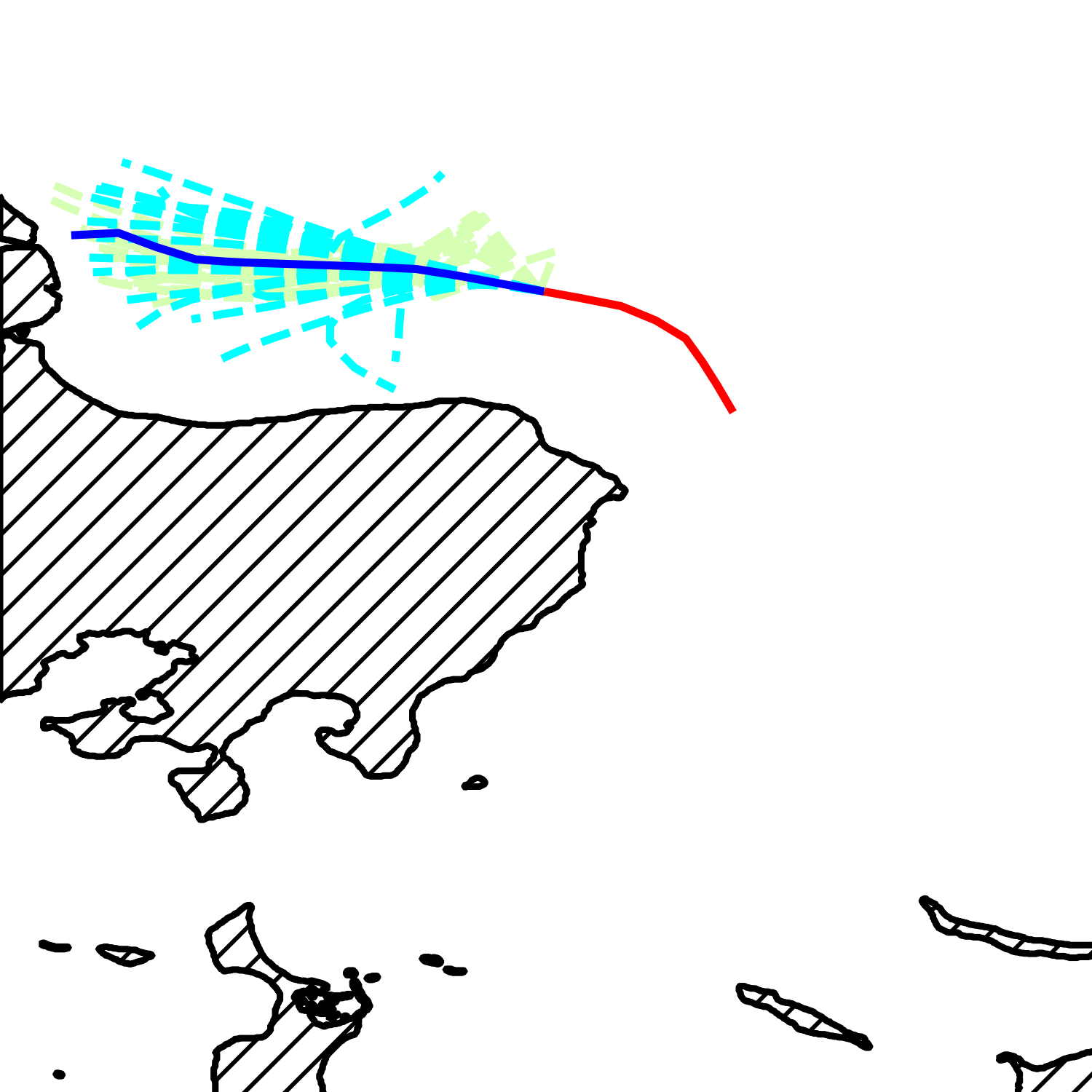}
	}%
	}%
	\centering
	\caption{Visualization of predicted best-of-20 trajectories by DiffuTraj (dashed cyan) and CVAE (light green) in different scenes, with the ground truth paths denoted in blue and observed trajectories in red. Our framework produces future trajectories that demonstrate both diversity and precision. Best viewed in color and Zoom-in for enhanced detail.}
	\label{fig8}
\end{figure*}

\subsection{Ablation Study}
% \subsubsection{Analysis of Diffusion Model}
% \mbox{}\\
% In order to examine the importance of our diffusion framework, we replaced the decoder to verify whether the performance improvement comes from the Transformer. 
\subsubsection{Analysis of Conditions}
To thoroughly assess the impact of each critical conditional component within our proposed model, we conducted comparative analyses between DiffuTraj and its various modified versions:

\begin{itemize}
    \item \textbf{DiffuTraj-w/o His}: History trajectory data is excluded to assess its impact, while maintaining the trajectory heatmap for the scene context module's functionality.

    \item \textbf{DiffuTraj-w/o Neigh}: Neighboring trajectory information is omitted to evaluate spatial interaction effects among vessels.
    
    \item \textbf{DiffuTraj-w/o Map}: The scene context and trajectory heatmap are omitted to assess the impact on physical and motion constraints.

    \item \textbf{DiffuTraj-w/o His\&Neigh}: Both history and neighboring features are removed to test the significance of the joint component.

    \item \textbf{DiffuTraj-w/o Map\&Neigh}: Only the historical trajectory information is retained as the condition.

    \item \textbf{DiffuTraj-w/o All}: Represents the model in an unconditional form, without specific conditional inputs.
\end{itemize}
\mbox{}\\
\textbf{Importance of the History Information} \quad The impact of historical trajectory data on prediction accuracy is assessed by omitting the past trajectories of the target vessel. Table \ref{tab2} and Table \ref{tab3} reveal that both DiffuTraj and its variant without historical data, DiffuTraj-w/o His, experience performance degradation with increasing forecast horizons. Notably, the absence of historical data leads to a pronounced decline in accuracy, resulting in an average increase of 68.4\% in ADE and 67.5\% in FDE. Moreover, comparing DiffuTraj-w/o His\&Neigh and DiffuTraj-w/o Neigh, the removal of historical information results in average increases of 70.3\% and 66.3\% in ADE and FDE, respectively. This underscores the critical nature of past motion states in trajectory prediction, highlighting their role in guiding for forecasting future vessel motion.
\mbox{}\\
\textbf{Importance of the Spatial Interactions} \quad To understand the influence of neighboring vessels, we incorporate spatial interaction data into DiffuTraj. While its contribution is less dramatic compared to historical data, it remains noteworthy. As shown in Table \ref{tab2} and Table \ref{tab3}, the removal of neighboring information leads to a modest increase in ADE and FDE by 6.0\% and 7.2\%, respectively, on average. When comparing DiffuTraj-w/o His\&Neigh with DiffuTraj-w/o His, the removal of this module results in an average increase of 7.3\% in ADE and 7.0\% in FDE. This suggests that while a vessel's long-term navigational objectives across a sea area predominantly dictate its overall trajectory, the impact of nearby vessels is more localized and significant when they are in close proximity.
% Besides, we compare DiffuTraj-w/o His&Neighbor with DiffuTraj-w/o His to further examine it under the absence of past trajectories, resulting in an average decrease of \(2.5\%\) in ADE and \(4\%\) in FDE.
\mbox{}\\
 \textbf{Importance of the Trajectory-On-Map Representation} \quad To evaluate the impact of the Trajectory-On-Map Representation, we conduct tests with this component removed from DiffuTraj. The results, presented in Table \ref{tab2} and Table \ref{tab3}, show a marked difference between DiffuTraj and DiffuTraj-w/o Map. Notably, the absence of scene context information leads to an average increase of 44.8\% in ADE and 52.1\% in FDE. Moreover, in comparison between DiffuTraj-w/o Map\&Neigh and DiffuTraj-w/o Neigh, ADE increases by 38.0\% and FDE increases by 45.0\%. Furthermore, the scene context representation demonstrates a gradually increasing impact over longer term prediction. Specifically, ADE and FDE increase by 0.8\% and 1.0\% at the 0.5-hour mark, respectively, These increases escalate to 74.0\% in ADE and 79.4\% in FDE at the 3.5-hour mark. This trend indicates that the Trajectory-On-Map Representation's influence grows significantly with longer prediction horizons. This finding confirms the vital role of scene context in constraining and guiding future vessel trajectories, particularly for long-term forecasts.

Meanwhile, we compare DiffuTraj with DiffuTraj-w/o All to assess the overall impact of three conditions mentioned above. Experimental results show that both ADE and FDE increase by an average of 5 times, in the case of unconditional trajectory generation, highlighting the significant influence of additional information on guiding accurate trajectory generation.

\subsubsection{Analysis of Network Architecture}
In order to further analyze the impact of the decoder network architecture, we conduct ablation experiments on the backbone network structure, as illustrated in Figure \ref{fig3}. Both the Transformer and LSTM architectures significantly outperform the Empty model, indicating the effectiveness of the backbone network in capturing the spatio-temporal dependencies in the fused feature maps. Additionally, the Transformer architecture exhibited a slight performance advantage over the LSTM architecture, suggesting its stronger contextual modeling capability. Moreover, compared to the optimal baseline, DiffuTraj without the decoder backbone network performs on par with it in terms of ADE and better in terms of FDE.

\subsubsection{Analysis of Hyperparameter Sensitivity}
In this section, we study the effect of different sampling steps $\gamma$ during inference on our model, which controls the number of iterations required for the reverse diffusion process. Figure \ref{fig4} presents the average and final displacement errors of DiffuTraj with different $\gamma$ values within the range of 100 at various prediction horizons in the Danish Straits dataset. Across different parameter settings, the prediction error steadily increases with the prediction time horizon. Within the first 2 hours, DiffuTraj exhibits the lowest average displacement errors at $\gamma=5$. However, for longer term predictions, $\gamma=2$ demonstrates superior performance over a span of 4 hours, significantly surpassing the performance of the model at $\gamma=5$. Specifically, the reductions in ADE and FDE are 10.8\% and 10.4\%, respectively, under the setting of 4 hours, while the difference in displacement errors remains within 1.5\% under the setting of 2 hours. Besides, employing longer sampling steps fails to enhance performance. Note that we use $\gamma=5$ as the fixed sampling step in experiments.

\subsubsection{Analysis of Reverse Diffusion Process}
To delve deeper into the reverse diffusion process, we perform 20 sample runs for each reverse diffusion step on all trajectories. For a more intuitive understanding, we create visualizations of example trajectories, showcased in Figure \ref{fig6}. We then calculate both ADE and FDE for these runs, with the results presented in Figure \ref{fig5}. Within our DiffuTraj framework, there is an explicit simulation of vessel motion transitioning from a state of uncertainty towards certainty. Specifically, the trajectory distribution exhibits a higher degree of indeterminacy in the initial steps of the reverse diffusion process, reflecting a wider array of potential outcomes. As the reverse diffusion progresses, we observe a reduction in diversity and an increase in determinacy. This culminates in the generation of plausible and determinate trajectories, illustrating the effectiveness of the reverse diffusion in refining trajectory predictions.

\subsection{Case Study}
Figure \ref{fig7} and Figure \ref{fig9} present the comparison of predicted trajectories by TrAISformer \cite{nguyen2021traisformer}, CVAE \cite{han2023interaction}, LSTM-Seq2Seq \cite{forti2020prediction}, and our DiffuTraj model across diverse maritime scenarios such as \textit{island restrictions}, \textit{narrow waterways}, \textit{multi-vessel encounters}, and \textit{sudden turns}. Note that we use the original TrAISformer for a more convincing comparison. The qualitative analysis reveals that DiffuTraj more accurately aligns with the actual ground truth paths compared to its predecessors. Specifically, in the first two cases, where navigation is restricted by terrain, DiffuTraj outperforms by predicting trajectories that both circumvent physical obstacles and correctly align with navigational paths. In the multi-vessel encounter scenarios, DiffuTraj excels by generating more realistic trajectories for the involved vessels, attributing this success to the sophisticated modeling of historical and neighboring trajectories. In scenarios involving sudden directional changes, DiffuTraj adeptly predicts the vessel's next move, demonstrating a deep understanding of motion patterns specific to different regions. Furthermore, Figure \ref{fig8} displays a comparison of multiple trajectory predictions by CVAE and DiffuTraj. This comparison highlights the variance in potential future trajectories as predicted by each model. Notably, DiffuTraj's approach to explicitly model trajectory uncertainty via a denoising process showcases superior flexibility and precision over CVAE's latent variable-based multimodal approach. This methodological distinction enables DiffuTraj to more closely match the true distribution of potential vessel movements, underscoring the effectiveness of our framework in complex maritime forecasting tasks.

\section{Conclusion}
In this study, we present DiffuTraj, a novel stochastic framework for predicting vessel trajectories through a guided motion pattern uncertainty diffusion process. This method effectively reduces uncertainty within maritime regions to accurately determine intended trajectories. By carefully managing each stage of the reverse diffusion process, DiffuTraj integrates scene context with vessel spatial interactions. This approach facilitates the creation of trajectories that are attuned to spatial discrepancies while remaining anchored in physical authenticity. Empirical evaluations on a real-world AIS dataset demonstrate that our framework delivers state-of-the-art results, particularly excelling in scenarios characterized by complex terrain and dynamic vessel interactions.

\ifCLASSOPTIONcaptionsoff
  \newpage
\fi

\bibliography{sample-base}
\bibliographystyle{IEEEtran}

\par\noindent 
\parbox[t]{\linewidth}{
\noindent\parpic{\includegraphics[height=1.9in,width=1.2in,clip,keepaspectratio]{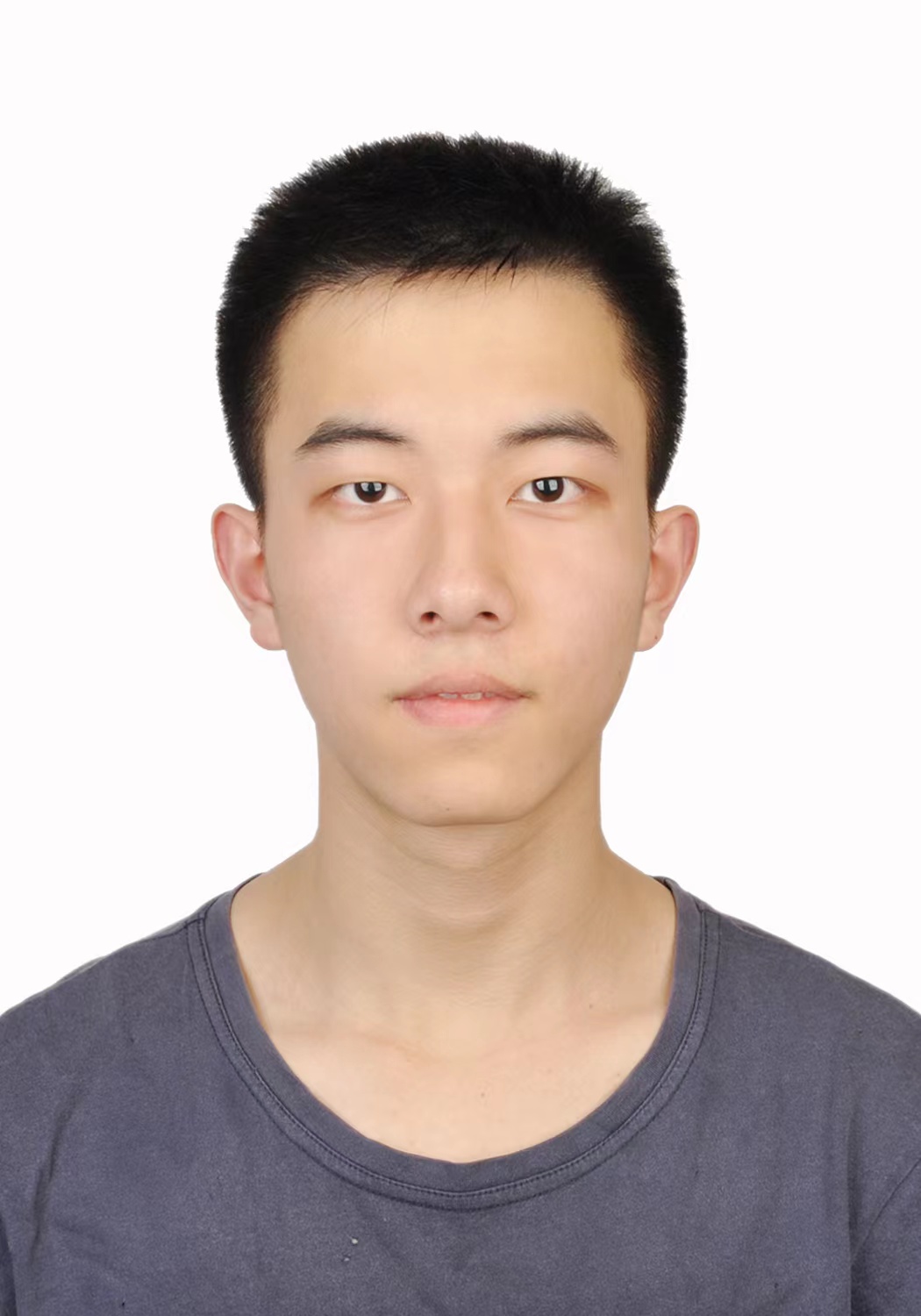}}
\noindent {\bf Changlin Li}\
received the B.E degree from Southwest Jiaotong University, Chengdu, China, in 2023, where he is currently pursuing the M.E degree in School of Computer Science and Engineering, University of Electronic Science and Technology of China. His research interests include machine learning, trajectory data mining, and intelligent transportation systems.}
\vspace{2\baselineskip}

\par\noindent 
\parbox[t]{\linewidth}{
\noindent\parpic{\includegraphics[height=1.9in,width=1.2in,clip,keepaspectratio]{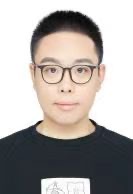}}
\noindent {\bf {Yanglei Gan}}\
earned his B.S degree from University of Connecticut, Storrs, CT in 2018 and his M.S degree from Boston University, Boston, MA in 2020. He is currently a Ph.D. student in School of Computer Science and Engineering, University of Electronic Science and Technology of China. His research interests include information extraction, knowledge graph, and time series analysis.}
\vspace{2\baselineskip}

\par\noindent 
\parbox[t]{\linewidth}{
\noindent\parpic{\includegraphics[height=1.9in,width=1.2in,clip,keepaspectratio]{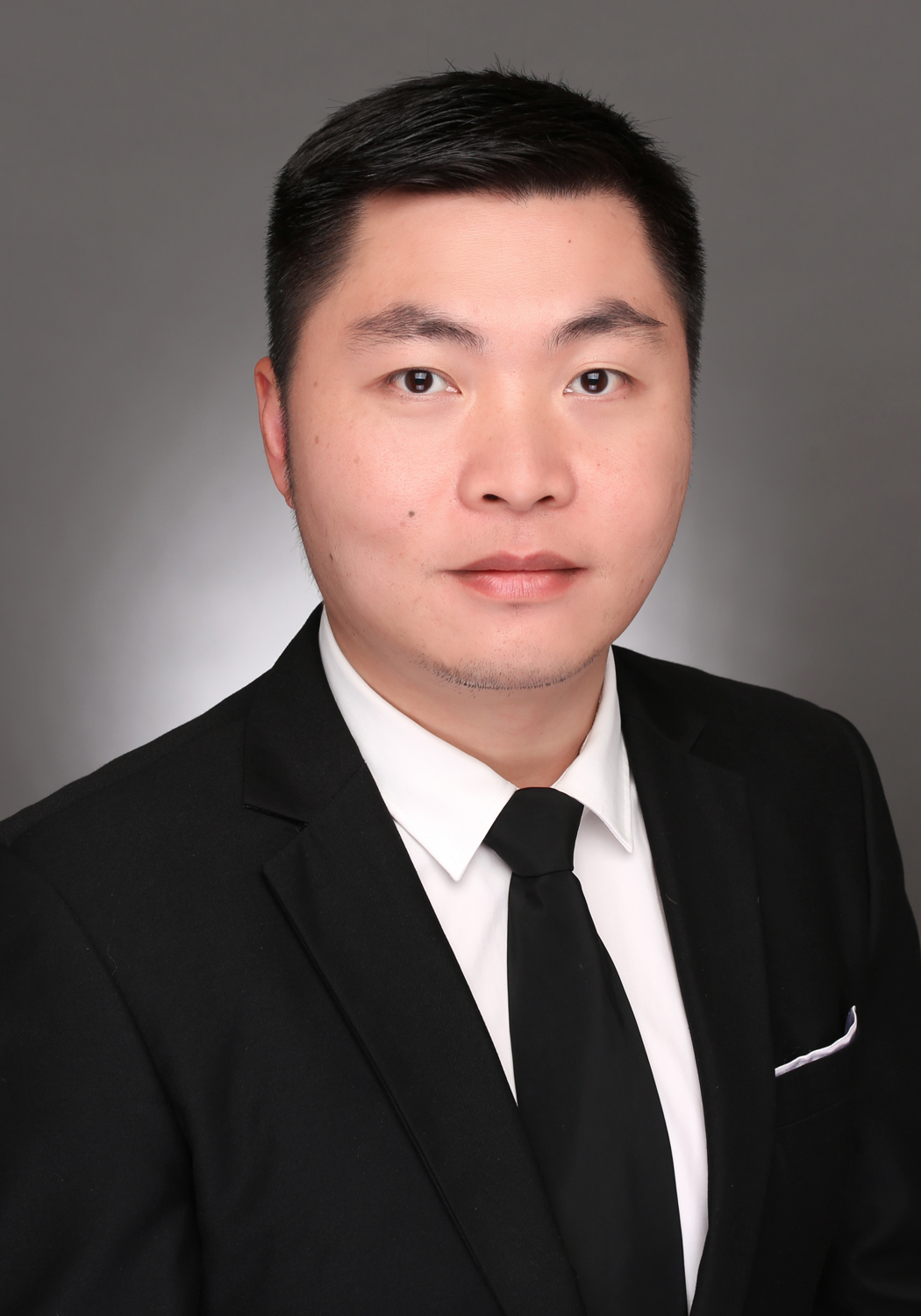}}
\noindent {\bf {Tian Lan}}\
(Member, IEEE) received the Ph.D.
degree in computer science from the University
of Electronic Science and Technology of China (UESTC), Chengdu, China, in 2009. He is currently an Associate Professor with the School of Information and Software Engineering, UESTC. His research interests include medical image processing, speech enhancement, and natural language processing.}
\vspace{2\baselineskip}

\par\noindent 
\parbox[t]{\linewidth}{
\noindent\parpic{\includegraphics[height=1.9in,width=1.2in,clip,keepaspectratio]{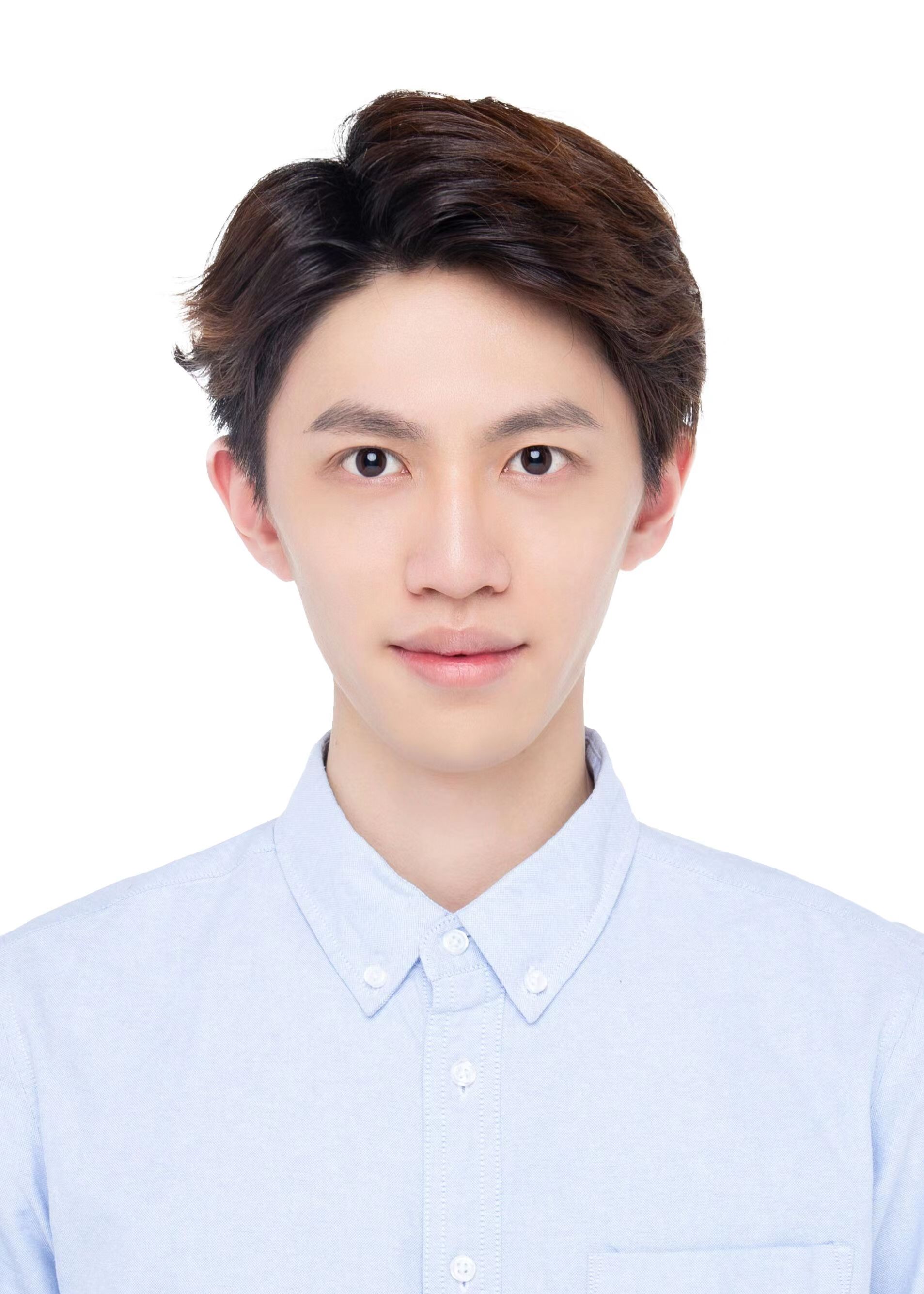}}
\noindent {\bf {Yuxiang Cai}}\
received the B.E. degree from Chongqing University of Posts and Telecommunications, Chongqing, China, in 2023. He is currently pursuing the M.E. degree with
the School of Computer Science and Engineering, University of Electronic Science and Technology of China, China. He has been working on natural language processing (NLP) and data mining.}
\vspace{2\baselineskip}

\par\noindent 
\parbox[t]{\linewidth}{
\noindent\parpic{\includegraphics[height=1.9in,width=1.2in,clip,keepaspectratio]{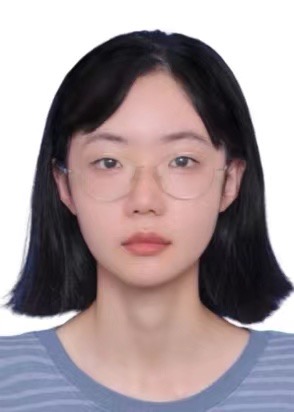}}
\noindent {\bf {Xueyi Liu}}\
received the B.E. degree from Southwest University, Chongqing, China, in 2023. She is currently pursuing the M.E. degree at the School of Computer Science and Engineering, University of Electronic Science and Technology of China. She has been working on natural language processing (NLP), data mining, and intelligent transportation systems.}
\vspace{3\baselineskip}

\par\noindent 
\parbox[t]{\linewidth}{
\noindent\parpic{\includegraphics[height=1.9in,width=1.2in,clip,keepaspectratio]{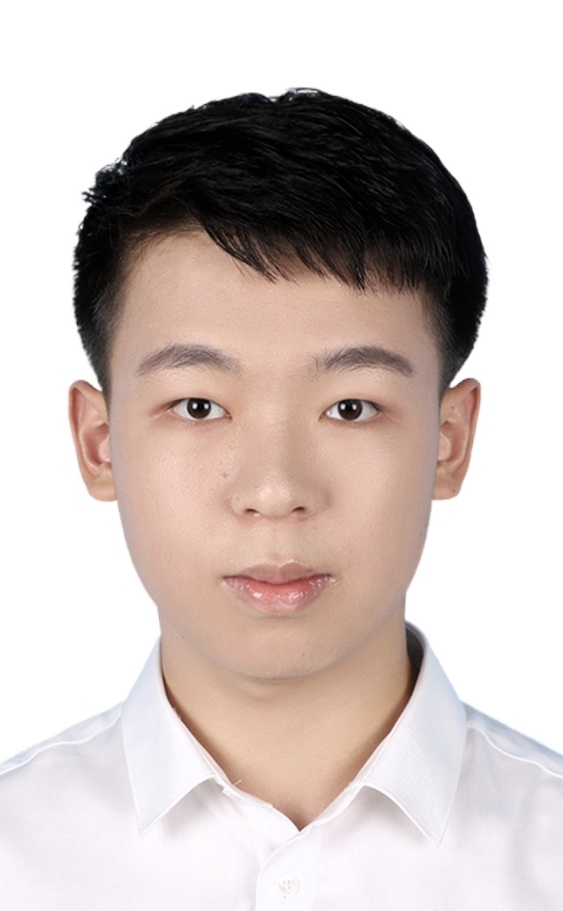}}
\noindent {\bf {Run Lin}}\
received the B.E. degree from China University of Mining and Technology, Beijing, in 2023. He is currently pursuing the M.E. degree at the School of Computer Science and Engineering (School of Cyber Security), University of Electronic Science and Technology of China.
He has been working on natural language processing (NLP) and data mining.}
\vspace{6\baselineskip}

\par\noindent 
\parbox[t]{\linewidth}{
\noindent\parpic{\includegraphics[height=1.9in,width=1.2in,clip,keepaspectratio]{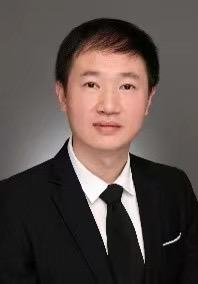}}
\noindent {\bf {Qiao Liu}}\
(Member, IEEE) received the Ph.D.
degree in computer science from the University
of Electronic Science and Technology of China
(UESTC), Chengdu, China, in 2010. He is currently a Full Professor with the School of Information and Software Engineering, UESTC. His current research interests include natural language processing, machine learning, and data mining.}

% \end{IEEEbiography}

\end{document}